\documentclass{article}

\usepackage{microtype}
\usepackage{graphicx}
\usepackage{subfigure}
\usepackage{booktabs} 
\usepackage{siunitx}
\usepackage[switch]{lineno}
\usepackage{newfloat}
\usepackage{listings}
\usepackage{multirow}
\usepackage{rotating}
\usepackage{caption}
\usepackage{subcaption}
\usepackage{tcolorbox}
\usepackage[dvipsnames,table]{xcolor}
\usepackage[normalem]{ulem} 
\usepackage{pifont}
\usepackage{doi}
\usepackage{floatrow}
\usepackage[export]{adjustbox}
\usepackage{longtable}
\usepackage{bbm}

\PassOptionsToPackage{numbers}{natbib}
\usepackage[accepted]{icml2025}
\setcitestyle{numbers}
\setcitestyle{square}

\usepackage{hyperref}

\usepackage{amsmath}
\usepackage{amssymb}
\usepackage{mathtools}
\usepackage{amsthm}

\hypersetup{
    colorlinks=true,
    citecolor=blue,
    linkcolor=blue, 
    urlcolor=cyan,  
}

\newcommand{\highlightfirst}[1]{\textcolor{black}{#1}}
\newcommand{\highlighthipe}[1]{\textcolor{black}{#1}} 
\newcommand{\highlightprism}[1]{\textcolor{black}{#1}}
\newcommand{\highlightsecond}[1]{\textcolor{black}{#1}} 
\newcommand{\highlightthird}[1]{\textcolor{black}{#1}} 
\newcommand{\highlightfour}[1]{\textcolor{black}{#1}} 
\newcommand{\highlightappendix}[1]{\textcolor{black}{#1}}
\newif\iffinalsubmission
\finalsubmissiontrue 

\newcommand{\highlightrev}[1]{\iffinalsubmission #1\else{\color{Red}#1}\fi} 
\newcommand{\highlightrevdel}[1]{\iffinalsubmission\else{\color{Red}\sout{#1}}\fi} 
\newcommand{\highlightrevdelraw}[1]{\iffinalsubmission\else{\color{Red}#1}\fi} 
\newcommand{\cmark}{\textcolor{ForestGreen}{\ding{51}}}
\newcommand{\xmark}{\textcolor{red}{\ding{55}}}

\usepackage[capitalize,noabbrev]{cleveref}

\theoremstyle{plain}

\theoremstyle{definition}

\theoremstyle{remark}

\usepackage[textsize=tiny]{todonotes}

\makeatletter
\newcommand{\safeRef}[2]{%
  \@ifundefined{r@#1}{%
    #2%
  }{%
    \ref{#1}%
  }%
}
\makeatother

\icmltitlerunning{Retrieval-aligned Tabular Foundation Models Enable Robust Clinical Risk Prediction in Electronic Health Records}

\begin{document}

\twocolumn[

\icmltitle{Retrieval-aligned Tabular Foundation Models Enable Robust Clinical Risk Prediction in Electronic Health Records Under Clinically Realistic Constraints}

\icmlsetsymbol{equal}{*}

\begin{icmlauthorlist}

\icmlauthor{Minh-Khoi Pham}{dcu,adapt}
\icmlauthor{Thang-Long Nguyen Ho}{dcu,adapt}
\icmlauthor{Thao Thi Phuong Dao}{vnu,tnh}
\icmlauthor{Tai Tan Mai}{dcu,adapt}
\icmlauthor{Minh-Triet Tran}{vnu,jvn}
\icmlauthor{Marie E. Ward}{hipe}
\icmlauthor{Una Geary}{hipe}
\icmlauthor{Rob Brennan}{adapt,ucd}
\icmlauthor{Nick McDonald}{tcd}
\icmlauthor{Martin Crane}{dcu,adapt}
\icmlauthor{Marija Bezbradica}{dcu,adapt}

\end{icmlauthorlist}

\icmlaffiliation{dcu}{School of Computing, Dublin City University, Dublin, Ireland}
\icmlaffiliation{adapt}{ADAPT Centre, Dublin City University, Dublin, Ireland}
\icmlaffiliation{vnu}{University of Science, Vietnam National University, Ho Chi Minh City, Vietnam}
\icmlaffiliation{tnh}{Department of Otolaryngology, Thong Nhat Hospital, Ho Chi Minh City, Vietnam}
\icmlaffiliation{jvn}{John von Neumann Institute, Vietnam National University, Ho Chi Minh City, Vietnam}
\icmlaffiliation{ucd}{School of Computer Science, University College Dublin, Dublin, Ireland}
\icmlaffiliation{tcd}{School of Psychology, Trinity College Dublin, Dublin, Ireland}
\icmlaffiliation{hipe}{St James’s Hospital, Dublin, Ireland}

\icmlcorrespondingauthor{Minh-Khoi Pham}{minhkhoi.pham@adaptcentre.ie}

\icmlkeywords{
Electronic Health Records, Tabular Foundation Models, Prior-Fitted Networks, Retrieval-Augmented In-context Learning, Patient Outcome Prediction
}

\vskip 0.3in

\begin{abstract}
Accurate clinical outcome prediction from structured electronic health records (EHRs) remains challenging due to high-dimensional heterogeneous features, severe outcome imbalance, and distribution shift. Although recent tabular foundation models, particularly in-context and retrieval-augmented approaches, achieve strong results on generic tabular benchmarks, their behavior under clinical conditions remains unclear.

We present the first systematic multi-cohort EHR benchmark comparing classical machine learning, deep tabular models, and tabular in-context learning (TICL) methods, including prior-fitted network (PFN)-based models and retrieval-augmented variants (RA-TICL), across small clinical datasets and large ICU cohorts. Through stress tests varying data scale, feature dimensionality, outcome rarity, and \highlightrevdel{cross-cohort generalization} \highlightrev{consistency across cohorts}, we identify regime-dependent performance patterns. TICL and RA-TICL models are notably sample-efficient in low-data settings, but they degrade quickly as feature heterogeneity and class imbalance increase.

\highlightrev{To address this limitation, we propose Attention Weighting for Aligned Retrieval Embeddings (AWARE), a retrieval-based adaptation strategy that trains a task-aligned retrieval encoder and a lightweight adapter to specialize a frozen tabular foundation model to each clinical task.} Across stress tests, AWARE improves AUPRC by up to 7.0\% under high feature dimensionality and 12.2\% under extreme outcome imbalance. \highlightrev{These findings establish retrieval quality and retrieval–inference coupling as key bottlenecks for adapting tabular foundation models to clinical prediction tasks.}
\end{abstract}

]

\printAffiliationsAndNotice{} 

\section{Introduction}
\label{sec:intro}

\highlightprism{Accurate prediction from structured electronic health records (EHRs) is central to modern clinical decision support, enabling risk stratification, resource allocation, and timely intervention. The growing availability of routinely collected EHR data offers substantial opportunities to develop predictive Machine Learning (ML) models that capture complex, personalized clinical patterns. However, EHR-based prediction remains methodologically challenging due to high-dimensional heterogeneous features, severe outcome imbalance, irregular temporal structure, and cross-institutional variation \cite{sarwar2022secondaryehr}. }

\highlightrevdelraw{
\begin{figure*}[htp]
    \setlength{\lineskip}{0pt}
    \centering
    \begin{tikzpicture}
        \node[anchor=north west,inner sep=0pt] (img1) at (0,0){\includegraphics[width=0.91\linewidth]{images/ehrlearning.pdf}};
        \node[font=\sffamily\bfseries\large] at (0ex, -2ex) {a};
        
        \draw[red, ultra thick] (img1.south west) -- (img1.north east);
        \draw[red, ultra thick] (img1.south east) -- (img1.north west);
    \end{tikzpicture}
\end{figure*}
}

\highlightrevdel{While recent years have seen rapid advances in foundation models for computer vision and natural language processing, tabular learning, and EHR modeling in particular, has also attracted growing interest, albeit at a more gradual pace and with comparatively fewer large-scale, systematic evaluations [2]. Most EHR foundation models have primarily relied on sequential or language-based representations that emphasize longitudinal structure and large-scale pretraining [3]. Although effective in selected settings, these paradigms often require extensive pretraining on real-world clinical data, operate over long token sequences, and impose complicated standardization of heterogeneous medical events into universal formats. Such design choices introduce (1) substantial training computational and architectural complexity, and (2) practical and governance-related constraints associated with real sensitive patient data, motivating interest in alternative paradigms that are more efficient. This motivates us to adopt a newly alternative approach that has been introduced recently in tabular deep learning, that can address these issues of EHRs. We highlight the main differences between these EHR-based learning paradigms in Figure 1.}

In a clinically realistic electronic health record, we may encounter thousands of variables, such as demographics, comorbidities, laboratory indices, vital signs, diagnosis codes, and procedural interventions~\cite{sarwar2022secondaryehr}, yet for a given task, such as predicting ICU mortality, only a small, often different subset (e.g., serum lactate, blood pressure, or SOFA score) carries substantial prognostic value. This heterogeneity requires ML models to identify clinically meaningful signals within a high-dimensional, largely noisy feature space~\cite{miotto_deeppatient_2016}; for retrieval-augmented approaches specifically, non-informative dimensions can distort the used similarity metrics, yielding neighbors that are not semantically relevant~\cite{neha2025retrieval, shi2024replug}.

\begin{figure*}[htp]
    \setlength{\lineskip}{0pt}
    \centering
    \highlightrevdelraw{
    \begin{tikzpicture}
        \node[anchor=north west,inner sep=0pt] (img1) at (0,0){\includegraphics[width=0.91\linewidth]{images/ehr.pdf}};
        \node[font=\sffamily\bfseries\large] at (0ex, -2ex) {a};
        
        \draw[red, ultra thick] (img1.south west) -- (img1.north east);
        \draw[red, ultra thick] (img1.south east) -- (img1.north west);
    \end{tikzpicture} \\
    \vspace{1.8em}}
    \begin{tikzpicture}
        \node[anchor=north west,inner sep=0pt] at (0,0){\includegraphics[width=0.31\linewidth]{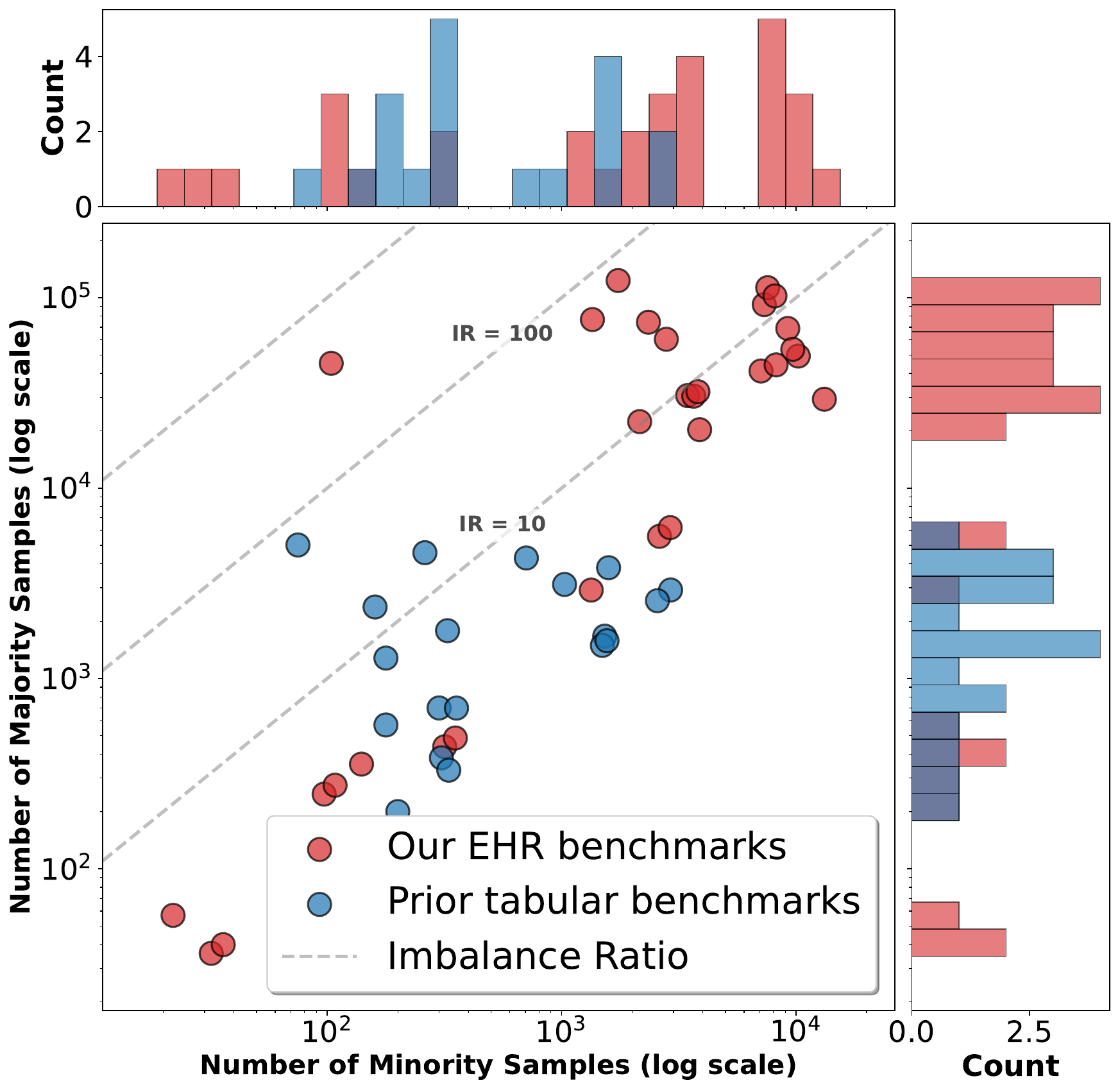} };
        \node[font=\sffamily\bfseries\large] at (0ex, 0ex) {a};
    \end{tikzpicture}
    \begin{tikzpicture}
        \node[anchor=north west,inner sep=0pt] at (0,0){\includegraphics[width=0.31\linewidth]{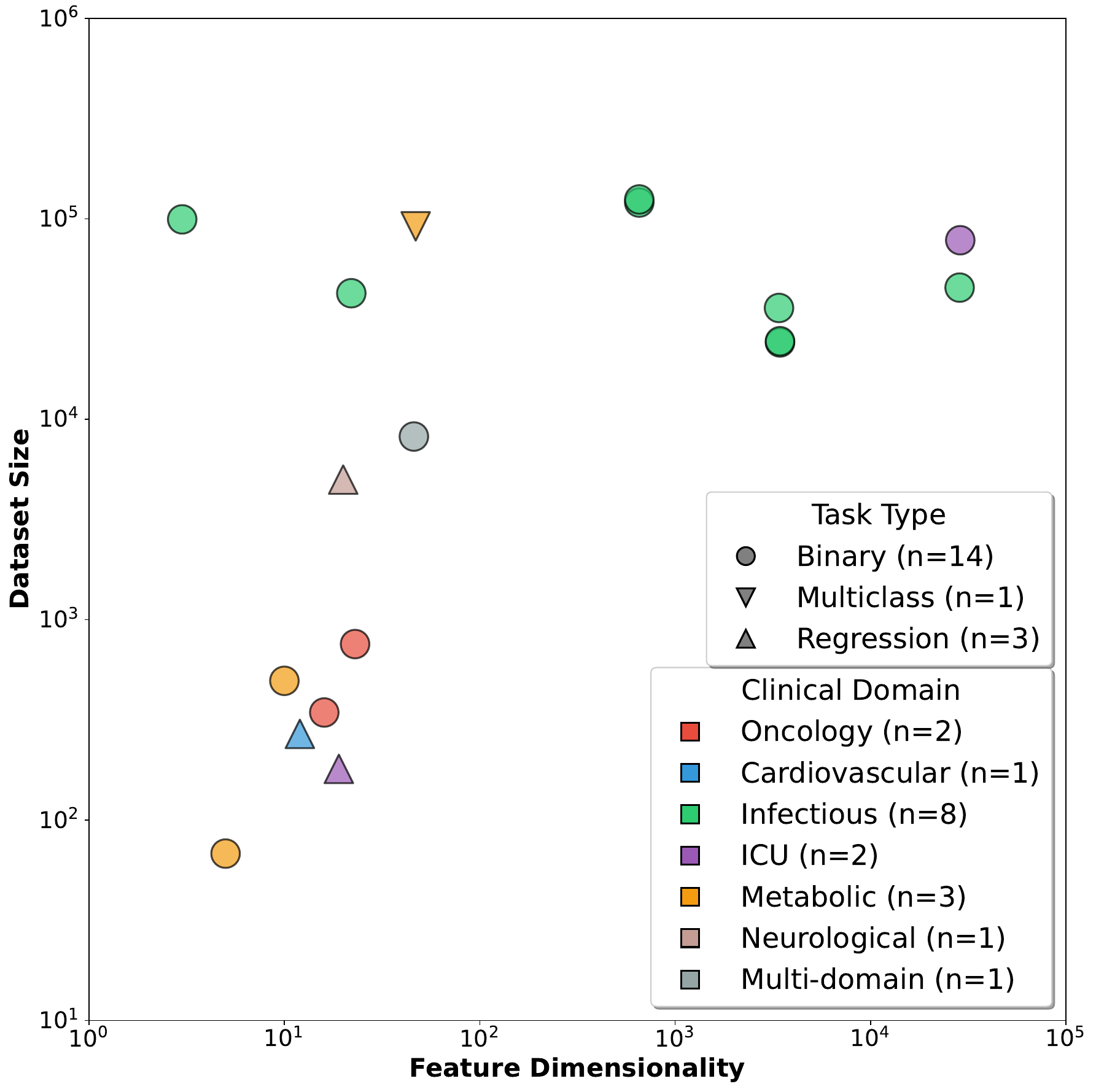}};
        \node[font=\sffamily\bfseries\large] at (0ex, 0ex) {b};
    \end{tikzpicture}
    \begin{tikzpicture}
        \node[anchor=north west,inner sep=0pt] at (0,0){\includegraphics[width=0.31\linewidth]{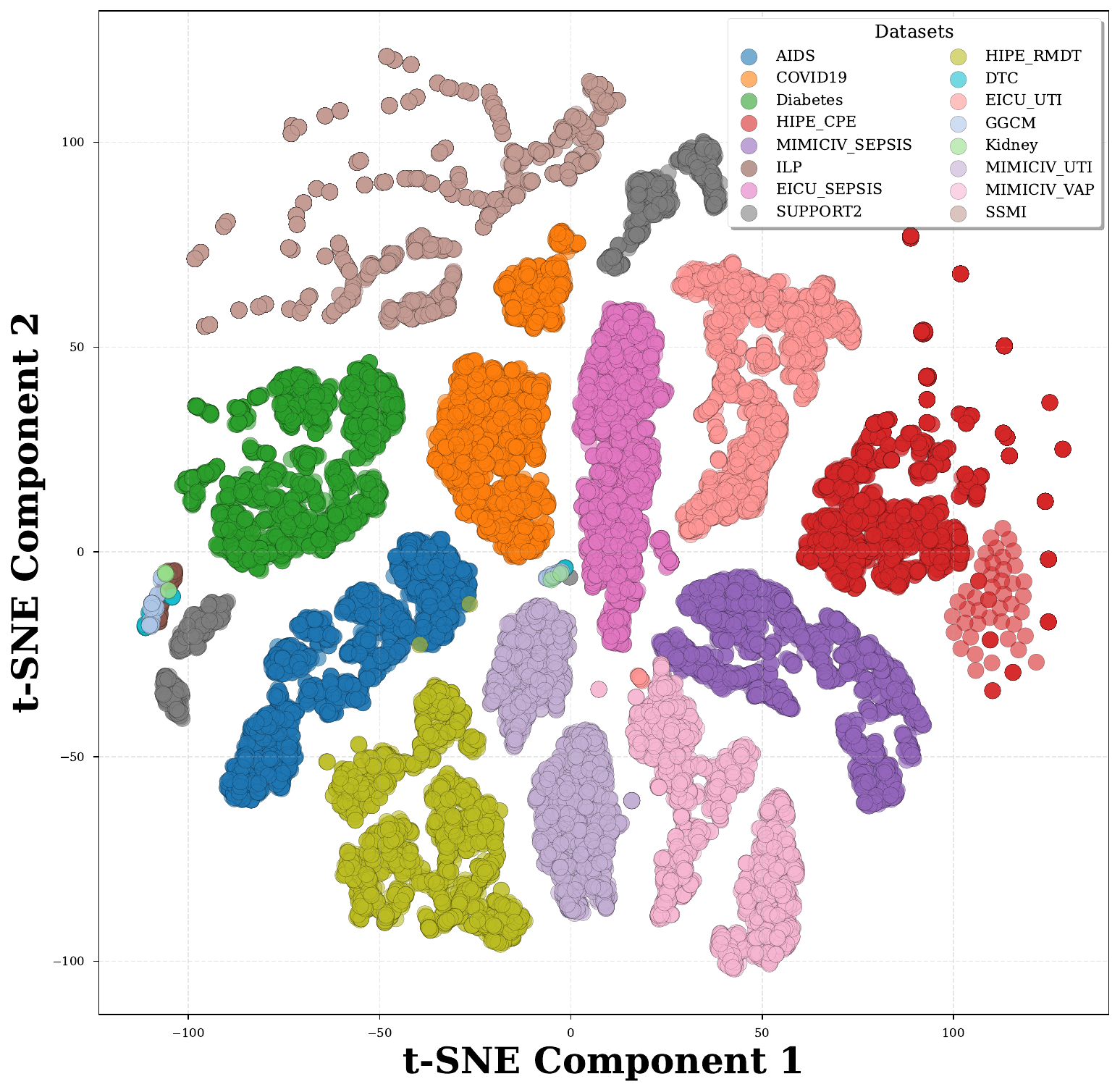}};
        \node[font=\sffamily\bfseries\large] at (0ex, 0ex) {c};
    \end{tikzpicture}
    \caption{Landscape of EHR benchmarks evaluated in this study. The datasets span wide variation in cohort size, feature dimensionality, outcome prevalence, and clinical domain, illustrating key challenges of \highlightrev{(a) rarity, (b) heterogeneity, (c) cross-task, cross-institutional, multi-cohort distribution shift faced by tabular learning methods in real-world EHR settings.}}
    \label{fig:dataset_landscape}
\end{figure*}

EHR-based prediction tasks are further characterized by severe outcome rarity and pervasive distribution shift: clinically significant events, such as hospital-acquired infections, septic shock, or mortality in low-risk populations, are relatively rare~\cite{moor2021early, zhu2026bridging}, producing extreme class imbalance where a model can achieve high aggregate accuracy while failing to detect rare, life-threatening cases~\cite{sarwar2022secondaryehr}. Distribution shift across institutions, departments, and time periods, driven by differences in demographics, treatment protocols, and coding practices, compounds this challenge, as models trained on one cohort often generalize poorly to another; for retrieval-based methods, class imbalance further skews neighborhoods toward majority-class cases, impoverishing contextual support for exactly the rare, high-risk cases that most require reliable prediction~\cite{nejjar2024imcontext}. Figure~\ref{fig:dataset_landscape} illustrates these structural challenges across our benchmark suite, spanning variation in cohort size, feature dimensionality, outcome prevalence, and clinical domain.

Our goal in this paper is to evaluate in-context and retrieval-based tabular learning under non-sequential EHR pipelines, which we call snapshot-based\footnote{or also called static EHR} EHR prediction tasks: each patient encounter is represented by a $d$-dimensional feature vector $x \in \mathbb{R}^d$ comprising numerical, ordinal, and categorical variables, aggregated into a fixed-length tabular representation rather than a longitudinal sequence. We formulate these tasks as supervised learning problems, where the prediction target $y$ may be a discrete label for classification (e.g., infection occurrence) or a continuous value for regression (e.g, length of stay), with multiple tasks possible for a same dataset. This setting differs from prior work~\cite{fallahpour2024EHRMamba, odgaard_core-behrt_2024} on trajectory-dependent outcomes (e.g., long-horizon readmission or mortality trajectories) that require explicit sequence modeling.



This snapshot-based formulation naturally places the problem within the tabular learning paradigm, which has long been central to EHR-based clinical prediction \cite{jensen2012mining}. In practice, this setting has been dominated by decision tree ensembles, most notably gradient-boosted methods such as XGBoost~\cite{chen_xgboost_2016}, LightGBM~\cite{ke2017lightgbm}, and CatBoost~\cite{prokhorenkova2018catboost}, which remain strong baselines for clinical outcome prediction due to their robustness to heterogeneous feature types and favorable interpretability--performance trade-offs~\cite{grinsztajn_why_2022, mcelfresh_when_2024}. These properties have made tabular models widely adopted across diverse EHR-based supervised learning tasks and institutional settings.

Motivated by the success of deep learning in other domains, a growing body of work has explored neural architectures tailored to tabular data. Models such as TabNet~\cite{arik2021tabnet} and TabTransformer~\cite{huang_tabtransformer_2020} introduced attention-based feature selection and contextualized embeddings, while more recent approaches, including ResNet-style architectures~\cite{gorishniy_revisiting_2023}, TabM~\cite{gorishniy2024tabm}, ExcelFormer~\cite{chen_excelformer_2024}, Trompt~\cite{chen2023trompt}, and DANets\cite{chen2022danets}, all aim to learn task-specific representations end-to-end in large-data regimes. A common theme across these models is the explicit modeling of feature relevance and structure: rather than treating all variables as equally informative, they incorporate mechanisms that implicitly select, weight, or group features to suppress noise from weakly relevant or redundant inputs. These models have been applied to diverse EHR-based clinical prediction tasks, including cardiovascular disease risk stratification~\cite{tabnetehr_natchathirah_2025, sumon2025cardiotabnet}, glaucoma surgery prediction using multimodal EHR and imaging features~\cite{koornwinder2025multimodal}, Carbapenemase-Producing Enterobacteriace (CPE) infection forecasting \cite{pham2025explainable}, preeclampsia risk prediction~\cite{preeclampsia_rabby_2025}, obstructive sleep apnea prediction \cite{chi2026osa}, and hepatocellular carcinoma risk estimation~\cite{park2025transformer}.  Table~\ref{tab:models_summary} provides a consolidated overview of all benchmarked models in this study, situating each method by its adaptation strategy, training paradigm, and suitability for core challenges in EHR-based clinical prediction.

\begin{table*}[htbp]
\centering
\caption{
Comparison of tabular learning paradigms evaluated in this study and their adaptation characteristics for EHR-based clinical prediction.
}
\label{tab:models_summary}
\resizebox{0.8\linewidth}{!}{
\renewcommand{\arraystretch}{1.05}
\begin{tabular}{lcccccc}
\toprule
\multirow{3}{*}{\textbf{Model}} &
\multicolumn{2}{c}{\textbf{Adaptation}} &
\multicolumn{2}{c}{\textbf{Training}} &
\multicolumn{2}{c}{\textbf{EHR Properties}} \\
\cmidrule(lr){2-3}
\cmidrule(lr){4-5}
\cmidrule(lr){6-7}
&
\multirow{2}{*}{\textbf{In-context}} &
\multirow{2}{*}{\textbf{Retrieval}} &
\multirow{2}{*}{\textbf{Pretrained}} &
\multirow{2}{*}{\textbf{Task}} &
\multirow{2}{*}{\textbf{Feature}} &
\multirow{2}{*}{\textbf{Imbalance}} \\ \addlinespace[3pt]
& \textbf{Learning} & \textbf{-augmented} & & \textbf{-adaptive} &  \textbf{Weighting} & \textbf{Handling}
\\
\midrule

\multicolumn{7}{l}{\textit{Classical tabular and tree-based models}} \\
\midrule
Logistic Regression & \xmark & \xmark & \xmark & \cmark & \cmark & \xmark \\
k-Nearest Neighbors & \xmark & \cmark & \xmark & \xmark & \xmark & \xmark  \\
XGBoost \cite{chen_xgboost_2016} & \xmark & \xmark & \xmark & \cmark & \cmark & \xmark \\
LightGBM \cite{ke2017lightgbm} & \xmark & \xmark & \xmark & \cmark & \cmark & \xmark \\
CatBoost \cite{prokhorenkova2018catboost} & \xmark & \xmark & \xmark & \cmark & \cmark & \xmark \\
MLP & \xmark & \xmark & \xmark & \cmark & \xmark & \xmark  \\

\midrule
\multicolumn{7}{l}{\textit{Deep tabular models}} \\
\midrule
TabNet \cite{arik2021tabnet} & \xmark & \xmark & \xmark & \cmark & \cmark & \xmark \\
TabTransformer \cite{huang_tabtransformer_2020} & \xmark & \xmark & \xmark & \cmark & \xmark & \xmark  \\
ResNet \cite{gorishniy_revisiting_2023} & \xmark & \xmark & \xmark & \cmark & \xmark & \xmark  \\
TabM \cite{gorishniy2024tabm} & \xmark & \xmark & \xmark & \cmark & \xmark & \xmark \\
Trompt \cite{chen2023trompt} & \xmark & \xmark & \xmark & \cmark & \cmark & \xmark    \\
ExcelFormer \cite{chen_excelformer_2024} & \xmark & \xmark & \xmark & \cmark & \cmark & \xmark  \\

\midrule
\multicolumn{7}{l}{\textit{Tabular in-context learning models}} \\
\midrule
ModernNCA \cite{ye2024revisiting} & \cmark & \cmark & \xmark & \cmark & \xmark & \xmark \\
TabPFN \cite{hollmann_accurate_2025} & \cmark & \xmark & \cmark & \xmark & \xmark & \cmark \\
kNNPFN \cite{thomas_retrieval_nodate} & \cmark & \cmark & \cmark & \xmark & \xmark & \xmark  \\
TabDPT \cite{ma_tabdpt_2025} & \cmark & \cmark & \cmark & \xmark & \xmark & \xmark  \\
\midrule
\midrule

\textbf{\emph{AWARE + RA-TICL (Ours)}} &
\cmark &
\cmark &
\cmark &
\cmark &
\cmark &
\cmark  \\

\bottomrule
\end{tabular}}
\end{table*}

Despite these advances, most deep tabular models for \highlightprism{EHR-based} clinical prediction remain \emph{task-specific}: they are trained to solve a single outcome at a time and typically require a moderate amount of labeled data for each new task \cite{brown2025large}. This paradigm limits scalability in real-world EHR settings, where institutions often face numerous prediction tasks across heterogeneous cohorts \cite{pang2025cehr}. With the increasing availability of large-scale EHR data and growing computational resources, recent work has therefore shifted toward \emph{foundation-style learning} for EHRs~\cite{ khan_comprehensive_2024,moor_foundation_2023}, aiming to build models that support reuse, transfer, and adaptation across multiple tasks and clinical contexts. Existing efforts in this direction also include language-based abstractions~\mbox{\cite{peng2023study, hegselmann_large_2025}} and sequential event-based representations~\mbox{\cite{wornow_shaky_2023, rasmy_med-bert_2021, li_behrt_2020}}, and tabular in-context paradigm \cite{tran2024predicting, karabacak2024advancing}. We explicitly focus on the latter in this work.

\highlightrevdel{Language-based abstractions convert structured EHR features into text to enable prompt-based inference with large language models, but rely heavily on prompt engineering and text abstractions that may obscure clinically relevant structure~\mbox{\cite{peng2023study, lee_emergency_2024}}. Sequential event-based representations instead model longitudinal EHR trajectories directly, for example via Transformer-based architectures such as Med-BERT~\mbox{\cite{rasmy_med-bert_2021}} and BEHRT~\mbox{\cite{li_behrt_2020}}, achieving strong performance on temporally complex tasks at the cost of large event vocabularies, rigid standardization pipelines, and extensive pretraining~\mbox{\cite{wornow_shaky_2023}}.}

Tabular in-context learning (TICL) offers a foundation-style paradigm for inference-time adaptation: TabPFN~\cite{hollmann_accurate_2025}\footnote{We refer to TabPFNv2 simply as TabPFN.} is a prior-fitted transformer trained on synthetic tabular tasks that adapts to new classification problems without gradient-based fine-tuning, and subsequent work has extended this framework with improved architectures~\cite{garg2025real, ma_tabdpt_2025, qu_tabicl_2025}. These methods instantiate \emph{Prior-Fitted Networks (PFNs)}~\cite{muller2021transformers}, which approximate amortized Bayesian inference over tabular datasets: given a test input $x$, prediction target $y$ and context $D$, PFNs learn $q_\theta(y \mid x, D) \approx p(y \mid x, D)$ by minimizing
\begin{equation}
\mathcal{L}_{\text{PFN}} =
\mathbb{E}_{(x, y, D)}
\left[ -\log q_\theta(y \mid x, D) \right].
\end{equation}
where each encounter is tokenized and processed via self-attention; predictions are then conditioned on a prompt of labeled reference examples and unlabeled queries, without updating model parameters. However, existing applications of PFN-based TICL to clinical prediction remain limited to clean, small, curated datasets \cite{tran2024predicting, karabacak2024advancing}, leaving it unclear how these methods behave under the constraints of real-world EHRs. In particular, clinical datasets are not only high-dimensional with sparse, noisy signal~\cite{sarwar2022secondaryehr} but also large in scale, making it infeasible to condition on the whole training cohort at inference time and thus requiring careful \emph{context construction}~\cite{qin2021retrieval, zheng2023dense}

\begin{figure*}[ht]
    \setlength{\lineskip}{0pt}
    \begin{tabular}{|l|}
    \hline
    \begin{tikzpicture}
        \node[font=\sffamily\bfseries\normalsize] at (-2, 0ex) {a. Raw EHR Projection: Heterogeneous, Sparse and Task-Agnostic};
        \node[anchor=north west,inner sep=0pt] at (-8,-0.5){\includegraphics[width=0.8\linewidth]{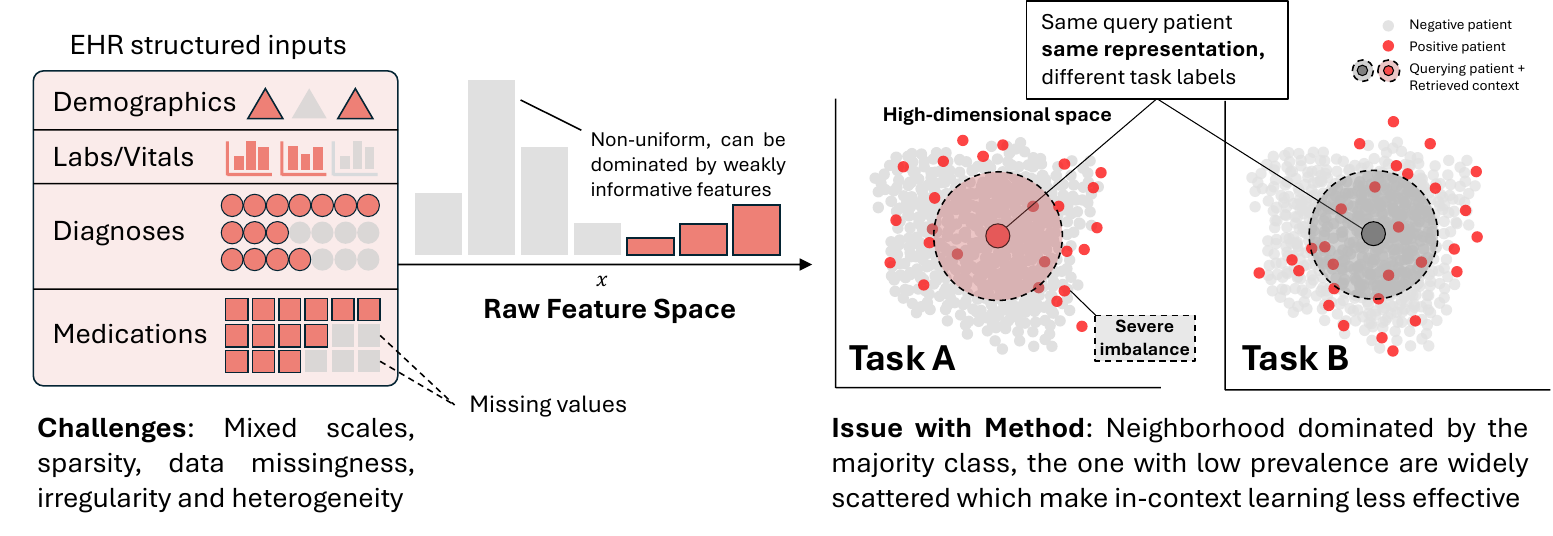}};
    \end{tikzpicture} \\
    \hline
    \begin{tikzpicture}
        \node[anchor=north west,inner sep=0pt] at (,0){\includegraphics[width=0.8\linewidth]{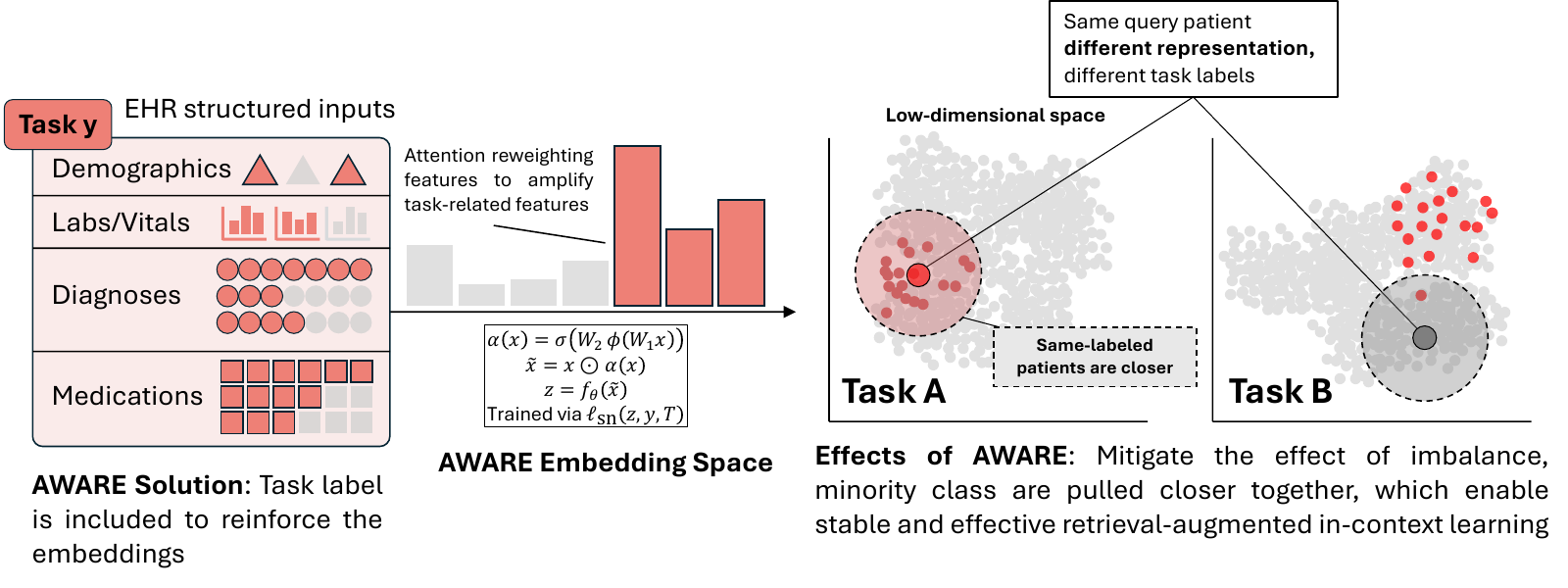}};
        \node[font=\sffamily\bfseries\normalsize] at (4.5, 0ex) {b. Our proposed AWARE Projection};

    \end{tikzpicture} \\
    \hline
    \end{tabular}
    \caption{\highlightthird{The challenges of raw EHR data for retrieval and the solution of task-aware alignment. (A) Raw vector is heterogeneous, sparse and task-agnostic, leading to noisy geometry and multi-task misalignment. (B) Task-aligned projection reshapes geometry for label-consistent, effective retrieval.}}
    \label{fig:aware_vs_baseline}
\end{figure*}

Retrieval-augmented variants of TICL (RA-TICL)~\cite{thomas_retrieval_nodate, wen2025scalable,gorishniy2023tabr,ye2024revisiting,xu_mixture_2024} address this by conditioning on retrieved training subsets\highlightrev{, and have seen clinical adoption via distance-based retrieval architectures such as TabR and kNN, including for cancer diagnosis, mortality, and readmission prediction~\cite{prasetyo2024exploring,wang2022sequential,zhiyun2025detectingoral}}. Formally, RA-TICL constructs, for each query $\mathbf{x}_q$, a context set
\begin{equation}
    D_{\text{context}}(\mathbf{x}_q)
    =
    \operatorname*{arg\,top\text{-}k}_{\mathbf{x}_i \in D_{\text{train}}}
    \; s(\mathbf{x}_q, \mathbf{x}_i),
    \label{eq:retrieval}
\end{equation}
where $s(\cdot,\cdot)$ is a similarity function and $k \ll |D_{\text{train}}|$; the retrieved set forms the in-context prompt. Retrieval-augmented prompting have been shown to improve robustness and accuracy of language foundation models on clinical texts~\cite{lewis2020rag,yang2025retrieval,brown2025large, neha2025retrieval}. However, regarding large-sized EHRs, their noisy features can significantly distort similarity metrics, and class imbalance skews neighborhoods toward majority-class examples, yielding retrieved encounters that are uninformative for clinical prediction~\cite{qin2021retrieval, zheng2023dense} (Figure \ref{fig:aware_vs_baseline}a). Recent work has begun training retrievers for better task alignment~\cite{lin2024ra, shi2024replug} to mitigate this challenge. Inspired by those, we propose AWARE (Figure~\ref{fig:aware_vs_baseline}b), a learned retrieval framework that optimizes neighborhood quality in embedding space, which scales inference-time conditioning to large, heterogeneous cohorts via compact, query-specific context construction tailored to any clinical prediction tasks.

We evaluate this approach alongside classical machine-learning baselines, deep tabular models, and PFN-based TICL methods across diverse small-scale benchmarks and large real-world EHR cohorts. Our experiments then isolate the effects of feature heterogeneity, class imbalance, and dataset scale, \highlightrevdel{and consistency across cohorts}\highlightrev{while assessing performance consistency across cohorts}, enabling a fine-grained analysis of when inference-time adaptation is effective or degrades and how retrieval quality mediates this behavior.

\noindent This work makes the following contributions:
\begin{itemize}

    \item \textbf{\highlightrevdel{A first}\highlightrev{To our knowledge, the first} systematic, multi-cohort benchmark of tabular foundation models for \highlightprism{EHR-based clinical prediction}.} We evaluate classical machine-learning baselines, deep tabular architectures, and TICL/RA-TICL methods across multiple real-world EHR cohorts and \highlightprism{EHR-based supervised learning tasks}, enabling controlled comparisons under shared preprocessing, tuning, and evaluation protocols. \highlightrev{The benchmark spans regression, binary and multiclass classification, and clinical settings ranging from small datasets to large, highly imbalanced longitudinal cohorts.}

    \item \textbf{An adaptive retrieval framework (AWARE) for RA-TICL.} We introduce \highlightrev{an adaptive framework combining attention-guided task-aligned embedding retrieval, imbalance-aware neighborhood learning, and ensemble retrieval stabilization, together with parameter-efficient adapter fine-tuning for compatible RA-TICL backbones}. These components improve neighborhood quality and prompt--backbone representation alignment under high-dimensional and imbalanced EHR conditions, enabling more effective retrieval-augmented in-context learning for clinical prediction.

    \item \textbf{\highlightprism{Controlled stress tests and retrieval-focused analysis under clinical constraints.}} We systematically vary dataset scale, feature dimensionality, and outcome rarity to characterize the performance of tabular foundation models under clinically relevant data constraints. \highlightrev{Through component-wise ablations across multiple retrieval backbones and analyses of retrieved-neighborhood quality, we identify task-aligned retrieval and minority-class coverage as the primary mechanisms underlying AWARE's improvements.} We show that TICL is most effective in low-data regimes, whereas unaligned retrieval increasingly degrades under larger, more heterogeneous, and more imbalanced settings. \highlightrev{AWARE's relative benefits consequently increase with feature dimensionality and outcome rarity, providing regime-specific guidance on when retrieval alignment is most valuable.}

\end{itemize}

\section{Results}
\label{sec:results}

\highlightprism{This section presents a systematic evaluation of tabular learning paradigms under clinically realistic EHR conditions. \highlightrev{We first conduct progressive ablation studies to isolate the contribution of each AWARE component across multiple retrieval-based backbones (\S\ref{sec:exp:retrieval}). We then benchmark the evaluated models on small- to medium-scale clinical datasets (\S\ref{sec:small_ehr_result}) and large-scale real-world EHR cohorts (\S\ref{sec:large_ehr_result}). Finally, we conduct three controlled stress-test experiments to characterize model behavior as training-set size, feature dimensionality, and outcome rarity are systematically varied (\S\ref{sec:results_ehr_constraints}).}}

\subsection{\highlightrevdel{Effectiveness of AWARE for EHR Retrieval}\highlightrev{Progressive Ablation of AWARE Components}}
\label{sec:exp:retrieval}

\highlightrev{\textbf{AWARE} (formally described later in \S\ref{sec:methods}) incorporates feature-level attention, imbalance-aware training, adapter-based domain adaptation, and ensemble aggregation to improve minority patient retrieval and stabilize context selection for rare infection prediction. Formally, the AWARE framework can be summarized as:
\[
\begin{aligned}
\text{AWARE}
= &\underbrace{\text{SNNL}}_{\text{metric learning}} + \underbrace{\text{attention}}_{\text{feature weighting}} + \underbrace{\text{balance sampling}}_{\text{imbalance correction}}
\\
&+ \underbrace{\text{ensemble}}_{\text{variance reduction}} + \underbrace{\text{adapter tuning}}_{\text{domain adaptation}}.
\end{aligned}
\]}

\begin{figure}[t]
    \centering
    \includegraphics[width=\linewidth]{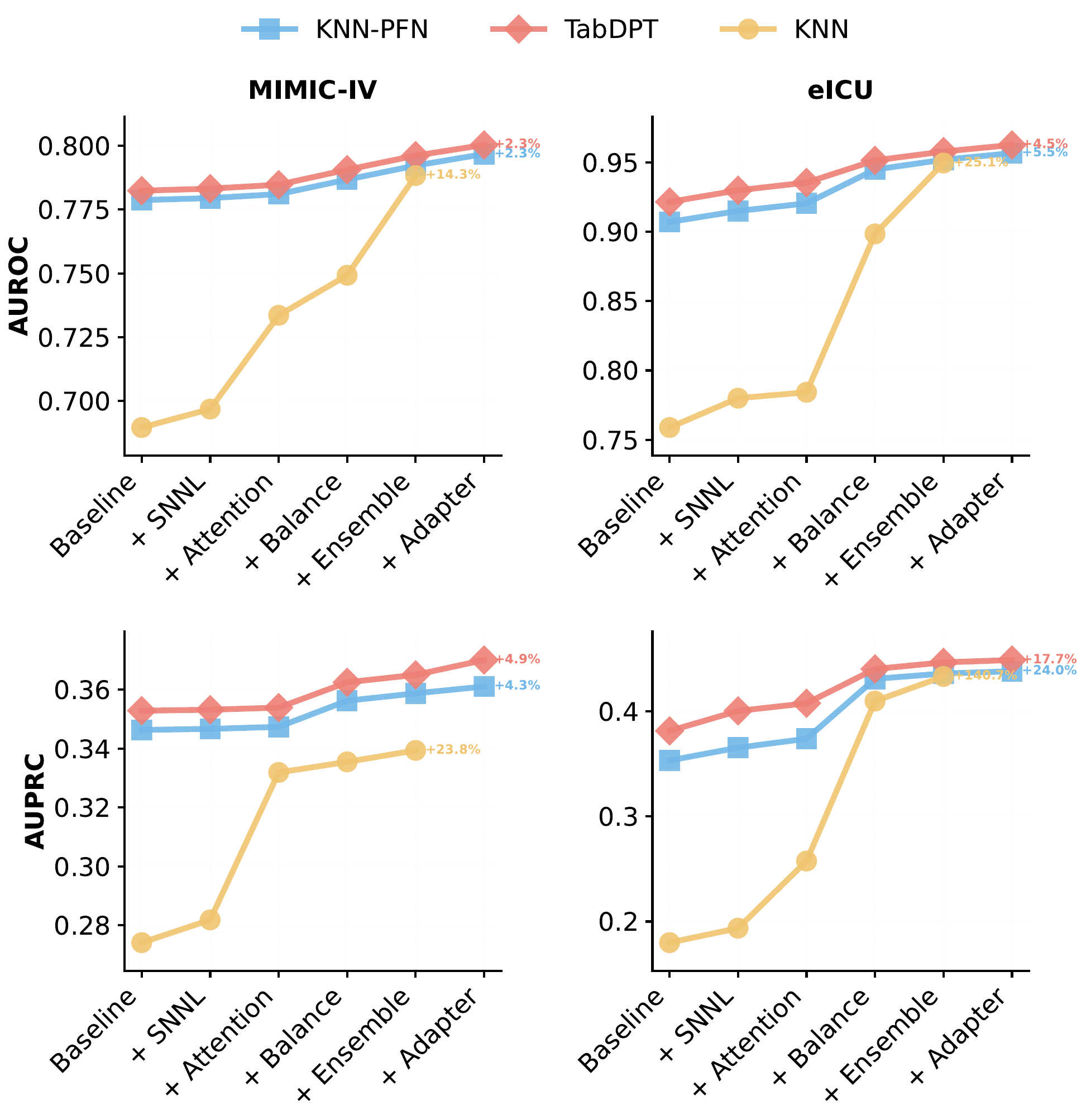}
    \caption{
\highlightfour{Mean performance improvement} over baseline as AWARE components are progressively introduced across three retrieval-based backbones on MIMIC-IV and eICU. Results are shown for AUROC (top) and AUPRC (bottom), averaged across \highlightfour{all clinical tasks and random seeds}.
}\label{fig:ablation_delta_pct}
\end{figure}

\highlightrev{Figure~\ref{fig:ablation_delta_pct} presents the resulting progressive ablation across three retrieval-based backbones (KNN, kNN-PFN, and TabDPT) and two large-scale ICU cohorts (MIMIC-IV and eICU). This cumulative design allows the performance change associated with each newly introduced component to be assessed conditional on the preceding configuration. Adapter tuning is applied only to kNN-PFN and TabDPT because these backbones contain trainable downstream parameters; for plain KNN, the progression ends with ensemble stabilization. The pretrained PFN backbones remain frozen throughout all configurations.}

Across architectures and datasets, task-aligned retrieval through SNNL initiates consistent performance gains, which generally increase with attention weighting and balanced sampling. The gains from balanced sampling are especially pronounced for AUPRC under class imbalance, supporting the interpretation that majority-class retrieval bias limits vanilla RA-TICL. For example, on eICU, the final AWARE-enhanced KNN configuration improves AUPRC by approximately 140\%, compared with a more moderate AUROC gain of 23.8\%. Improvements occur across all three backbones, suggesting that they arise from the AWARE components rather than increased model capacity. KNN exhibits the largest relative improvement, indicating that naive distance-based retrieval is particularly sensitive to poorly structured neighborhoods in heterogeneous EHR feature spaces. Cross-validation ensembling and adapter fine-tuning provide smaller, complementary gains. Overall, the results indicate that AWARE's principal contribution is task-aligned restructuring of retrieval neighborhoods, with balanced sampling further supporting rare-event conditioning and subsequent stabilization techniques providing secondary improvements.

\highlightrevdel{
Comprehensive ablation studies are provided in the Supplementary Information. We conduct comparison studies across embedding-based retrieval strategies in Supplementary Table \ref{tab:retrieval_objectives},  Supplementary Figure \ref{fig:retrieval_comparison_icl} and Supplementary Figure \ref{fig:retrieval_comparison_ret}. Furthermore, Supplementary Tables \ref{tab:ablation_aware_only}, \ref{tab:ablation_aware_wo}, and \ref{tab:ablation_aware_progressive} present isolated-component, leave-one-out, and progressive ablation analyses for these components, respectively.}

\subsection{Overall Results on Small-medium EHR datasets}
\label{sec:small_ehr_result}

\begin{figure}[t]
    \centering
    \includegraphics[width=\linewidth]{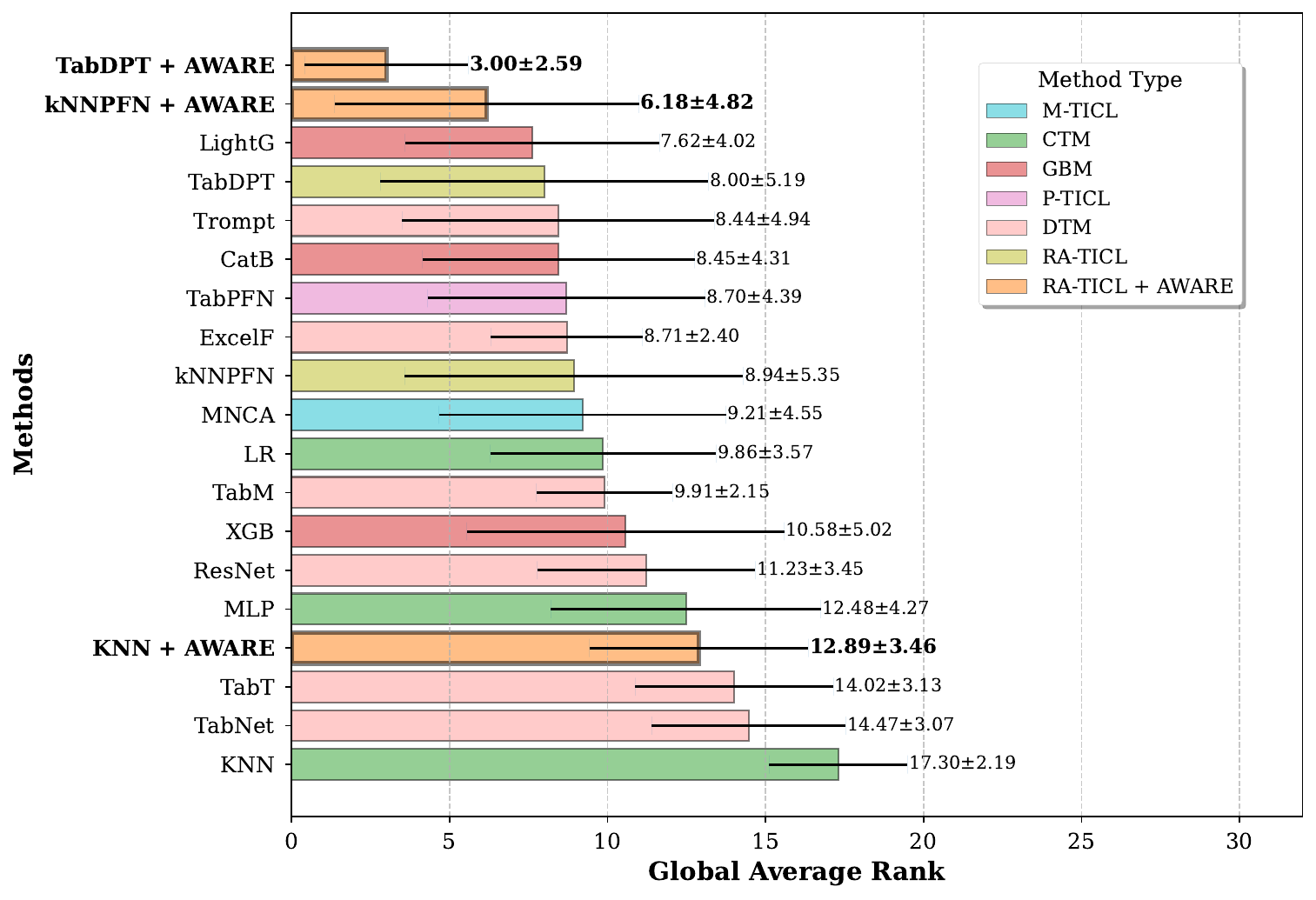}
    \caption{Aggregate performance rankings of tabular models on 12 small-medium EHR datasets, rankings are measured by average score of performance. Rankings summarize relative performance across tasks within each regime.}
    \label{fig:smallmediumehrs}
\end{figure}

Figure~\ref{fig:smallmediumehrs} summarizes model performance across 12 small- to medium-scale EHR datasets. In this regime, TICL methods (e.g., TabPFN) and retrieval-augmented variants (e.g., kNNPFN, TabDPT, ModernNCA) achieve strong average AUROC rankings without requiring task-specific fine-tuning or extensive hyperparameter search (with the exception of ModernNCA). These findings align with prior work demonstrating superior sample efficiency of tabular foundation models on general benchmarks~\cite{erickson2025tabarena}. In smaller datasets where labeled data are limited, inference-time conditioning appears sufficient to capture predictive structure without additional optimization.

\highlightfour{Incorporating AWARE yields substantial gains even in this small-to-medium regime.
TabDPT + AWARE achieves the best overall average rank ($3.0 \pm 2.59$), an improvement of approximately 4.5 rank positions over TabDPT alone ($8.00 \pm 5.19$), and kNNPFN + AWARE ($6.18 \pm 4.82$) similarly outperforms its base variant ($8.94 \pm 5.35$).
We attribute these gains to two key enhancements in the AWARE framework: embedding-based retrieval, which produces more semantically coherent neighbor sets than raw feature similarity, and imbalance-aware sampling during retrieval, which ensures that retrieved contexts are class-informative even when positive cases are sparse.}
While tree-based models remain competitive and exhibit stable performance, they are not uniformly superior in our experiments, contrary to some prior claims in the EHR literature. Given the heterogeneity of feature types across these datasets, the results suggest that \highlightfour{retrieval-augmented TICL methods, particularly when combined with AWARE,} provide a favorable balance of efficiency and effectiveness for small- to medium-scale clinical prediction tasks. \highlightappendix{Complete per-dataset performance metrics for all models on small- to medium-scale EHR datasets are provided in Supplementary Table \safeRef{tab:smallmediumehrs}{5}.}

\begin{figure*}[t]
    \setlength{\lineskip}{0pt}
    \centering
    \begin{tikzpicture}
        \node[anchor=north west,inner sep=0pt] at (0,0){\includegraphics[width=0.48\linewidth]{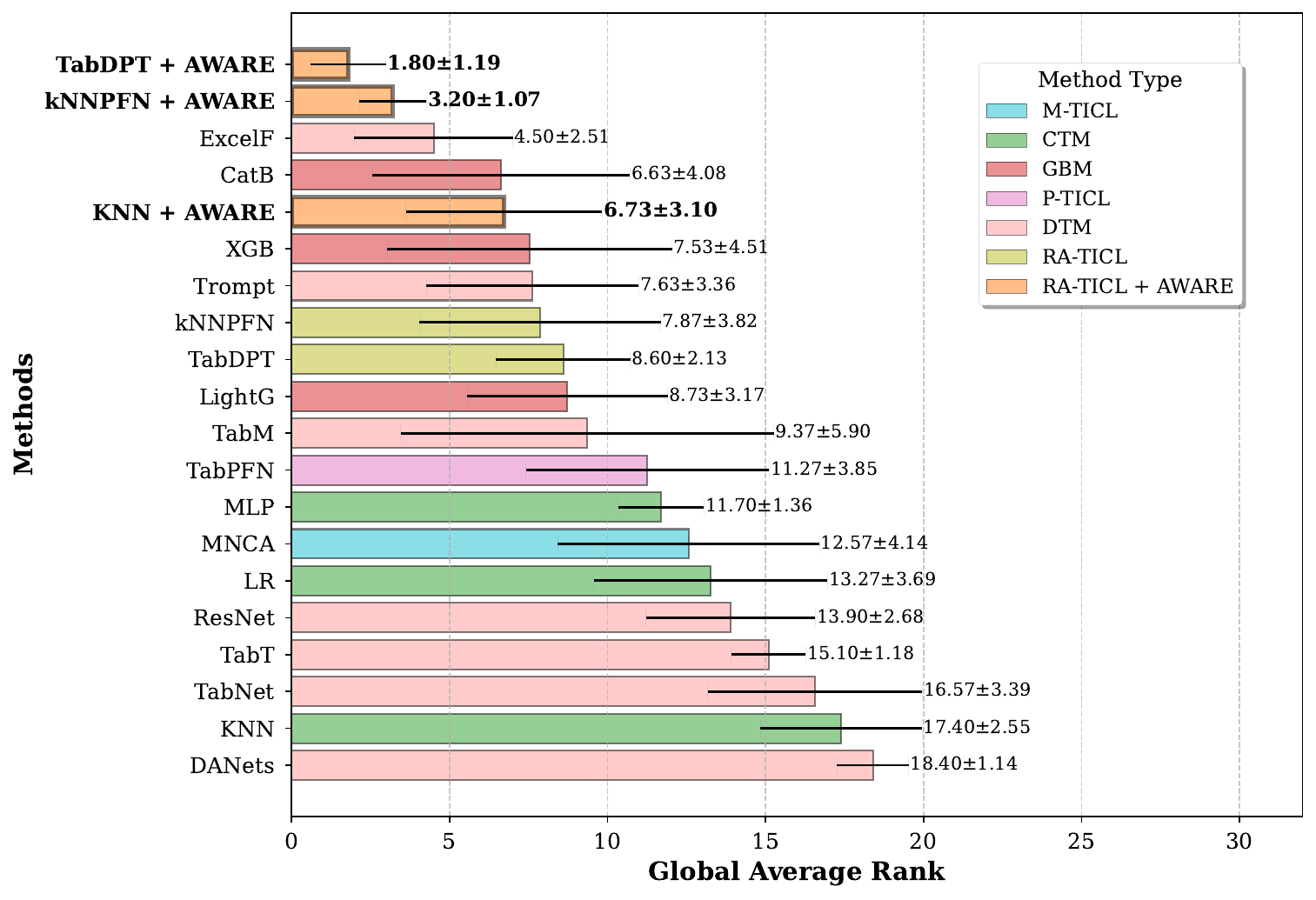}};
        \node[font=\sffamily\bfseries\large] at (0ex, 0ex) {a};
    \end{tikzpicture}
    \begin{tikzpicture}
        \node[anchor=north west,inner sep=0pt] at (0,0){\includegraphics[width=0.48\linewidth]{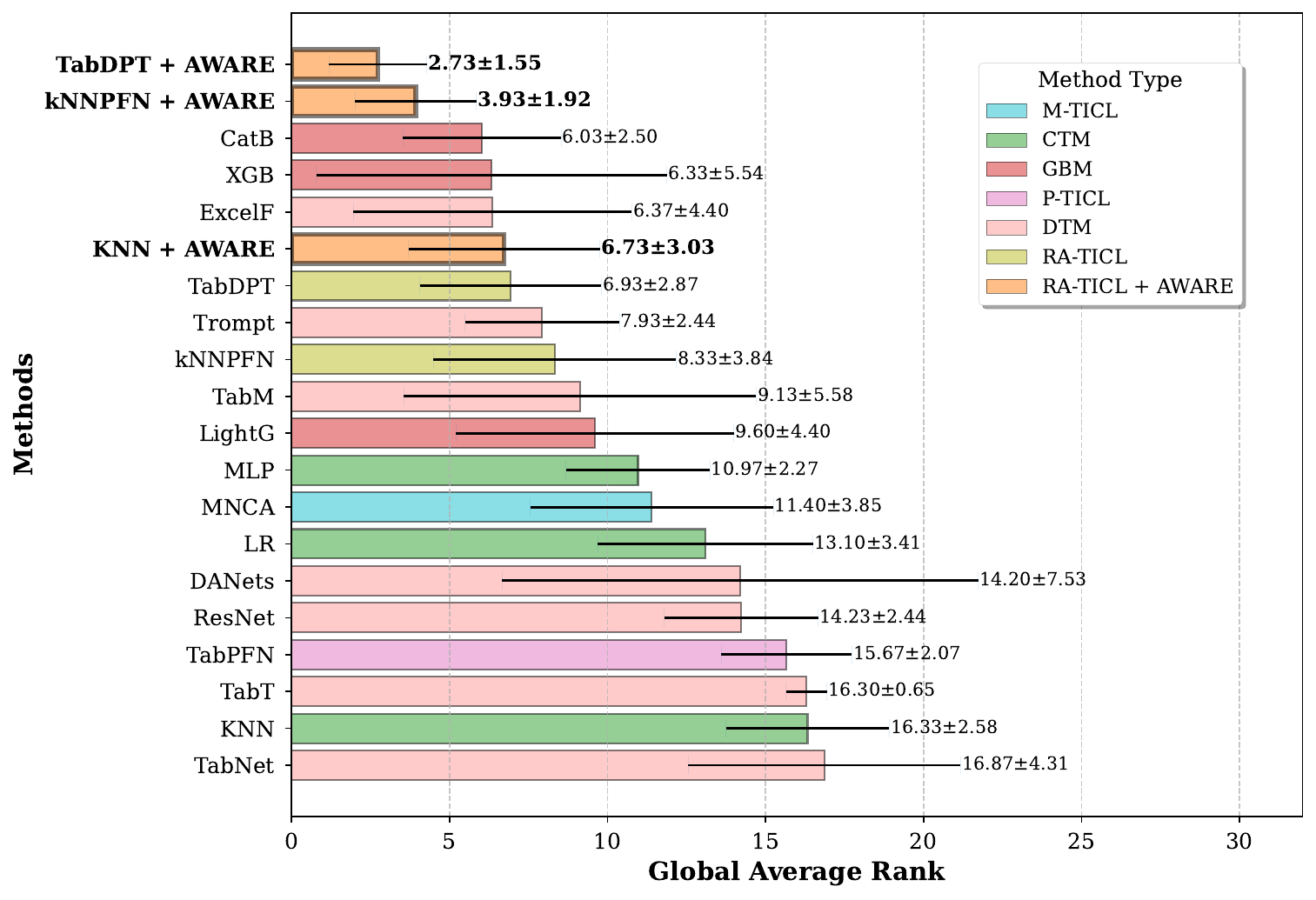}};
        \node[font=\sffamily\bfseries\large] at (0ex, 0ex) {b};
    \end{tikzpicture}
    \caption{Aggregate performance rankings of tabular models on large-scale EHR datasets (MIMIC-IV and eICU), measured by average (a) AUROC and (b) AUPRC across tasks within each regime.}
    \label{fig:largeehrs}
\end{figure*}

\subsection{Overall Results on Large-scale EHR datasets}
\label{sec:large_ehr_result}

As dataset scale and feature heterogeneity increase, the performance landscape shifts substantially. Figure~\ref{fig:largeehrs} shows that, when averaged across MIMIC-IV and eICU tasks, models with explicit feature selection mechanisms (e.g., ExcelFormer and Trompt) or tree-based architectures (e.g., CatBoost, LightGBM, and XGBoost) rise in the rankings relative to TICL and RA-TICL methods. Several deep tabular architectures, including TabTransformer, ResNet, and TabNet, rank below simpler baselines such as logistic regression, suggesting instability when modeling heterogeneous EHR features without strong inductive bias. In contrast, gradient-boosted tree methods remain consistently competitive, reinforcing their status as reliable baselines for complex clinical datasets. Notably, ExcelFormer and Trompt improve markedly at larger scales, likely benefiting from built-in feature filtering mechanisms that attenuate noise from weakly informative or sparse variables. \highlightappendix{Complete per-dataset performance metrics for all models on large-scale EHR datasets are provided in Supplementary Table~\safeRef{tab:hai_results}{6}.}

These results challenge the intuition that retrieval universally enhances in-context learning. Instead, they suggest that retrieval quality deteriorates in high-dimensional, sparse, and imbalanced EHR settings, where raw feature similarity becomes less aligned with predictive relevance. Consistent with this interpretation, ModernNCA exhibits a pronounced decline in ranking on large-scale cohorts, indicating increased sensitivity to neighborhood construction as data complexity increases. \highlightfour{Incorporating AWARE mitigates this degradation and yields substantial gains in this large-scale regime. TabDPT~+~AWARE achieves the best overall average rank for both AUROC ($1.80 \pm 1.19$) and AUPRC ($2.73 \pm 1.55$), improving by approximately 6.8 and 4.2 rank positions over TabDPT alone ($8.60 \pm 2.13$ and $6.93 \pm 2.87$, respectively). Similarly, kNNPFN~+~AWARE ranks second for both AUROC and AUPRC ($3.20 \pm 1.07$), improving by approximately 4.7 and 5.1 rank positions over kNNPFN alone ($7.87 \pm 3.82$ and $8.33 \pm 3.84$, respectively).} Together, these results suggest that task-aligned retrieval and imbalance-aware context construction become increasingly important as dataset scale, feature heterogeneity, and outcome imbalance increase.

\subsection{Performance Under Clinically Realistic EHR Constraints}
\label{sec:results_ehr_constraints}

We next investigate how core structural properties of EHR data shape model behavior through controlled analyses of training-set size, feature dimensionality, outcome rarity, and consistency across cohorts.

\begin{figure*}[t]
    \setlength{\lineskip}{0pt}
    \centering
    \begin{tikzpicture}
        \node[anchor=north west,inner sep=0pt] at (0,0){\includegraphics[width=0.304\linewidth,valign=t]{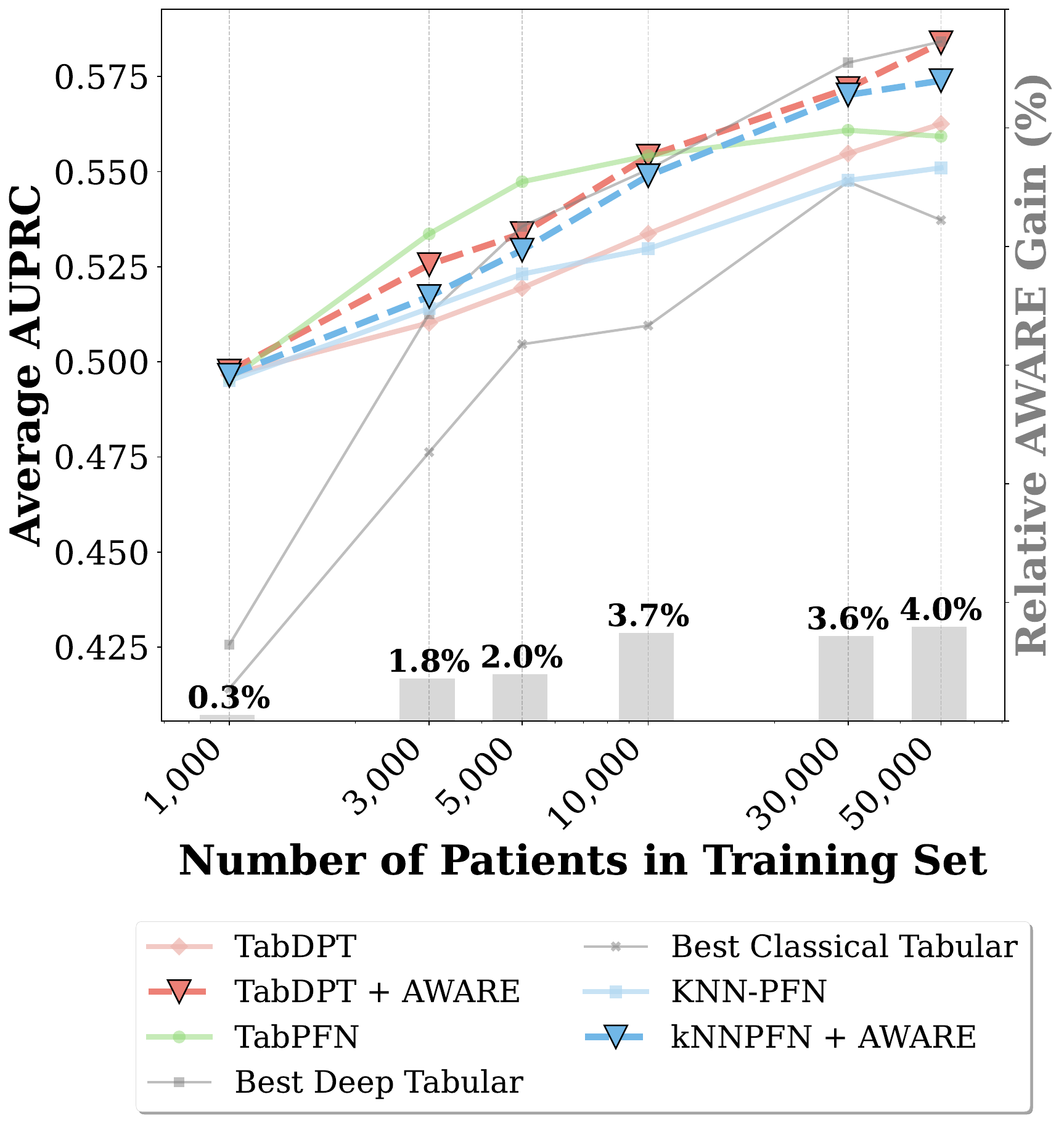}};
        \node[font=\sffamily\bfseries\large] at (0ex, -1ex) {a};
    \end{tikzpicture}
    \begin{tikzpicture}
        \node[anchor=north west,inner sep=0pt] at (0,0){\includegraphics[width=0.31\linewidth,valign=t]{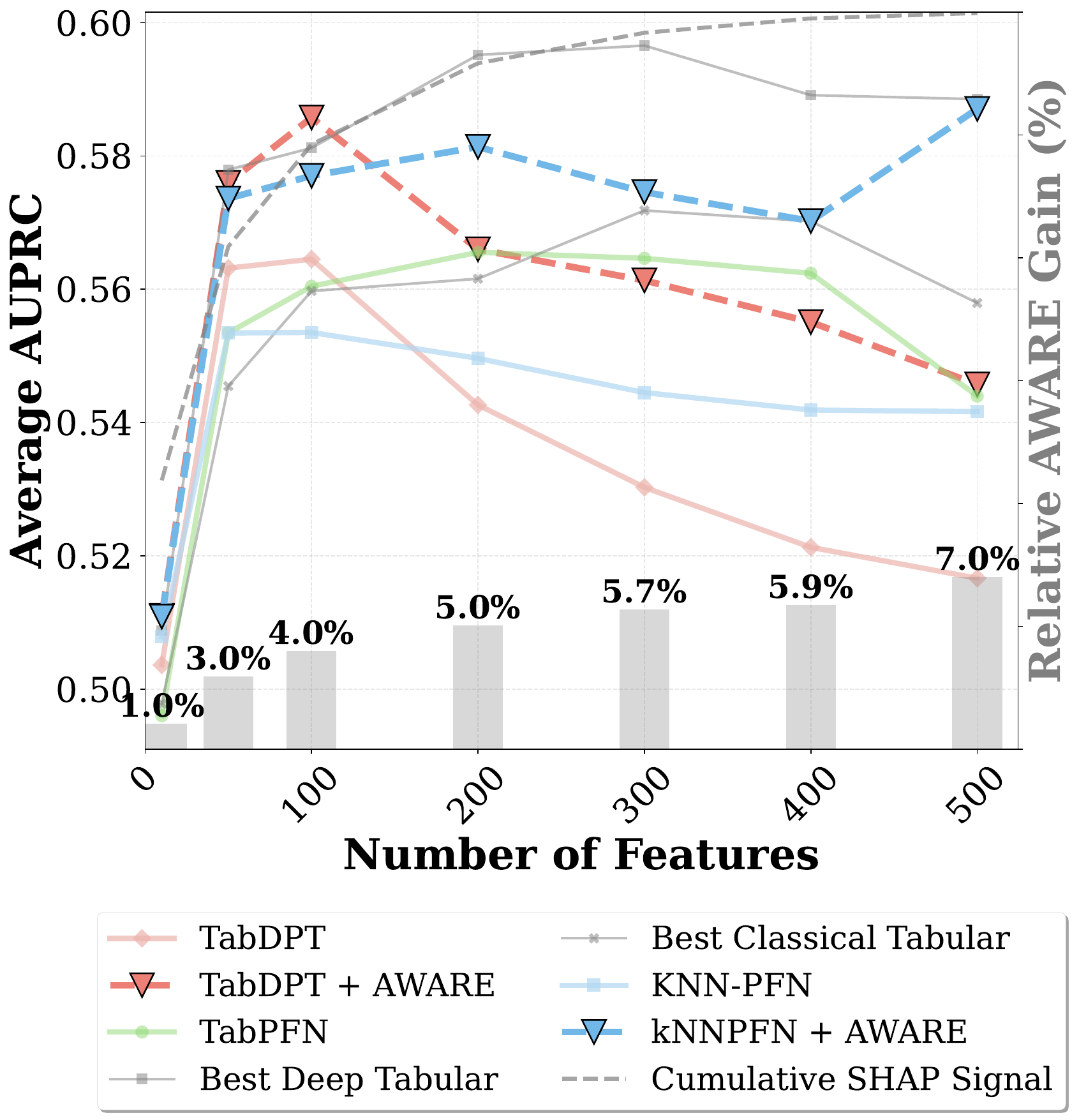}};
        \node[font=\sffamily\bfseries\large] at (0ex, 0ex) {b};
    \end{tikzpicture}
    \begin{tikzpicture}
        \node[anchor=north west,inner sep=0pt] at (0,0){\includegraphics[width=0.31\linewidth,valign=t]{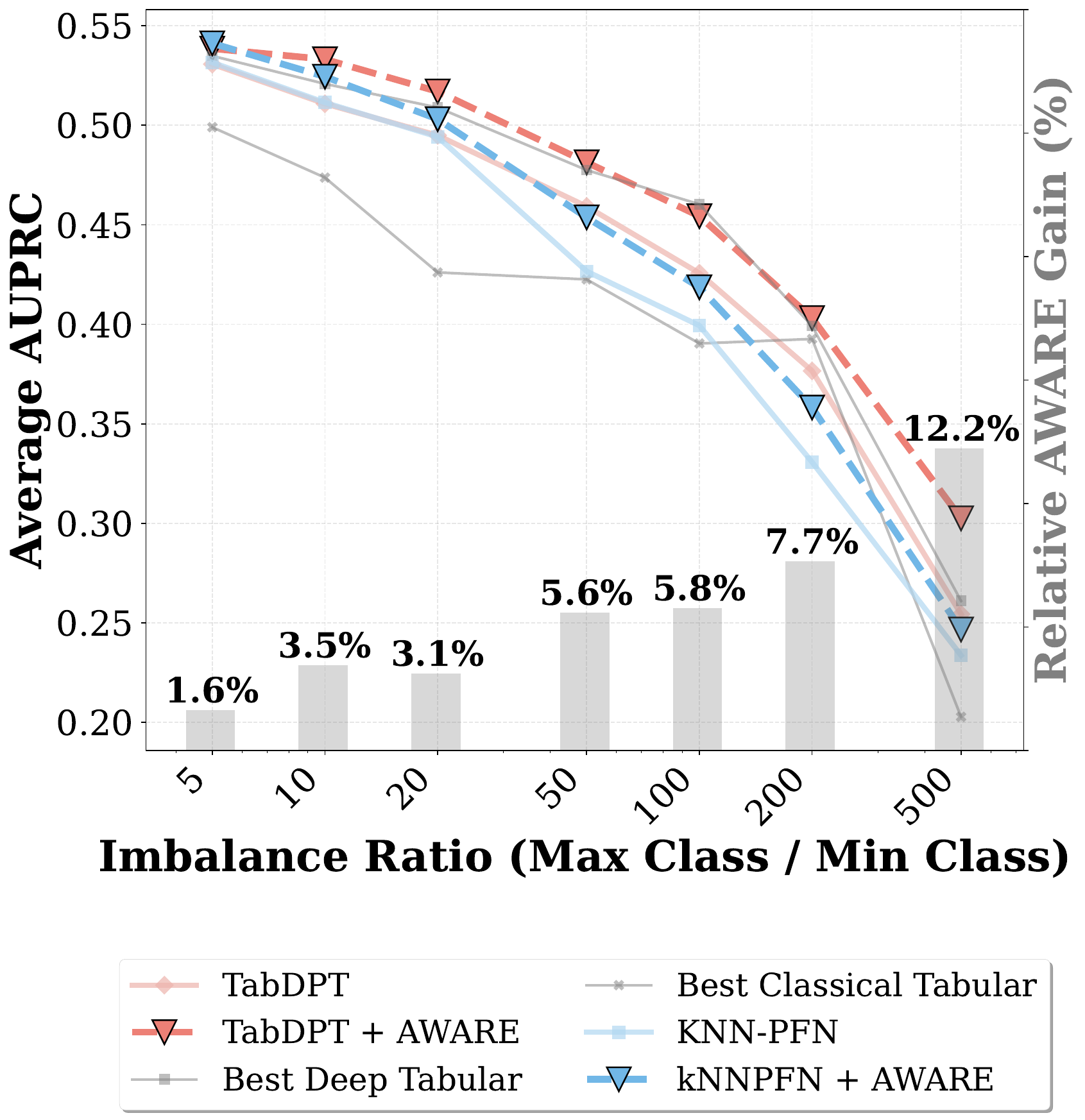}};
        \node[font=\sffamily\bfseries\large] at (0ex, 0ex) {c};
    \end{tikzpicture}
\caption{
Performance of tabular models under three representative clinical constraints. To improve readability given the large number of models evaluated, highly competitive baselines within each paradigm are consolidated into a single curve labeled \emph{Best Classical Model} and \emph{Best Deep Tabular Model}, representing the strongest-performing model in each family under the respective constraint. These are compared with individual tabular in-context learning methods and their AWARE-enhanced variants (dashed lines). Performance is measured using AUPRC across all settings, while overlaid bar charts report the average relative performance gains (in \%) obtained by incorporating AWARE. (a) Performance scaling with training data size. (b) Performance degradation under increasing feature heterogeneity. (c) Performance under increasing outcome imbalance.
}
\label{fig:comparison}
\end{figure*}

\paragraph{Data Scale Sensitivity.}
\label{sec:exp:data_scaling}

We first examine how model performance evolves as the amount of labeled training data increases, a central factor in \highlightrevdel{real-world clinical deployment}\highlightrev{clinical settings} where data availability varies substantially across institutions and tasks. Figure~\ref{fig:comparison}a reports AUPRC as a function of training-set size, ranging from 1{,}000 to 50{,}000 patients randomly sampled from both MIMIC-IV and eICU datasets. As expected, performance improves generally for all models as more training data become available. In the low-data regime around 1{,}000 patients, TICL and RA-TICL models achieve the strongest performance, consistent with their ability to leverage inductive priors and context-based prediction. However, performance gains for baseline TICL methods saturate earlier than for deep tabular models. Beyond approximately 10{,}000 patients, deep representation-learning approaches catch up with or outperform several standard TICL variants, suggesting that additional data are not exploited equally across model families.

Notably, incorporating the proposed AWARE framework into RA-TICL models (TabDPT and kNNPFN) alters this scaling behavior. As shown in Figure~\ref{fig:comparison}a, AWARE-enhanced variants maintain performance improvements as dataset size increases and remain competitive even at 50{,}000 patients. The overlaid bar charts further quantify these effects by reporting the relative performance gains achieved by AWARE at each sample size. These gains generally increase with data scale, from approximately 0.3\% in the smallest regime to around 3--4\% in larger datasets, suggesting that the observed benefit of AWARE becomes more pronounced as training-set size increases.

\paragraph{Feature Heterogeneity.}

We next examine sensitivity to feature heterogeneity by progressively expanding the input dimensionality from 10 to 500 features. Features are added in descending order of importance according to mean absolute SHapley Additive exPlanations (SHAP) values, such that early increments introduce high-signal variables while later increments increasingly incorporate lower-signal and potentially noisy dimensions. As illustrated by the cumulative SHAP signal curve in Figure~\ref{fig:comparison}b, informative signal saturates relatively early, and subsequent feature additions contribute progressively less additional signal. Under this controlled expansion, several baseline TICL models, most notably TabDPT, exhibit increasing sensitivity as dimensionality increases. In contrast, the best deep and classical tabular baselines remain comparatively stable from 200 features and beyond, suggesting greater robustness to feature noise through learned representations or implicit feature selection. These results suggest that the limitations of TICL methods in high-dimensional EHR settings may be related not merely to dataset size, but also to the dilution of task-relevant signal within heterogeneous feature spaces.

Incorporating AWARE consistently mitigates this degradation. As shown in Figure~\ref{fig:comparison}b, AWARE-enhanced RA-TICL variants consistently maintain higher AUPRC across the full dimensionality range and exhibit reduced performance drop in the high-feature regime. The overlaid bar charts quantify these improvements as relative percentage gains over the corresponding baseline RA-TICL models. Notably, the gains increase monotonically with dimensionality—from approximately 1--4\% at low feature counts to nearly 7\% at 500 features—suggesting that AWARE becomes increasingly beneficial as heterogeneity intensifies. 

\highlightrevdel{Taken together, these findings suggest that in heterogeneous EHR feature spaces, limitations of RA-TICL models arise not only from representational capacity but from context construction in noisy, high-dimensional inputs. By introducing task-aligned retrieval and implicit feature reweighting, AWARE helps adapt RA-TICL models to EHR complexity and mitigates heterogeneity-induced instability.}

\paragraph{Outcome Rarity.}
Rare outcomes are pervasive in clinical prediction tasks, particularly for adverse events such as hospital-acquired infections, in-hospital mortality, and treatment complications~\cite{cohen2006learning}. To systematically evaluate robustness to outcome rarity, we adopt a controlled imbalance-scaling protocol. The training-set size is fixed at 10{,}000 samples, while the negative-to-positive ratio (imbalance ratio, IR) is progressively increased from 5 to 500 through stratified subsampling without replacement. All models are evaluated on a fixed held-out test set with the original class distribution~\cite{zhu2026bridging}. Figure~\ref{fig:comparison}c illustrates how AUPRC degrades as class imbalance intensifies. All methods generally exhibit performance decline as positive prevalence decreases. However, the rate and stability of this degradation differ substantially across paradigms. The best deep tabular model maintains comparatively strong absolute performance across moderate imbalance levels, while the best classical model degrades more sharply, particularly beyond IR~=~50. Baseline RA-TICL models (TabDPT and kNNPFN) demonstrate competitive robustness under mild-to-moderate imbalance but experience increasingly pronounced degradation in the extreme regime (IR~$\geq$~200), reflecting the difficulty of constructing representative contextual examples when minority instances become scarce.

Incorporating AWARE consistently mitigates this effect. As shown in Figure~\ref{fig:comparison}c, AWARE-enhanced variants consistently remain above their baseline counterparts across all imbalance levels, with the performance gap \highlightrev{generally} widening as rarity increases. The overlaid bar charts quantify these improvements as relative percentage gains, which grow from approximately 1.6\% at IR~=~5 to over 12\% at IR~=~500. This \highlightrev{overall} increase suggests that AWARE becomes progressively more beneficial as minority-class sparsity intensifies.

\highlightrevdel{These results suggest that under extreme prevalence skew, limitations of baseline RA-TICL models stem not only from representational capacity but from instability in retrieval and context construction when positive examples are scarce. By introducing task-aligned retrieval and improved minority coverage, AWARE enhances robustness in rare-event regimes while preserving the retraining-free nature of inference-time adaptation.}

\paragraph{\highlightrevdel{Generalizability.}\highlightrev{Multi-Cohort Consistency.}}

\begin{table}[t]
\centering
\caption{\highlighthipe{Performance on HIPE prediction tasks. Results are mostly reused from Pham et al. \cite{pham2025explainable}. With our new inclusion of \highlightfour{KNN, kNNPFN and TabDPT baselines and AWARE versions.}} \highlightrev{Markers ($^{**}$p$<$0.01) on AWARE models indicate they significantly outperform their vanilla (non-AWARE) counterparts (paired bootstrap, Holm-Bonferroni).}}
\label{tab:results:hipe_ehrs}
\resizebox{\textwidth}{!}{
\renewcommand{\arraystretch}{0.95}
\begin{tabular}{lcccc}
\toprule
\multirow{3}{*}{\textbf{Method}} & \multicolumn{4}{c}{\textbf{HIPE}} \\
\cmidrule(lr){2-5}
 & \multicolumn{2}{c}{RDMT}  & \multicolumn{2}{c}{CPE} \\
\cmidrule(lr){2-3} \cmidrule(lr){4-5} 
 & AUROC $\uparrow$ & AUPRC $\uparrow$ & AUROC $\uparrow$ & AUPRC $\uparrow$ \\
\midrule

XGB & \ensuremath{0.824_{\textcolor{gray}{\pm 0.010}}} & \ensuremath{0.432_{\textcolor{gray}{\pm 0.020}}} & \ensuremath{0.755_{\textcolor{gray}{\pm 0.043}}} & \ensuremath{0.036_{\textcolor{gray}{\pm 0.019}}} \\
CatB & \ensuremath{0.827_{\textcolor{gray}{\pm 0.005}}} & \ensuremath{0.446_{\textcolor{gray}{\pm 0.015}}} & \ensuremath{0.645_{\textcolor{gray}{\pm 0.078}}} & \ensuremath{0.027_{\textcolor{gray}{\pm 0.025}}} \\
LightG & \ensuremath{0.814_{\textcolor{gray}{\pm 0.055}}} & \ensuremath{0.438_{\textcolor{gray}{\pm 0.065}}} & \ensuremath{0.719_{\textcolor{gray}{\pm 0.092}}} & \ensuremath{0.030_{\textcolor{gray}{\pm 0.021}}} \\

TabNet & \ensuremath{0.827_{\textcolor{gray}{\pm 0.005}}} & \ensuremath{0.440_{\textcolor{gray}{\pm 0.009}}} & \ensuremath{0.558_{\textcolor{gray}{\pm 0.083}}} & \ensuremath{{0.111}_{\textcolor{gray}{\pm 0.055}}} \\
ResNet & \ensuremath{{0.830}_{\textcolor{gray}{\pm 0.001}}} & \ensuremath{{0.457}_{\textcolor{gray}{\pm 0.006}}} & \ensuremath{{0.773}_{\textcolor{gray}{\pm 0.030}}} & \ensuremath{0.046_{\textcolor{gray}{\pm 0.030}}} \\
TabT & \ensuremath{\underline{0.834}_{\textcolor{gray}{\pm 0.004}}} & \ensuremath{{0.461}_{\textcolor{gray}{\pm 0.007}}} & \ensuremath{\underline{0.781}_{\textcolor{gray}{\pm 0.057}}} & \ensuremath{0.080_{\textcolor{gray}{\pm 0.017}}} \\
TabPFN & \ensuremath{0.828_{\textcolor{gray}{\pm 0.002}}} & \ensuremath{0.455_{\textcolor{gray}{\pm 0.004}}} & \ensuremath{0.719_{\textcolor{gray}{\pm 0.051}}} & \ensuremath{0.022_{\textcolor{gray}{\pm 0.010}}} \\

\midrule
KNN & \ensuremath{0.681_{\textcolor{gray}{\pm 0.001}}} & \ensuremath{0.307_{\textcolor{gray}{\pm 0.001}}} & \ensuremath{0.733_{\textcolor{gray}{\pm 0.001}}} & \ensuremath{0.020_{\textcolor{gray}{\pm 0.001}}} \\

kNNPFN & \ensuremath{0.706_{\textcolor{gray}{\pm 0.001}}} & \ensuremath{0.365_{\textcolor{gray}{\pm 0.001}}} & \ensuremath{0.760_{\textcolor{gray}{\pm 0.001}}} & \ensuremath{{0.141}_{\textcolor{gray}{\pm 0.001}}} \\
TabDPT & \ensuremath{0.709_{\textcolor{gray}{\pm 0.001}}} & \ensuremath{0.362_{\textcolor{gray}{\pm 0.001}}} & \ensuremath{0.746_{\textcolor{gray}{\pm 0.001}}} & \ensuremath{0.017_{\textcolor{gray}{\pm 0.001}}} \\

\midrule
\multicolumn{5}{c}{\textbf{\textit{Ours (Tabular In-Context Learning Models + AWARE)}}} \\
\midrule
KNN & \ensuremath{0.720_{\textcolor{gray}{\pm 0.001}}} & \ensuremath{0.349_{\textcolor{gray}{\pm 0.001}}} & \ensuremath{0.774_{\textcolor{gray}{\pm 0.001}}} & \ensuremath{0.072_{\textcolor{gray}{\pm 0.001}}} \\
kNNPFN & \ensuremath{\highlightrev{\textbf{0.843}^{**}}_{\textcolor{gray}{\pm 0.005}}} & \ensuremath{\highlightrev{\textbf{0.480}^{**}}_{\textcolor{gray}{\pm 0.006}}} & \ensuremath{\highlightrev{\textbf{0.782}^{**}}_{\textcolor{gray}{\pm 0.024}}} & \ensuremath{\highlightrev{\textbf{0.232}^{**}}_{\textcolor{gray}{\pm 0.020}}} \\
TabDPT & \ensuremath{\highlightrev{\underline{0.834}^{**}}_{\textcolor{gray}{\pm 0.005}}} & \ensuremath{\highlightrev{\underline{0.465}^{**}}_{\textcolor{gray}{\pm 0.002}}} & \ensuremath{\highlightrev{\underline{0.781}^{**}}_{\textcolor{gray}{\pm 0.017}}} & \ensuremath{\highlightrev{\underline{0.221}^{**}}_{\textcolor{gray}{\pm 0.002}}} \\

\midrule
\bottomrule
\end{tabular}}
\end{table}

\highlightrev{Taken together, results in \S\ref{sec:small_ehr_result} and \S\ref{sec:large_ehr_result} show that TICL and RA-TICL models remain competitive with classical and deep tabular baselines across small-to-medium and large-scale cohorts alike, across diverse medical domains, feature types, and institutional practices, without dataset-specific hyperparameter search. We next extend this multi-cohort view to HIPE, the most heterogeneous and imbalanced cohort in our benchmark.}

\highlighthipe{This privately held HIPE cohort has been originally analyzed in prior work~\cite{pham_forecasting_2024,pham2025explainable}, by incorporating RA-TICL variants into the same experimental framework (Table~\ref{tab:results:hipe_ehrs}). HIPE represents an extreme setting characterized by substantial feature heterogeneity, high-dimensional sparse coding, and severe outcome imbalance. In this environment, baseline retrieval-augmented TICL models (kNNPFN and TabDPT) exhibit noticeable performance degradation relative to the strongest task-specific deep models, reflecting the compounded challenges of sparsity, noisy similarity structure, and rare-event prevalence in raw EHR feature space. }

\highlighthipe{However, when combined with the proposed AWARE retrieval alignment, performance improves substantially across both tasks and metrics. The AWARE variants consistently outperform their vanilla counterparts and achieve the best overall performance in most settings, with particularly pronounced gains in AUPRC for the highly imbalanced CPE task (e.g., from 0.141 to 0.232 for kNNPFN). Notably, these improvements allow AWARE-based TICL models to match or surpass the strongest deep tabular baselines such as TabTransformer and ResNet\highlightrevdel{, despite requiring no task-specific retraining.} \highlightrev{, while requiring substantially less training time (as shown in Supplementary Figure \safeRef{fig:apdx:efficiency}{4}).} \highlightrevdel{These results suggest that much of the degradation observed in naive retrieval stems from misaligned similarity structure in heterogeneous EHR spaces, and that task-aware retrieval alignment can substantially improve the effectiveness of in-context conditioning under severe imbalance.}}

\highlightrevdel{These findings suggest that inference-time tabular conditioning remains viable even under substantial structural and statistical shift. While extreme imbalance and heterogeneity can attenuate performance, degradation is gradual rather than catastrophic. Importantly, TICL methods achieve this level of cross-cohort robustness without retraining, hand-crafted feature pipelines, or domain-specific engineering.}

\highlightrev{Taken together, the controlled stress tests and multi-cohort results demonstrate the robustness of AWARE under clinically realistic EHR constraints. Its benefits become increasingly pronounced as dataset scale, feature dimensionality, and outcome rarity increase, and persist in the heterogeneous and severely imbalanced HIPE cohort. By replacing task-agnostic distance-based retrieval with task-aligned context construction, AWARE consistently reduces the performance degradation of RA-TICL models across diverse clinical tasks, feature regimes, and cohorts. These findings support our central conclusion that retrieval-aligned tabular foundation models enable more robust clinical risk prediction from EHRs under clinically realistic constraints, particularly in settings where naive neighborhood construction is least reliable.}

\section{Discussion}
\label{sec:discussion}

\highlightrev{This study provides a systematic multi-cohort assessment of TICL and RA-TICL for EHR-based clinical prediction. Three principal findings emerge. First, TICL models demonstrate strong sample efficiency on small- and medium-scale datasets, where pretrained tabular priors support competitive prediction with limited labeled data. Second, these advantages do not automatically extend to larger and more complex cohorts, where feature heterogeneity, sparsity, outcome imbalance, and unstable context composition increasingly limit vanilla TICL and RA-TICL. Third, AWARE consistently mitigates these limitations, with its relative benefits generally increasing as the clinical constraints intensify. Thus, transferring the success of foundation models from generic tabular benchmarks to clinical data is neither straightforward nor directly beneficial~\cite{bohr-etal-2025-perspective}; retrieval alignment and context construction, rather than model capacity alone, are important determinants of RA-TICL performance in EHR data.}

\highlightrev{The comparative results reveal a clear regime-dependent pattern. On small- and medium-scale datasets, TICL and RA-TICL remain competitive with classical and deep tabular approaches, consistent with the reported sample efficiency of tabular foundation models~\cite{erickson2025tabarena}. Incorporating AWARE further improves both TabDPT and kNNPFN, indicating that task-aligned retrieval is beneficial even in relatively compact cohorts. As dataset scale and feature heterogeneity increase, however, tree-based and feature-selective deep models become more competitive, whereas several vanilla neighborhood-based methods decline in ranking. AWARE mitigates much of this degradation and moves its RA-TICL backbones to the strongest overall average rankings in the large-scale ICU experiments. Together with evidence that count-based models can remain competitive with sequential transformers and LLM-based pipelines on structured EHR tasks~\cite{gao2025countbased}, these findings suggest that no modeling paradigm is uniformly superior: performance depends on the clinical data regime and, for retrieval-based models, on the quality of context construction.}

\highlightrev{The ablation and stress tests identify retrieval alignment as a key determinant of performance in complex EHR settings. Raw distance can be misleading because predictive information is often concentrated in task-specific variables, while sparse or weakly informative features may dominate patient similarity~\cite{scheurwegs2017selecting,qin2021retrieval,zheng2023dense}. AWARE addresses this mismatch by combining task-aligned embeddings, attention-based feature weighting, and imbalance-aware sampling. Its consistent gains across KNN, kNNPFN, and TabDPT suggest that the improvements arise from better context construction rather than increased PFN capacity alone. Moreover, relative AUPRC gains increase from approximately 0.3\% to 3--4\% with dataset size, approach 7\% at 500 features, and exceed 12\% under extreme imbalance. Thus, as candidate pools become larger and more heterogeneous, selecting relevant and class-informative examples becomes increasingly important~\cite{thomas_retrieval_nodate,xu_mixture_2024}.}

\highlightrev{This interpretation is supported by the consistent improvements observed across the evaluated cohorts. AWARE improves both RA-TICL backbones across smaller clinical datasets and the public ICU cohorts, despite differences in clinical domains, feature representations, and outcome prevalence. Its benefit is especially evident in HIPE~\cite{pham_forecasting_2024,pham2025explainable}, which combines sparse high-dimensional features with severe imbalance: for the CPE task, AWARE increases kNNPFN AUPRC from 0.141 to 0.232. These findings position AWARE as a parameter-efficient form of task adaptation that keeps the PFN backbone frozen, although its retrieval encoder and optional adapter remain cohort- and task-specific. More broadly, TICL may be most useful when labeled data are limited, whereas tree-based, feature-selective, or sequential models may remain preferable for highly heterogeneous or strongly temporal tasks. Comparative computational efficiency is reported in Supplementary Figure~\safeRef{fig:apdx:efficiency}{4}.}

\highlightrev{Several limitations qualify these findings and motivate future work. The evaluation focuses on structured, often temporally aggregated EHR features and may therefore obscure fine-grained dynamics and sequential dependencies~\cite{choi_retain_2016,lipton2015learning}; the conclusions may not extend to end-to-end longitudinal, text-based, imaging, or multimodal models. Missing-data preprocessing is applied consistently, but informative missingness mechanisms are not explicitly modeled. Furthermore, all datasets are retrospective and originate from a limited number of healthcare systems, leaving possible effects from unobserved confounding, institutional practices, and temporal change. External-site, temporal-shift, cross-institutional transfer, and prospective evaluations are therefore needed.  Finally, the present evaluation focuses primarily on discrimination. Deployment-oriented studies should additionally examine calibration, uncertainty, subgroup performance, threshold-specific utility, computational latency, and decision-curve outcomes. Because we do not evaluate clinician interaction, workflow integration, or prospective decision-making, the present study should be interpreted as methodological benchmarking rather than evidence of clinical utility or deployment readiness.}

Taken together, this work positions tabular foundation models as complementary rather than general-purpose solutions for EHR modeling. TICL provides strong sample efficiency in favorable regimes but is not inherently robust to feature heterogeneity, retrieval mismatch, outcome rarity, or cohort variation. AWARE demonstrates that task-aligned and stabilized context construction can substantially mitigate these limitations without fine-tuning the pretrained PFN backbone. Within the evaluated settings, retrieval-aligned tabular foundation models therefore provide a more robust basis for clinical risk prediction under real-world EHR constraints, while requiring further prospective and cross-institutional validation before clinical deployment.

\section{Methods}
\label{sec:methods}

\highlightrev{This study comprises two complementary components: (1) a systematic evaluation of tabular learning methods for EHR-based clinical prediction and (2) the development of AWARE, our task-aligned retrieval framework for retrieval-augmented tabular in-context learning. The evaluation examines model performance under four clinically relevant characteristics of real-world EHR data: data scale, feature heterogeneity, outcome rarity, and performance consistency across datasets and healthcare systems.}

\subsection{AWARE Framework Overview}

\begin{figure}[ht]
    \setlength{\lineskip}{0pt}
    \centering
    \includegraphics[width=0.8\linewidth]{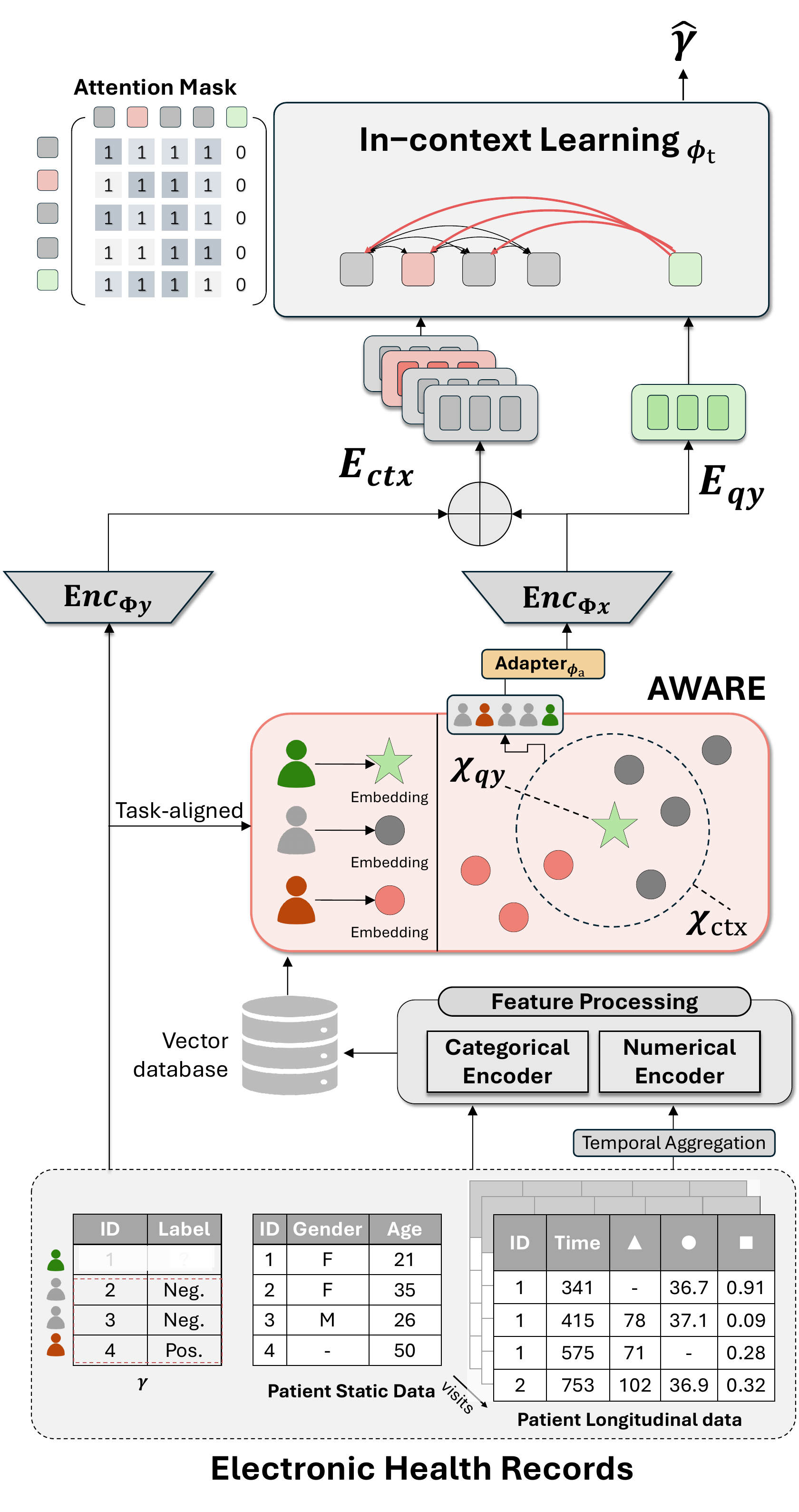}
    
    \caption{
    Overview of our designed task-aligned retrieval (AWARE) for retrieval-augmented in-context learning EHR model. 
    }
    \label{fig:diagram}
\end{figure}

\highlightrev{Existing RA-TICL methods commonly construct context sets using fixed similarity metrics, such as Euclidean or cosine distance, in the original feature space~\cite{thomas_retrieval_nodate}. This assumption is restrictive for EHR data because patient representations are high-dimensional and heterogeneous, and the same feature space may support multiple prediction tasks requiring different notions of similarity. Consequently, examples that are close in the raw feature space may not be relevant to the outcome being predicted and may provide label-misaligned context.}

To address this mismatch, we propose AWARE (Attention Weighting for Aligned Retrieval Embeddings), which comprises two sequential stages, as illustrated in Figure~\ref{fig:diagram}. First, a supervised attention-based encoder is trained using the Soft Nearest Neighbor Loss (SNNL) to produce task-aligned embeddings used exclusively for retrieval. This stage reshapes local neighborhood structure so that proximity more closely reflects outcome relevance. Second, for RA-TICL backbones that support trainable downstream components, an optional lightweight adapter is optimized using bootstrapped prompts constructed from the retrieved neighborhoods, while the pretrained backbone remains frozen. This adapter stage is used with kNNPFN and TabDPT; KNN+AWARE consists only of the retrieval-alignment stage.

\highlightrevdelraw{
\begin{figure*}[htp]
    \setlength{\lineskip}{0pt}
    \centering
    \begin{tikzpicture}
        \node[anchor=north west,inner sep=0pt] (img1) at (0,0){\includegraphics[width=0.91\linewidth]{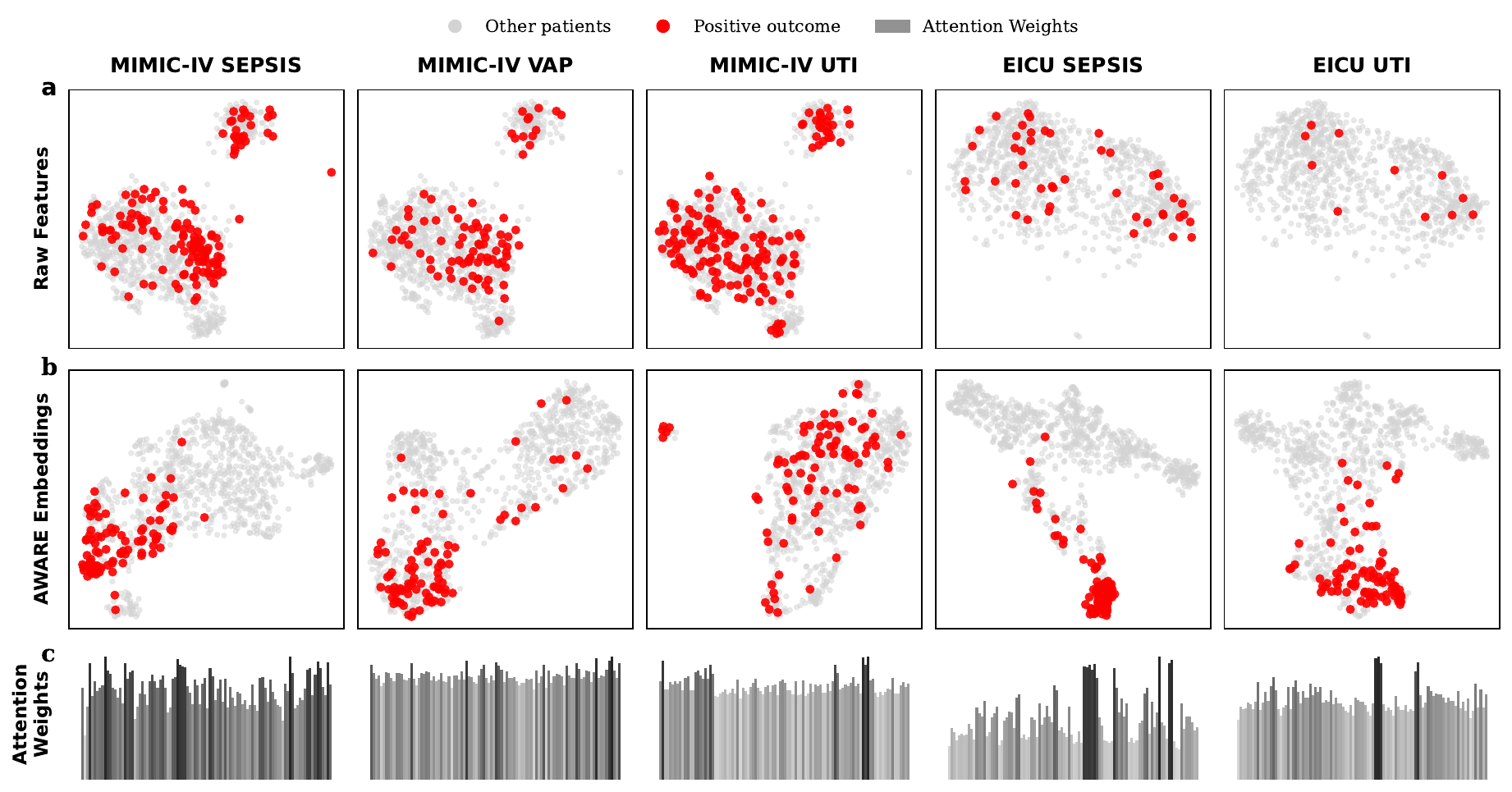}};
        \node[font=\sffamily\bfseries\large] at (0ex, -2ex) {a};
        
        \draw[red, ultra thick] (img1.south west) -- (img1.north east);
        \draw[red, ultra thick] (img1.south east) -- (img1.north west);
    \end{tikzpicture}
\end{figure*}
}

\highlightrev{At inference, the query patient is embedded by the trained retrieval encoder, and its nearest neighbors are selected exclusively from the training support pool. The retrieved examples are combined with the query to form the in-context prompt. Where applicable, the prompt is transformed by the trained adapter and then passed to the frozen TICL backbone to obtain the final prediction. No encoder, adapter, or backbone parameters are updated during inference. AWARE therefore combines task-aligned neighborhood construction with optional representation alignment while preserving the pretrained PFN backbone.}

\subsection{Task-Aligned Retrieval Embedding and Context Construction.}
AWARE employs a lightweight neural encoder that combines feature-wise attention with embedding-based metric learning. Given a patient feature vector $x \in \mathbb{R}^d$, the encoder computes instance-specific attention weights
\begin{equation}
\alpha(x) = \sigma\!\left(W_2 \,\phi(W_1 x)\right) \in (0,1)^d,
\end{equation}
where $\phi(\cdot)$ is a nonlinear activation and $\sigma(\cdot)$ denotes the sigmoid function. These weights are applied via element-wise scaling,
\begin{equation}
\tilde{x} = x \odot \alpha(x)
\end{equation}
where a $\odot$ b denotes element-wise multiplication. This operation provides soft, instance-specific feature reweighting by emphasizing features associated with the prediction objective while down-weighting less informative or noisy features. From the reweighted representation $\tilde{x}$, the encoder produces a low-dimensional embedding:
\begin{equation}
z = f_\theta(\tilde{x}),
\end{equation}
which is used exclusively for retrieval. 

To directly align neighborhood structure with clinical outcomes, the encoder \highlightfour{is} trained using the Soft Nearest Neighbor Loss \cite{frosst_2019_snnl}. Given a batch of $b$ embeddings $\{z_i\}_{i=1}^b$ with labels $\{y_i\}_{i=1}^b$, the loss is defined as
\begin{equation}
\label{eq:snnl}
    \ell_{\mathrm{sn}}(z,y,T) = -\frac{1}{b} \sum_{i=1}^{b} \log \frac{ \sum_{j \neq i,\, y_j = y_i} \exp\!\left(-\frac{d(z_i,z_j)}{T}\right) }{ \sum_{k \neq i} \exp\!\left(-\frac{d(z_i,z_k)}{T}\right) },
\end{equation}
where $d(\cdot,\cdot)$ denotes a distance metric (e.g, cosine distance) and $T$ is a temperature parameter. Minimizing Equation ~\ref{eq:snnl} encourages embeddings from the same outcome class to form compact local neighborhoods while separating embeddings from different classes, yielding task-aligned similarity for retrieval in heterogeneous feature spaces. For regression tasks, \highlightrev{which is not supported natively by SNNL}, we instead construct pseudo-label groups using quantile-based target binning $\hat{y}={b(y)}$ \highlightrev{as follows:}
\begin{equation}
    \widehat{y}_i = b_(y_i) = \sum_{r=1}^{Q-1} \mathbb{I}\!\left(y_i > \tau_r\right), \qquad \widehat{y}_i \in \{0,\ldots,Q-1\},
\label{eq:regression_pseudolabel}
\end{equation}
\highlightrev{where \(\mathbb{I}(\cdot)\) denotes the indicator function. The function \(b_(y_i)\) assigns \(y_i\) to one of \(Q\) zero-indexed quantile bins by counting the number of training-derived quantile boundaries below \(y_i\).}

\highlightsecond{
\subsection{Adapter-Based Representation Alignment.}
While SNNL-based embedding learning improves neighborhood quality during retrieval, the retrieved context must still be mapped into a representation space compatible with the frozen in-context backbone. To address this, we introduce a parameter-efficient adapter $A_\psi$ that operates prior to the backbone model. Rather than modifying internal Transformer layers, the adapter learns a task-specific projection that maps both support and query representations into a feature space aligned with the backbone’s pretrained representation.}

\highlightsecond{
Inspired by Xu et al.~\cite{xu_mixture_2024}, we train the adapter using bootstrapped prompt simulations constructed from the downstream dataset. Let $\mathcal{D}_{\text{train}} = \{(x_i, y_i)\}_{i=1}^{N_{\text{train}}}$ denote the downstream dataset. We approximate the inference-time prompt distribution by sampling $ \mathcal{D}_{\text{bootstrap}} \sim p(\mathcal{D} \mid \mathcal{D}_{\text{train}})$. For large datasets ($N_{\text{train}} > 3000$), we first sample an anchor instance $(x_q, y_q) \sim p(x,y \mid \mathcal{D}_{\text{train}})$, and construct its context set via SNNL-based retrieval:
\begin{equation}
    \mathcal{D}_{\text{context}} = \mathrm{SNNL\text{-}Retrieval}(x_q \mid \mathcal{D}_{\text{train}}, B)
\end{equation}
where $B$ denotes the context size.  The bootstrapped prompt is then formed as $\mathcal{D}_{\text{bootstrap}}=\mathcal{D}_{\text{context}} \cup \{(x_q, y_q)\}$. For smaller datasets, we instead sample a large subset of $\mathcal{D}_{\text{train}}$ as $\mathcal{D}_{\text{context}}$ and treat the remaining samples as pseudo-query points.}

We freeze the pretrained backbone transformer  $q_{\theta_0}(y \mid x, \mathcal{D})$ and optimize only the adapter parameters $\psi$.  The adapted model is defined as
\begin{equation}
    q_{\theta_0,\psi}(y \mid x, \mathcal{D}) = q_{\theta_0}\big(y \mid A_\psi(x), A_\psi(\mathcal{D})\big)
\end{equation}
The adapter is trained by minimizing the negative log-likelihood over bootstrapped prompts:
\begin{equation}
    \mathcal{L}_{\text{adapter}} = - \mathbb{E}_{\mathcal{D}_{\text{bootstrap}}} \log q_{\theta_0,\psi} \big( y_q \mid x_q, \mathcal{D}_{\text{context}} \big)
\end{equation}
Intuitively, the frozen backbone encodes the synthetic pretraining prior $p(\mathcal{D} \mid \phi)p(\phi)$, while the adapter learns a dataset-specific correction that aligns the model with the empirical prompt distribution induced by SNNL-based retrieval, $p(\mathcal{D} \mid \mathcal{D}_{\text{train}})$. This preserves prior structure and stability while enabling task-aware representation alignment under realistic EHR context construction.

\subsection{Imbalance-aware Training and Retrieval Stabilization.}
Many EHR-based clinical prediction tasks exhibit severe class imbalance, where majority-class examples can dominate embedding structure and distort neighborhood quality. To mitigate this effect during AWARE training, we adopt balanced mini-batch sampling. Specifically, \highlightfour{we propose two variants for classification tasks and regression tasks.}

Regarding classification, for a mini-batch $\mathcal{B}$, samples are drawn with probability inversely proportional to class frequency:
\begin{equation}
    \mathbb{P}\big((x_i, y_i) \in \mathcal{B}\big) \propto \frac{1}{N_{y_i}},
\end{equation}
where $N_{y_i}$ denotes the number of training instances with label $y_i$. This ensures approximately uniform class representation within each mini-batch, stabilizing neighborhood formation in embedding space, particularly for rare outcomes.

For regression tasks, where targets are continuous rather than discrete, we employ a distribution-aware balancing strategy by first discretizing the target space into quantile-based bins and sampling instances inversely proportional to bin frequency:
\begin{equation}
\label{eqn:regression}
    \mathbb{P}\big((x_i, y_i) \in \mathcal{B}\big) \propto \frac{1}{N_{b(y_i)}},
\end{equation}
where $b(y_i)$ denotes the quantile bin associated with target value $y_i$, and $N_{b(y_i)}$ is the number of training instances within that bin. This encourages more uniform coverage of the continuous target distribution and reduces overrepresentation of densely populated target ranges during training.

To further reduce variance in retrieval behavior, we employ a $K$-fold cross-validation ensemble of AWARE encoders (with $K=5$ in our experiments). Let $ \{\mathcal{D}^{(k)}_{\text{train}}\}_{k=1}^{K} $ denote the $K$ disjoint training folds. For each fold $k$, we train an independent encoder $ f_{\theta^{(k)}} : \mathbb{R}^d \rightarrow \mathbb{R}^m $ using SNNL with balanced sampling, yielding embeddings $ z^{(k)} = f_{\theta^{(k)}}(x). $ At inference time, embeddings are aggregated across folds to obtain a stabilized representation:
\begin{equation}
    \bar{z}(x) = \frac{1}{K} \sum_{k=1}^{K} f_{\theta^{(k)}}(x).
\end{equation}
Retrieval is then performed in the ensemble-averaged embedding space using $\bar{z}(x)$ for both query and candidate support instances.

\highlightprism{
The final AWARE training objective thus consists solely of the Soft Nearest Neighbor Loss, while balanced sampling and cross-validated ensembling act as regularization mechanisms that reduce sampling variance and improve retrieval stability in high-dimensional and imbalanced EHR settings.}

\highlightrev{In summary, AWARE is a complementary adaptation layer for RA-TICL rather than a modification of the underlying in-context learning mechanism. The retrieval encoder, and optional adapter where applicable, are trained on the downstream training data and remain fixed at inference. For each query, AWARE constructs a per-query task-aligned context that is passed to the frozen RA-TICL backbone, whose pretrained parameters and prediction procedure remain unchanged. AWARE therefore improves retrieval and representation alignment around an existing RA-TICL model rather than constituting a purely inference-time method or a new ICL mechanism. }

\subsection{Experimental Setup}
\label{sec:experiments}

\subsubsection{Benchmark Design.}
We evaluate \highlightfirst{various methods} on a diverse collection of EHR datasets selected to reflect the heterogeneity, sparsity, and distribution complexity of real-world clinical data. The datasets include structured diagnosis and procedure codes, demographic variables, laboratory measurements, and time-series vital signs. To minimize potential overlap with model pretraining data, we exclude datasets reported as part of the pretraining corpus of the benchmarked models. Prediction tasks span regression, binary classification, and multiclass classification, with both balanced and highly imbalanced outcome distributions.

\subsubsection{Datasets and Tasks Choice.}
Our experiments draw from four complementary data sources. First, we use MIMIC-IV~\cite{johnson_mimic-iv_2023}, a large single-center critical care dataset, and derive \highlightprism{infection-related prediction tasks such as sepsis, urinary tract infection (UTI), and ventilator-associated pneumonia (VAP)}. Second, we evaluate on eICU~\cite{pollard2018eicu}, a multi-center intensive care unit (ICU) dataset spanning over 200 hospitals, \highlightrevdel{which supports assessment of robustness under institutional and cohort shifts, particularly for sepsis and UTI} \highlightrev{providing a heterogeneous multi-center cohort for evaluating sepsis and UTI prediction}. \highlighthipe{Third, we use a private daily-recorded cohort from a large acute hospital system, codenamed HIPE~\cite{pham_forecasting_2024, pham2025explainable}, for prediction of rare infections such as CPE}. Finally, we include a curated set of 12 OpenML/UCI clinical \highlightprism{datasets that provide small- to medium-scale EHR-based supervised learning tasks.} \highlightprism{These datasets support a systematic study of data scale, feature heterogeneity, severe outcome rarity, and} \highlightrevdel{cross-cohort generalizability} \highlightrev{consistency across cohorts} \highlightprism{across diverse EHR-based clinical prediction settings.} \highlightappendix{We additionally report experiments with Synthea~\cite{walonoski_synthea_2018} in Supplementary Table~\safeRef{tab:results:synthea}{7}. The full list of datasets used and their statistics can be found in Supplementary Table~\safeRef{tab:dataset_stats}{8}.}

\subsubsection{Feature Engineering and Preprocessing.}

For the OpenML/UCI datasets, we retain all available features and apply standard preprocessing, including mean imputation for numerical variables, mode imputation for categorical variables, and feature normalization. \highlightsecond{Missingness is handled through this standardized imputation pipeline to ensure comparability across models and datasets. }  \highlightrev{Because explicitly modeling MAR or MNAR mechanisms is beyond the scope of this methodological evaluation, we treat missingness as an inherent component of EHR heterogeneity and compare learning paradigms under consistent preprocessing assumptions~\cite{getzen2023mining}. All imputation and normalization parameters are estimated from the training split and applied unchanged to the validation and test splits.}

For MIMIC-IV and eICU, we follow established EHR preprocessing pipelines~\cite{gupta_extensive_2022, wu2024multimodal}. Because these cohorts contain thousands of highly sparse variables, dimensionality reduction is required for computational feasibility and cross-model compatibility. We therefore apply model-agnostic filtering based on variance and feature prevalence, removing near-constant variables and extremely rare features before capping the dimensionality at 500 when necessary. \highlightrev{All filtering criteria are determined using the training split only.} This procedure avoids architecture-specific inductive bias in feature selection while retaining a broad and heterogeneous representation of the EHR feature space. Time-series measurements are summarized using simple aggregation statistics (mean, minimum, maximum, and standard deviation).

\subsubsection{Training Configuration and Hyperparameters.}

Most tabular models are trained and evaluated under a unified experimental protocol to ensure fair and controlled comparisons. In particular, their hyperparameters are optimized independently for each dataset using 30 validation trials, ensuring a matched tuning budget across approaches. For each selected configuration, models are retrained with three random seeds to estimate performance variability. \highlightrev{The full hyperparameter search space for every classical machine learning and deep tabular baseline is reported in Supplementary Note \safeRef{sec:apdx:hyperparams}{5.6} (Supplementary Table \safeRef{tab:hyperparameter_ranges}{9}).}

For retrieval-based methods using AWARE, the retrieval encoders are trained independently of the downstream TICL models. We implemented SNNL as described in Equation~\ref{eq:snnl}, using cosine distance and temperature (T=0.1). For TabPFN-based models, we use a fixed version of TabPFNv2 throughout all experiments to ensure reproducibility. \highlightrevdel{All training is performed for 50 epochs using the AdamW optimizer with learning rate ($10^{-3}$) and task-dependent batch sizes determined by memory constraints.} \highlightrev{Retrieval encoders are trained for 50 epochs using the AdamW optimizer with a learning rate of $10^{-3}$.} Retrieval encoders are trained separately for each dataset, task, and ensemble fold, where applicable, using training data only. They remain frozen during adapter optimization and TICL inference, thereby preventing joint optimization with the pretrained backbone.

For adapter fine-tuning, we train the adapter for five epochs with relatively small learning rate. The adapter parameters account for approximately (0.01\%) of the total parameters of the corresponding RA-TICL backbone. \highlightfour{This adapter fine-tuning is applicable only to backbones with trainable downstream parameters (kNNPFN and TabDPT). For plain KNN, which has no learnable parameters beyond the retrieval encoder, AWARE reduces to the retrieval-alignment stage only; the notation KNN+AWARE throughout this paper refers to this retrieval-only configuration.}

For all retrieval-augmented models, unless otherwise specified, we use a fixed context size of 1024 labeled instances (for large-scale datasets) or a multiplier of 32 instances (capped at 1024, for small-to-medium scaled ones) to balance retrieval diversity and computational feasibility. This value is held constant across experiments to ensure comparability. Full definitions and descriptions of all benchmarked models are provided in Supplementary Note \safeRef{sec:apdx:algorithms}{5.5}.

\subsubsection{Evaluation Configuration.}
\highlighthipe{We use stratified patient-level train / test splits for all longitudinal EHR datasets (MIMIC-IV, eICU, and HIPE), so patients are unique across splits. HIPE provides an additional independent cohort with distinct coding practices and patient populations; however, models are trained and evaluated separately within each dataset. Feature extraction and preprocessing use statistics computed on the training split only (e.g., normalization and imputation parameters), and these are reused unchanged for validation and test.} 

\subsubsection{Evaluation Metrics.}
We report evaluation metrics commonly used in clinical prediction. For classification tasks, we report the area under the receiver operating characteristic curve (AUROC), which measures the probability that a randomly chosen positive instance is assigned a higher predicted score than a randomly chosen negative instance. The area under the precision--recall curve (AUPRC) summarizes the trade-off between precision and recall across decision thresholds and is particularly informative under severe class imbalance \cite{saito2015precision}.

\noindent
\highlightrevdel{For retrieval-based models, we evaluate retrieval quality using Precision@$k$, defined as:
where $y_i$ denotes the ground-truth label of the $i$-th retrieved instance. }

\noindent
For multiclass classification, we additionally report the F1-score, defined as the harmonic mean of precision and recall:
\[
\mathrm{F1} = \frac{2 \cdot \mathrm{Precision} \cdot \mathrm{Recall}}{\mathrm{Precision} + \mathrm{Recall}}.
\]

\noindent
For regression tasks, performance is measured using mean absolute error (MAE) and root mean squared error (RMSE):
\[
\mathrm{MAE} = \frac{1}{n} \sum_{i=1}^{n} |y_i - \hat{y}_i|,
\qquad
\mathrm{RMSE} = \sqrt{\frac{1}{n} \sum_{i=1}^{n} (y_i - \hat{y}_i)^2 }.
\]

\section*{Data Availability}

The datasets analyzed in this study are publicly available or accessible through established data use agreements. MIMIC-IV is available via PhysioNet (\url{https://physionet.org/}) to credentialed users who complete the required data use training and agreements. The eICU Collaborative Research Database is available through PhysioNet under similar access requirements. The OpenML and UCI datasets used in this study are publicly accessible via their respective repositories. The privately held HIPE dataset was used under institutional data use agreements and is not publicly available due to privacy and regulatory restrictions. Access to this dataset may be considered by the data custodians upon reasonable request and subject to appropriate approvals. All data used in this study are either publicly accessible through the sources described above or governed by third-party data use agreements.

\section*{Code Availability}

The source code used for data preprocessing, model training, evaluation, and generation of figures in this study is publicly available at \url{https://github.com/kaylode/aware}, with documentation sufficient to reproduce the main results.

\section*{Acknowledgements}

This research was conducted with the financial support of Taighde Éireann-Research Ireland under Grant Agreement No. 13/RC/2106\_P2 at ADAPT, the Research Ireland Centre for AI-Driven Digital Content Technology at DCU funded through the Research Ireland Research Centres Programme. For the purpose of Open Access, the author has applied a CC BY public copyright licence to any author-accepted manuscript version arising from this submission.

\section*{Author Contributions}

\textbf{M.-K.P.} conceptualized the study, developed the methodology, implemented the software, performed the experiments, conducted the formal analysis, and prepared the figures and manuscript. 
\textbf{T.-L.N.H.} contributed to the methodology, software implementation, experimental validation, and manuscript preparation. 
\textbf{T.T.M.}, \textbf{T.T.P.D.}, and \textbf{M.-T.T.} contributed to data analysis, visualization, validation of results, and manuscript writing. 
\textbf{M.E.W.}, \textbf{U.G.}, \textbf{R.B.}, and \textbf{N.M.} contributed to formal analysis and manuscript preparation. 
\textbf{M.B.} and \textbf{M.C.} supervised the project, contributed to study design and manuscript writing, and provided overall project administration. All authors reviewed and approved the final manuscript.


\section*{Competing Interests}
The authors declare no competing financial or non-financial interests.
\bibliographystyle{sn-nature}
\bibliography{references}

\clearpage
\onecolumn

\renewcommand\thefigure{\arabic{figure}}
\renewcommand\thetable{\arabic{table}}
\captionsetup[figure]{name=Supplementary Figure}
\captionsetup[figure*]{name=Supplementary Figure}
\captionsetup[table]{name=Supplementary Table}
\captionsetup[table*]{name=Supplementary Table}

\setcounter{figure}{0}
\setcounter{table}{0}

\section{Supplementary Information}

\subsection{Supplementary Note 1 | Ablation Studies}
\label{sec:apdx:ablation}

\highlightrevdel{
\textbf{AWARE} incorporates feature-level attention, imbalance-aware training, adapter-based domain adaptation, and ensemble aggregation to improve minority patient retrieval and stabilize context selection for rare infection prediction. Formally, the AWARE framework can be summarized as:}

\highlightrevdel{We hypothesize that retrieval-aligned neighborhood learning combined with imbalance-aware context selection will yield the largest improvements in AUPRC for rare clinical outcomes, where standard nearest-neighbor retrieval is strongly biased toward majority patient populations.}

\paragraph{Retrieval Objective Comparison.}
We compare several embedding-based retrieval objectives to identify the most effective strategy for retrieval-aligned in-context learning under severe clinical class imbalance. These methods differ primarily in how they structure the patient representation space and utilize outcome supervision: \textbf{kNN-L2} retrieves patients via Euclidean distance without representation learning; \textbf{Siamese} employs triplet-based metric learning but its rigid pairwise constraints fragment local neighborhood structure; \textbf{UnsupCon} learns label-free invariant representations that may fail to preserve rare disease structure; and \textbf{SupCon} improves class separability via label supervision but risks over-collapsing heterogeneous minority subpopulations. In contrast, \textbf{SNNL} directly optimizes local neighborhood composition via a soft probabilistic objective, preserving label coherence without sacrificing geometric spread. Supplementary Table~\ref{tab:retrieval_objectives} summarizes these objectives.

\begin{table*}[ht]
\centering
\caption{Comparison of different embedding-based retrieval objectives on large-scale ICU EHR tasks (MIMIC-IV and eICU). Performance is evaluated across models and datasets to identify the most effective objective for AWARE's retrieval stage. Markers ($^{*}$p$<$0.05, $^{**}$p$<$0.01) on a retrieval objective indicate it significantly outperforms SNNL (Full AWARE) (paired bootstrap, Holm-Bonferroni).}
\label{tab:retrieval_objectives}
\resizebox{\textwidth}{!}{
\renewcommand{\arraystretch}{0.95}
\begin{tabular}{lcccccccccc}
\toprule
\multirow{3}{*}{\textbf{Objective}} 
& \multicolumn{6}{c}{\textbf{MIMIC-IV}} 
& \multicolumn{4}{c}{\textbf{eICU}} \\
\cmidrule(lr){2-7} \cmidrule(lr){8-11}
& \multicolumn{2}{c}{\textbf{SEPSIS}} 
& \multicolumn{2}{c}{\textbf{VAP}}
& \multicolumn{2}{c}{\textbf{UTI}} 
& \multicolumn{2}{c}{\textbf{SEPSIS}}
& \multicolumn{2}{c}{\textbf{UTI}} \\
\cmidrule(lr){2-3} \cmidrule(lr){4-5} \cmidrule(lr){6-7}
\cmidrule(lr){8-9} \cmidrule(lr){10-11} 
& AUROC $\uparrow$ & AUPRC $\uparrow$
& AUROC $\uparrow$ & AUPRC $\uparrow$
& AUROC $\uparrow$ & AUPRC $\uparrow$
& AUROC $\uparrow$ & AUPRC $\uparrow$
& AUROC $\uparrow$ & AUPRC $\uparrow$ \\
\midrule
\multicolumn{11}{l}{\textbf{\textit{KNN}}} \\
\midrule
\hspace{3mm} SNNL & \ensuremath{0.795_{{\textcolor{gray}{\pm 0.002}}}} & \ensuremath{0.394_{{\textcolor{gray}{\pm 0.011}}}} & \ensuremath{0.710_{{\textcolor{gray}{\pm 0.005}}}} & \ensuremath{0.233_{{\textcolor{gray}{\pm 0.011}}}} & \ensuremath{0.586_{{\textcolor{gray}{\pm 0.005}}}} & \ensuremath{0.218_{{\textcolor{gray}{\pm 0.007}}}} & \ensuremath{0.803_{{\textcolor{gray}{\pm 0.006}}}} & \ensuremath{0.325_{{\textcolor{gray}{\pm 0.018}}}} & \ensuremath{0.757_{{\textcolor{gray}{\pm 0.006}}}} & \ensuremath{0.062_{{\textcolor{gray}{\pm 0.001}}}}  \\
\hspace{3mm} kNN L2 & \ensuremath{0.794_{{\textcolor{gray}{\pm 0.013}}}} & \ensuremath{0.391_{{\textcolor{gray}{\pm 0.026}}}} & \ensuremath{0.709_{{\textcolor{gray}{\pm 0.015}}}} & \ensuremath{0.213_{{\textcolor{gray}{\pm 0.008}}}} & \ensuremath{0.569_{{\textcolor{gray}{\pm 0.025}}}} & \ensuremath{0.209_{{\textcolor{gray}{\pm 0.025}}}} & \ensuremath{0.795_{{\textcolor{gray}{\pm 0.023}}}} & \ensuremath{0.314_{{\textcolor{gray}{\pm 0.027}}}} & \ensuremath{0.734_{{\textcolor{gray}{\pm 0.041}}}} & \ensuremath{0.056_{{\textcolor{gray}{\pm 0.011}}}}  \\
\hspace{3mm} Siamese & \ensuremath{0.868_{{\textcolor{gray}{\pm 0.006}}}} & \ensuremath{0.453_{{\textcolor{gray}{\pm 0.011}}}} & \ensuremath{0.756_{{\textcolor{gray}{\pm 0.005}}}} & \ensuremath{0.235_{{\textcolor{gray}{\pm 0.014}}}} & \ensuremath{0.608_{{\textcolor{gray}{\pm 0.012}}}} & \ensuremath{0.216_{{\textcolor{gray}{\pm 0.009}}}} & \ensuremath{0.847_{{\textcolor{gray}{\pm 0.010}}}} & \ensuremath{0.332_{{\textcolor{gray}{\pm 0.017}}}} & \ensuremath{0.677_{{\textcolor{gray}{\pm 0.010}}}} & \ensuremath{0.037_{{\textcolor{gray}{\pm 0.004}}}}  \\
\hspace{3mm} UnsupCon & \ensuremath{\underline{0.881}_{{\textcolor{gray}{\pm 0.001}}}} & \ensuremath{\underline{0.473}_{{\textcolor{gray}{\pm 0.002}}}} & \ensuremath{\underline{0.775}_{{\textcolor{gray}{\pm 0.000}}}} & \ensuremath{{\underline{0.259}}^{**}_{{\textcolor{gray}{\pm 0.002}}}} & \ensuremath{\textbf{0.647}_{{\textcolor{gray}{\pm 0.001}}}} & \ensuremath{\textbf{0.253}_{{\textcolor{gray}{\pm 0.000}}}} & \ensuremath{\underline{0.905}_{{\textcolor{gray}{\pm 0.002}}}} & \ensuremath{\underline{0.466}_{{\textcolor{gray}{\pm 0.021}}}} & \ensuremath{\textbf{0.861}_{{\textcolor{gray}{\pm 0.006}}}} & \ensuremath{\textbf{0.088}_{{\textcolor{gray}{\pm 0.001}}}}  \\
\hspace{3mm} SupCon & \ensuremath{\textbf{0.887}_{{\textcolor{gray}{\pm 0.009}}}} & \ensuremath{\textbf{0.498}_{{\textcolor{gray}{\pm 0.005}}}} & \ensuremath{\textbf{0.778}_{{\textcolor{gray}{\pm 0.003}}}} & \ensuremath{{\textbf{0.273}}^{**}_{{\textcolor{gray}{\pm 0.007}}}} & \ensuremath{\underline{0.646}_{{\textcolor{gray}{\pm 0.002}}}} & \ensuremath{\underline{0.247}_{{\textcolor{gray}{\pm 0.003}}}} & \ensuremath{\textbf{0.957}_{{\textcolor{gray}{\pm 0.002}}}} & \ensuremath{\textbf{0.651}_{{\textcolor{gray}{\pm 0.013}}}} & \ensuremath{\underline{0.853}_{{\textcolor{gray}{\pm 0.009}}}} & \ensuremath{\underline{0.084}_{{\textcolor{gray}{\pm 0.003}}}}  \\
\midrule
\multicolumn{11}{l}{\textbf{\textit{kNNPFN}}} \\
\midrule
\hspace{3mm} SNNL & \ensuremath{\underline{0.898}_{{\textcolor{gray}{\pm 0.000}}}} & \ensuremath{\underline{0.510}_{{\textcolor{gray}{\pm 0.000}}}} & \ensuremath{\underline{0.782}_{{\textcolor{gray}{\pm 0.000}}}} & \ensuremath{\textbf{0.277}_{{\textcolor{gray}{\pm 0.000}}}} & \ensuremath{\textbf{0.658}_{{\textcolor{gray}{\pm 0.000}}}} & \ensuremath{\textbf{0.254}_{{\textcolor{gray}{\pm 0.000}}}} & \ensuremath{\textbf{0.957}_{{\textcolor{gray}{\pm 0.001}}}} & \ensuremath{\textbf{0.638}_{{\textcolor{gray}{\pm 0.002}}}} & \ensuremath{\underline{0.873}_{{\textcolor{gray}{\pm 0.004}}}} & \ensuremath{\underline{0.093}_{{\textcolor{gray}{\pm 0.002}}}}  \\
\hspace{3mm} kNN L2 & \ensuremath{0.897_{{\textcolor{gray}{\pm 0.000}}}} & \ensuremath{0.509_{{\textcolor{gray}{\pm 0.000}}}} & \ensuremath{\underline{0.782}_{{\textcolor{gray}{\pm 0.000}}}} & \ensuremath{\underline{0.276}_{{\textcolor{gray}{\pm 0.000}}}} & \ensuremath{\underline{0.657}_{{\textcolor{gray}{\pm 0.000}}}} & \ensuremath{\underline{0.253}_{{\textcolor{gray}{\pm 0.000}}}} & \ensuremath{0.947_{{\textcolor{gray}{\pm 0.000}}}} & \ensuremath{0.615_{{\textcolor{gray}{\pm 0.000}}}} & \ensuremath{0.867_{{\textcolor{gray}{\pm 0.000}}}} & \ensuremath{0.091_{{\textcolor{gray}{\pm 0.000}}}}  \\
\hspace{3mm} Siamese & \ensuremath{0.892_{{\textcolor{gray}{\pm 0.004}}}} & \ensuremath{0.501_{{\textcolor{gray}{\pm 0.009}}}} & \ensuremath{\textbf{0.783}_{{\textcolor{gray}{\pm 0.002}}}} & \ensuremath{0.259_{{\textcolor{gray}{\pm 0.018}}}} & \ensuremath{0.641_{{\textcolor{gray}{\pm 0.008}}}} & \ensuremath{0.233_{{\textcolor{gray}{\pm 0.006}}}} & \ensuremath{0.935_{{\textcolor{gray}{\pm 0.003}}}} & \ensuremath{0.566_{{\textcolor{gray}{\pm 0.011}}}} & \ensuremath{0.824_{{\textcolor{gray}{\pm 0.015}}}} & \ensuremath{0.069_{{\textcolor{gray}{\pm 0.004}}}}  \\
\hspace{3mm} UnsupCon & \ensuremath{\textbf{0.899}_{{\textcolor{gray}{\pm 0.001}}}} & \ensuremath{\textbf{0.520}_{{\textcolor{gray}{\pm 0.002}}}} & \ensuremath{0.777_{{\textcolor{gray}{\pm 0.002}}}} & \ensuremath{0.260_{{\textcolor{gray}{\pm 0.000}}}} & \ensuremath{0.653_{{\textcolor{gray}{\pm 0.001}}}} & \ensuremath{0.251_{{\textcolor{gray}{\pm 0.002}}}} & \ensuremath{\underline{0.956}_{{\textcolor{gray}{\pm 0.000}}}} & \ensuremath{\underline{0.636}_{{\textcolor{gray}{\pm 0.003}}}} & \ensuremath{\textbf{0.874}_{{\textcolor{gray}{\pm 0.004}}}} & \ensuremath{\textbf{0.097}_{{\textcolor{gray}{\pm 0.002}}}}  \\
\hspace{3mm} SupCon & \ensuremath{0.894_{{\textcolor{gray}{\pm 0.008}}}} & \ensuremath{0.501_{{\textcolor{gray}{\pm 0.011}}}} & \ensuremath{0.779_{{\textcolor{gray}{\pm 0.001}}}} & \ensuremath{0.252_{{\textcolor{gray}{\pm 0.007}}}} & \ensuremath{\textbf{0.658}_{{\textcolor{gray}{\pm 0.008}}}} & \ensuremath{0.249_{{\textcolor{gray}{\pm 0.008}}}} & \ensuremath{0.955_{{\textcolor{gray}{\pm 0.002}}}} & \ensuremath{0.633_{{\textcolor{gray}{\pm 0.004}}}} & \ensuremath{0.871_{{\textcolor{gray}{\pm 0.005}}}} & \ensuremath{\underline{0.093}_{{\textcolor{gray}{\pm 0.006}}}}  \\
\midrule
\multicolumn{11}{l}{\textbf{\textit{TabDPT}}} \\
\midrule
\hspace{3mm} SNNL & \ensuremath{\underline{0.906}_{{\textcolor{gray}{\pm 0.000}}}} & \ensuremath{\underline{0.529}_{{\textcolor{gray}{\pm 0.000}}}} & \ensuremath{\underline{0.785}_{{\textcolor{gray}{\pm 0.000}}}} & \ensuremath{0.274_{{\textcolor{gray}{\pm 0.000}}}} & \ensuremath{0.659_{{\textcolor{gray}{\pm 0.000}}}} & \ensuremath{0.257_{{\textcolor{gray}{\pm 0.000}}}} & \ensuremath{\underline{0.967}_{{\textcolor{gray}{\pm 0.001}}}} & \ensuremath{0.689_{{\textcolor{gray}{\pm 0.006}}}} & \ensuremath{\underline{0.892}_{{\textcolor{gray}{\pm 0.003}}}} & \ensuremath{0.112_{{\textcolor{gray}{\pm 0.005}}}}  \\
\hspace{3mm} kNN L2 & \ensuremath{0.905_{{\textcolor{gray}{\pm 0.000}}}} & \ensuremath{0.528_{{\textcolor{gray}{\pm 0.000}}}} & \ensuremath{0.784_{{\textcolor{gray}{\pm 0.000}}}} & \ensuremath{0.274_{{\textcolor{gray}{\pm 0.000}}}} & \ensuremath{0.658_{{\textcolor{gray}{\pm 0.000}}}} & \ensuremath{0.257_{{\textcolor{gray}{\pm 0.000}}}} & \ensuremath{0.961_{{\textcolor{gray}{\pm 0.000}}}} & \ensuremath{0.659_{{\textcolor{gray}{\pm 0.000}}}} & \ensuremath{0.882_{{\textcolor{gray}{\pm 0.000}}}} & \ensuremath{0.103_{{\textcolor{gray}{\pm 0.000}}}}  \\
\hspace{3mm} Siamese & \ensuremath{0.860_{{\textcolor{gray}{\pm 0.004}}}} & \ensuremath{0.438_{{\textcolor{gray}{\pm 0.006}}}} & \ensuremath{0.704_{{\textcolor{gray}{\pm 0.012}}}} & \ensuremath{0.194_{{\textcolor{gray}{\pm 0.006}}}} & \ensuremath{0.591_{{\textcolor{gray}{\pm 0.014}}}} & \ensuremath{0.203_{{\textcolor{gray}{\pm 0.009}}}} & \ensuremath{0.851_{{\textcolor{gray}{\pm 0.006}}}} & \ensuremath{0.382_{{\textcolor{gray}{\pm 0.021}}}} & \ensuremath{0.640_{{\textcolor{gray}{\pm 0.011}}}} & \ensuremath{0.032_{{\textcolor{gray}{\pm 0.001}}}}  \\
\hspace{3mm} UnsupCon & \ensuremath{\textbf{0.909}_{{\textcolor{gray}{\pm 0.001}}}} & \ensuremath{\textbf{0.543}_{{\textcolor{gray}{\pm 0.003}}}} & \ensuremath{\underline{0.785}_{{\textcolor{gray}{\pm 0.003}}}} & \ensuremath{\underline{0.275}_{{\textcolor{gray}{\pm 0.003}}}} & \ensuremath{\underline{0.661}_{{\textcolor{gray}{\pm 0.001}}}} & \ensuremath{\underline{0.258}_{{\textcolor{gray}{\pm 0.002}}}} & \ensuremath{\textbf{0.969}_{{\textcolor{gray}{\pm 0.001}}}} & \ensuremath{\underline{0.693}_{{\textcolor{gray}{\pm 0.002}}}} & \ensuremath{\textbf{0.899}_{{\textcolor{gray}{\pm 0.001}}}} & \ensuremath{\textbf{0.125}_{{\textcolor{gray}{\pm 0.005}}}}  \\
\hspace{3mm} SupCon & \ensuremath{0.902_{{\textcolor{gray}{\pm 0.008}}}} & \ensuremath{0.524_{{\textcolor{gray}{\pm 0.002}}}} & \ensuremath{\textbf{0.788}_{{\textcolor{gray}{\pm 0.004}}}} & \ensuremath{\textbf{0.286}_{{\textcolor{gray}{\pm 0.005}}}} & \ensuremath{\textbf{0.663}_{{\textcolor{gray}{\pm 0.003}}}} & \ensuremath{\textbf{0.259}_{{\textcolor{gray}{\pm 0.003}}}} & \ensuremath{\underline{0.967}_{{\textcolor{gray}{\pm 0.001}}}} & \ensuremath{\textbf{0.694}_{{\textcolor{gray}{\pm 0.003}}}} & \ensuremath{0.883_{{\textcolor{gray}{\pm 0.004}}}} & \ensuremath{\underline{0.113}_{{\textcolor{gray}{\pm 0.002}}}}  \\

\bottomrule
\end{tabular}}
\end{table*}

\begin{figure}[ht]
    \centering
    \includegraphics[width=0.6\linewidth]{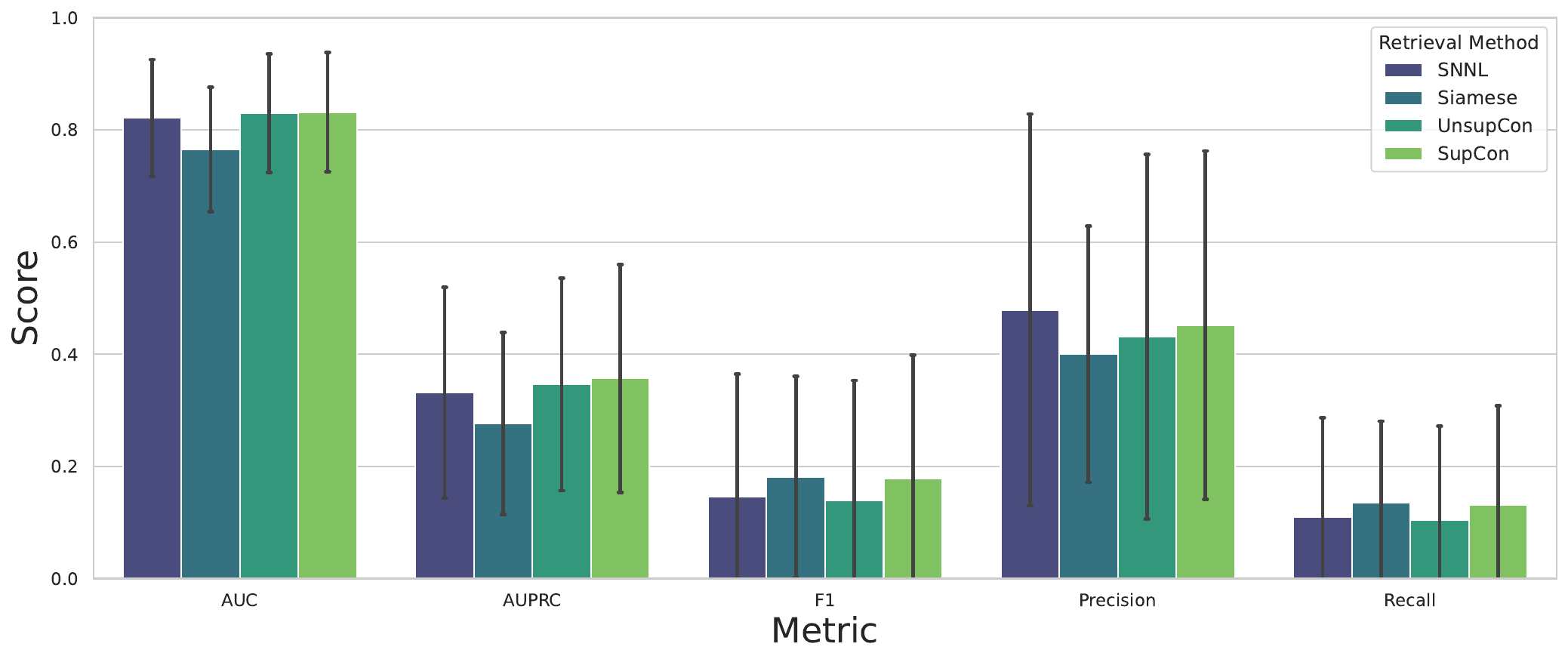}
    \caption{Downstream predictive performance across retrieval objectives. 
    Distributions overlap substantially, motivating retrieval-specific evaluation.}
    \label{fig:retrieval_comparison_icl}
\end{figure}

As shown in Supplementary Figure~\ref{fig:retrieval_comparison_icl}, downstream predictive performance is largely comparable across all methods. To look beyond end-to-end metrics, we evaluate retrieval quality directly via three metrics across varying context sizes. \highlightrevdel{\textit{Neighborhood Purity@k}, the fraction of retrieved neighbors sharing the query's outcome label; \textit{Contradiction Rate@k}, the proportion of retrieved neighbors carrying a conflicting label, which directly penalizes misleading in-context evidence; and \textit{Retrieval Diversity@k}, the geometric spread among retrieved neighbors, which detects representation collapse.}
\highlightrev{Given a query $\mathbf{x}_i$ with label $y_i$, its $k$ retrieved neighbors with labels $\{y_{i,1}, \dots, y_{i,k}\}$, and their corresponding embeddings $\{\mathbf{e}_{i,1}, \dots, \mathbf{e}_{i,k}\}$, we define the following metrics, each averaged over all $N$ queries in the evaluation set.

\textit{Neighborhood Purity@k} is the fraction of retrieved neighbors sharing the query's label:
\[
\mathrm{Purity@}k = \frac{1}{N}\sum_{i=1}^{N} \frac{1}{k}\sum_{j=1}^{k} \mathbbm{1}[y_{i,j} = y_i].
\]

\textit{Contradiction Rate@k} is the fraction of queries for which the retrieved neighborhood is \emph{majority-opposed} to the query label, i.e., where more than half of the $k$ retrieved neighbors carry a different label than the query:
\[
\mathrm{ContradictionRate@}k = \frac{1}{N}\sum_{i=1}^{N} \mathbbm{1}\!\left[\sum_{j=1}^{k}\mathbbm{1}[y_{i,j} = y_i] < \frac{k}{2}\right].
\]
Unlike Neighborhood Purity, which is a continuous per-neighbor average, Contradiction Rate is a discrete per-query indicator: it flags queries whose retrieved context would outvote the correct label if used naively for majority-vote inference, directly penalizing misleading in-context evidence.

\textit{Retrieval Diversity@k} measures the geometric spread among the $k$ retrieved neighbor embeddings, computed as the mean pairwise cosine distance over all ordered pairs of retrieved embeddings:
\[
\mathrm{Diversity@}k = \frac{1}{N}\sum_{i=1}^{N} \frac{1}{k(k-1)}\sum_{a \neq b} \mathrm{dist}_{\cos}(\mathbf{e}_{i,a}, \mathbf{e}_{i,b}), \qquad \mathrm{dist}_{\cos}(\mathbf{u}, \mathbf{v}) = 1 - \frac{\mathbf{u}\cdot\mathbf{v}}{\|\mathbf{u}\|\|\mathbf{v}\|}.
\]
Low diversity indicates representation collapse, where retrieved neighbors occupy a narrow region of the embedding space rather than spanning diverse but label-consistent examples.}

\begin{figure}[ht]
    \centering
    \includegraphics[width=\linewidth]{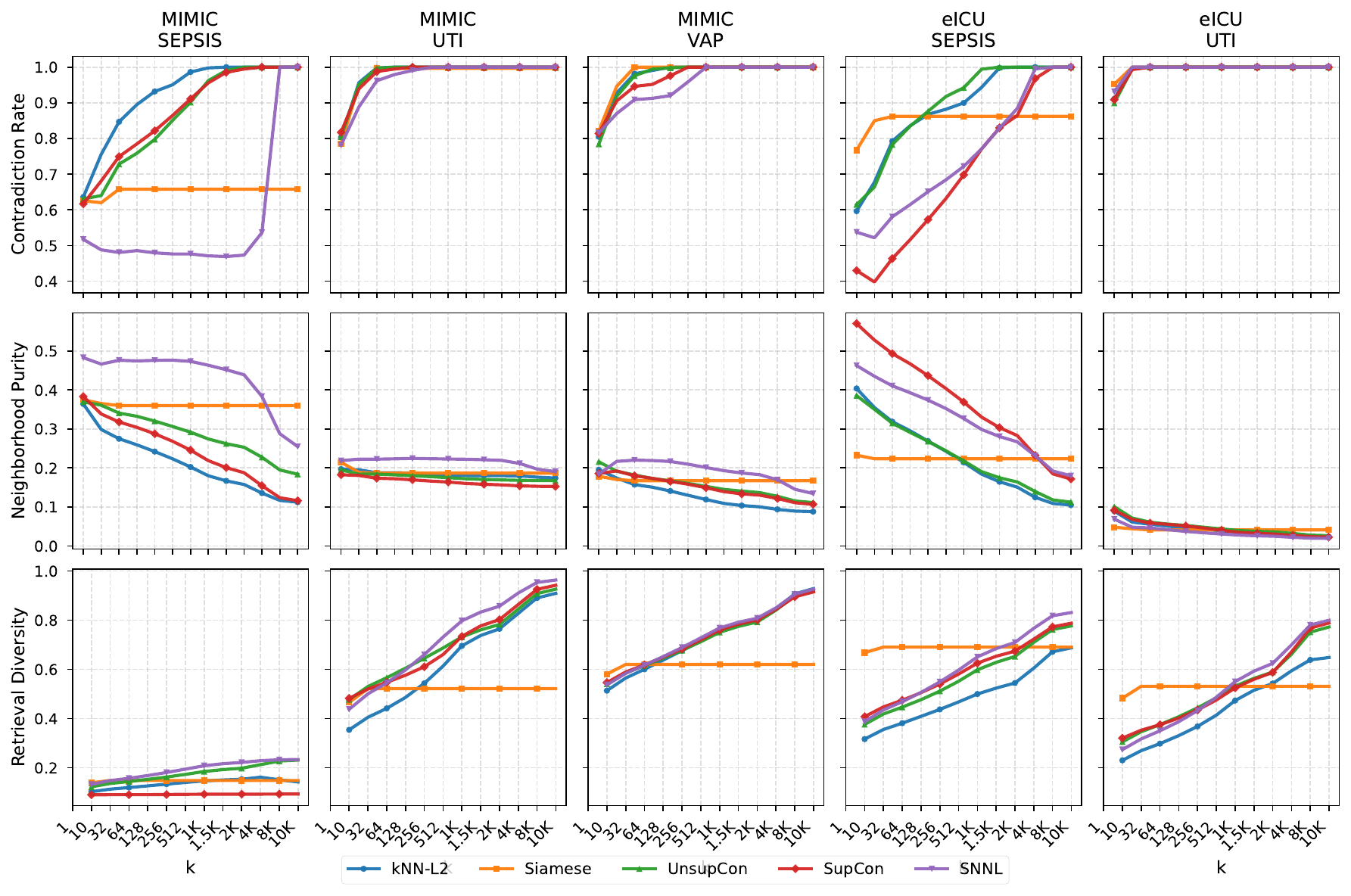}
    \caption{Retrieval quality across methods, tasks, and context sizes $k$. 
    SNNL achieves the best neighborhood purity and lowest contradiction rates at 
    small $k$, while maintaining diversity comparable to SupCon.}
    \label{fig:retrieval_comparison_ret}
\end{figure}

Supplementary Figure~\ref{fig:retrieval_comparison_ret} reveals structural differences not 
captured by predictive metrics. At small $k$, SNNL achieves the highest Neighborhood Purity on SEPSIS tasks, and uniquely suppresses Contradiction Rate growth that afflicts SupCon as $k$ increases. Siamese remains stable but consistently low across all metrics, while kNN-L2 and Siamese both exhibit near-zero Retrieval Diversity, indicating representation collapse. SupCon achieves comparable  diversity to SNNL but at the cost of coherence. On UTI and VAP tasks, all methods converge to modest purity, reflecting a prevalence floor beyond which representation learning alone cannot compensate. Overall, SNNL is the only objective that jointly achieves high purity, low contradiction rate, and high diversity, providing the most reliable retrieval foundation for imbalanced  clinical settings. Thus, we chose it as our primary loss function in the main text.

\begin{figure*}[ht]
\centering
\includegraphics[width=0.99\linewidth]{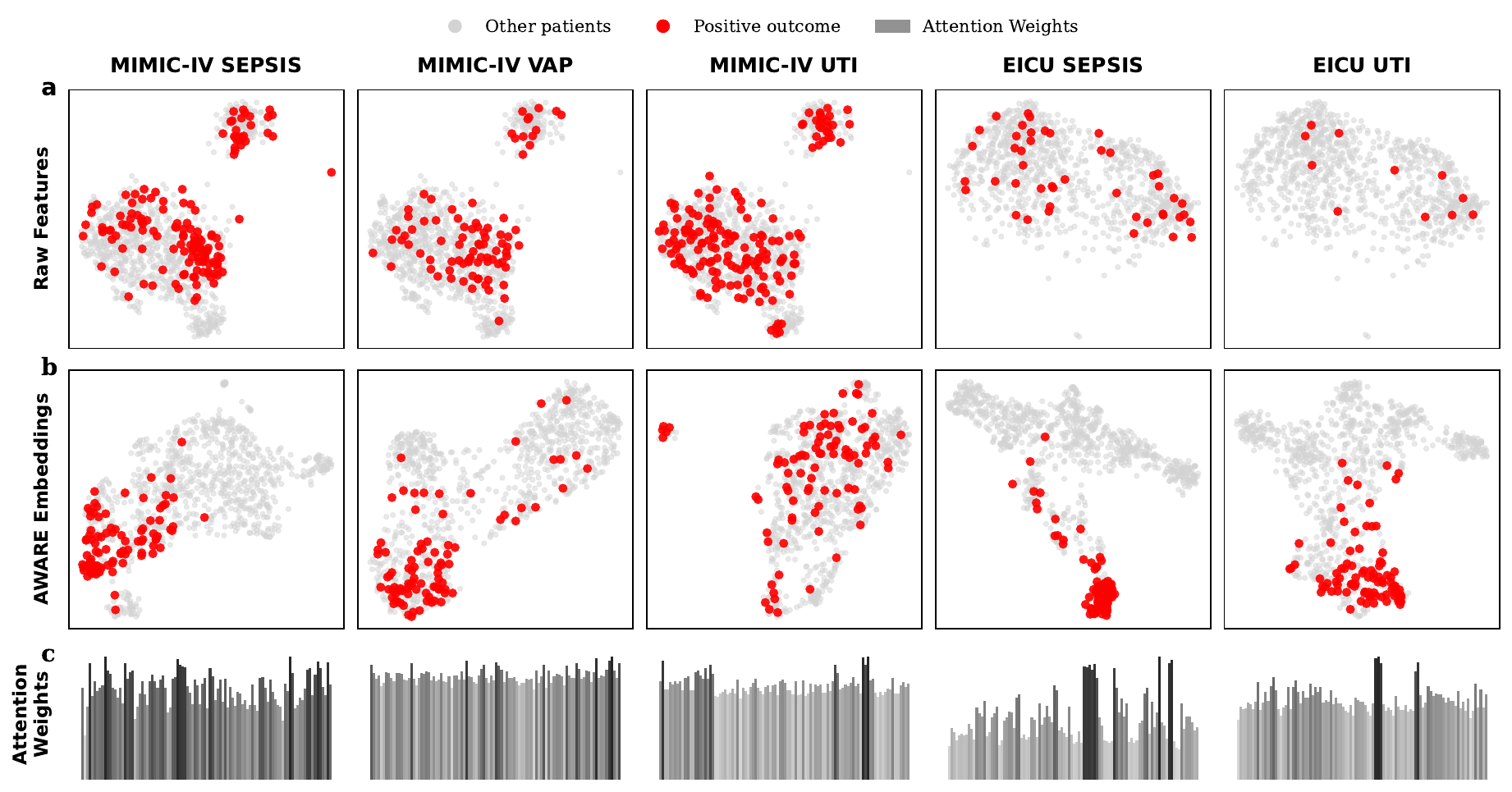}
\caption{
Visualization of patient neighborhood structure across MIMIC-IV and eICU dataset, illustrated by projecting data into 2D space. (a) In raw feature space, proximity does not reliably correspond to predictive relevance across different clinical tasks, resulting in scattered positive examples. (b) In contrast, task-aligned embedding retrieval (AWARE) reshapes the representation space such that label-consistent examples form coherent local clusters while preserving smooth structure for downstream in-context learning. (c) Black bars illustrate task-specific feature reweighting induced by the attention-based encoder, highlighting how retrieval adapts to heterogeneous clinical variables.
}
\label{fig:combined_umap_all_tasks}
\end{figure*}

\highlightrev{Figure~\ref{fig:combined_umap_all_tasks} illustrates a central limitation of relying on fixed similarity metrics in the raw EHR feature space. As shown in Figure~\ref{fig:combined_umap_all_tasks}a, positive cases are often dispersed or intermixed with negative cases, suggesting that patients retrieved using Euclidean or cosine distance may be feature-similar but clinically irrelevant or label-misaligned. Moreover, the same feature space supports multiple prediction tasks, each requiring a different notion of clinically meaningful similarity. In contrast, the task-aligned embeddings in Figure~\ref{fig:combined_umap_all_tasks}b form more coherent outcome-related neighborhoods, while the attention-weight profiles in Figure~\ref{fig:combined_umap_all_tasks}c vary across tasks and cohorts. Under these conditions, a single task-agnostic retrieval metric is unlikely to be universally appropriate, motivating adaptive, task-aligned retrieval for heterogeneous EHR data.}

\newpage

\paragraph{Component Ablation.}
Supplementary Tables~\ref{tab:ablation_aware_only}, \ref{tab:ablation_aware_wo}, and~\ref{tab:ablation_aware_progressive} present three complementary views of the AWARE ablation: individual component strength, leave-one-out removal, and progressive build-up, respectively. Across all three backbones (KNN, kNNPFN, TabDPT) and all five tasks, Full AWARE consistently achieves the best or near-best AUROC and AUPRC in the vast majority of configurations, confirming that the components are mutually reinforcing rather than redundant.

The progressive ablation (Supplementary Table~\ref{tab:ablation_aware_progressive}) reveals that balance sampling contributes the largest single-step gains, particularly on AUPRC for minority-heavy tasks such as eICU SEPSIS and eICU UTI, where adding balanced sampling alone raises AUPRC by substantial margins over the attention-only configuration. SNNL and attention individually provide modest but consistent improvements over the respective baselines, while ensemble aggregation and adapter tuning deliver the final incremental gains that close the gap to full model performance. The leave-one-out results (Supplementary Table~\ref{tab:ablation_aware_wo}) corroborate this ordering: removing balance sampling or attention causes the largest degradation, whereas removing the adapter has the smallest impact, particularly for KNN which does not include adapter tuning. Taken together, the ablations support retaining all components in the full AWARE configuration.

\begin{table*}[ht]
\centering
\caption{Component ablation of AWARE on large-scale ICU EHR tasks (MIMIC-IV and eICU). Each row represents a configuration where only one trained component is active (added on top of the frozen PFN backbone). $^\dagger$ Full AWARE version for KNN does not include Adapter Finetuning. Markers ($^{*}$p$<$0.05, $^{**}$p$<$0.01) on Full AWARE indicate it significantly outperforms all single-component variants (paired bootstrap, Holm-Bonferroni).}
\label{tab:ablation_aware_only}
\resizebox{\textwidth}{!}{
\renewcommand{\arraystretch}{0.95}
\begin{tabular}{lcccccccccc}
\toprule
\multirow{3}{*}{\textbf{Model}} 
& \multicolumn{6}{c}{\textbf{MIMIC-IV}} 
& \multicolumn{4}{c}{\textbf{eICU}} \\
\cmidrule(lr){2-7} \cmidrule(lr){8-11}
& \multicolumn{2}{c}{\textbf{SEPSIS}} 
& \multicolumn{2}{c}{\textbf{VAP}}
& \multicolumn{2}{c}{\textbf{UTI}} 
& \multicolumn{2}{c}{\textbf{SEPSIS}}
& \multicolumn{2}{c}{\textbf{UTI}} \\
\cmidrule(lr){2-3} \cmidrule(lr){4-5} \cmidrule(lr){6-7}
\cmidrule(lr){8-9} \cmidrule(lr){10-11} 
& AUROC $\uparrow$ & AUPRC $\uparrow$
& AUROC $\uparrow$ & AUPRC $\uparrow$
& AUROC $\uparrow$ & AUPRC $\uparrow$
& AUROC $\uparrow$ & AUPRC $\uparrow$
& AUROC $\uparrow$ & AUPRC $\uparrow$ \\
\midrule
\multicolumn{11}{l}{\textbf{\textit{KNN}}} \\
\midrule
\hspace{3mm} Full AWARE$^\dagger$ & \ensuremath{{\textbf{0.903}}^{**}_{{\textcolor{gray}{\pm 0.001}}}} & \ensuremath{{\textbf{0.507}}^{**}_{{\textcolor{gray}{\pm 0.006}}}} & \ensuremath{{\textbf{0.788}}^{**}_{{\textcolor{gray}{\pm 0.003}}}} & \ensuremath{0.250_{{\textcolor{gray}{\pm 0.005}}}} & \ensuremath{{\textbf{0.671}}^{**}_{{\textcolor{gray}{\pm 0.002}}}} & \ensuremath{{\textbf{0.260}}^{**}_{{\textcolor{gray}{\pm 0.004}}}} & \ensuremath{{\textbf{0.971}}^{**}_{{\textcolor{gray}{\pm 0.001}}}} & \ensuremath{{\textbf{0.711}}^{**}_{{\textcolor{gray}{\pm 0.002}}}} & \ensuremath{{\textbf{0.928}}^{**}_{{\textcolor{gray}{\pm 0.002}}}} & \ensuremath{{\textbf{0.155}}^{**}_{{\textcolor{gray}{\pm 0.003}}}}  \\
\hspace{3mm} Attention Only & \ensuremath{\underline{0.862}_{{\textcolor{gray}{\pm 0.006}}}} & \ensuremath{\underline{0.448}_{{\textcolor{gray}{\pm 0.021}}}} & \ensuremath{0.763_{{\textcolor{gray}{\pm 0.001}}}} & \ensuremath{\underline{0.257}_{{\textcolor{gray}{\pm 0.001}}}} & \ensuremath{0.629_{{\textcolor{gray}{\pm 0.007}}}} & \ensuremath{\underline{0.236}_{{\textcolor{gray}{\pm 0.001}}}} & \ensuremath{\underline{0.911}_{{\textcolor{gray}{\pm 0.033}}}} & \ensuremath{\underline{0.492}_{{\textcolor{gray}{\pm 0.112}}}} & \ensuremath{0.810_{{\textcolor{gray}{\pm 0.005}}}} & \ensuremath{0.065_{{\textcolor{gray}{\pm 0.001}}}}  \\
\hspace{3mm} SNNL Only & \ensuremath{0.795_{{\textcolor{gray}{\pm 0.002}}}} & \ensuremath{0.394_{{\textcolor{gray}{\pm 0.011}}}} & \ensuremath{0.710_{{\textcolor{gray}{\pm 0.005}}}} & \ensuremath{0.233_{{\textcolor{gray}{\pm 0.011}}}} & \ensuremath{0.586_{{\textcolor{gray}{\pm 0.005}}}} & \ensuremath{0.218_{{\textcolor{gray}{\pm 0.007}}}} & \ensuremath{0.803_{{\textcolor{gray}{\pm 0.006}}}} & \ensuremath{0.325_{{\textcolor{gray}{\pm 0.018}}}} & \ensuremath{0.757_{{\textcolor{gray}{\pm 0.006}}}} & \ensuremath{0.062_{{\textcolor{gray}{\pm 0.001}}}}  \\
\hspace{3mm} Balance Only & \ensuremath{0.851_{{\textcolor{gray}{\pm 0.005}}}} & \ensuremath{0.422_{{\textcolor{gray}{\pm 0.012}}}} & \ensuremath{0.764_{{\textcolor{gray}{\pm 0.005}}}} & \ensuremath{0.248_{{\textcolor{gray}{\pm 0.013}}}} & \ensuremath{0.630_{{\textcolor{gray}{\pm 0.004}}}} & \ensuremath{0.234_{{\textcolor{gray}{\pm 0.001}}}} & \ensuremath{0.851_{{\textcolor{gray}{\pm 0.009}}}} & \ensuremath{0.335_{{\textcolor{gray}{\pm 0.013}}}} & \ensuremath{0.821_{{\textcolor{gray}{\pm 0.009}}}} & \ensuremath{0.065_{{\textcolor{gray}{\pm 0.001}}}}  \\
\hspace{3mm} Ensemble Only & \ensuremath{\underline{0.862}_{{\textcolor{gray}{\pm 0.002}}}} & \ensuremath{0.440_{{\textcolor{gray}{\pm 0.011}}}} & \ensuremath{\underline{0.768}_{{\textcolor{gray}{\pm 0.004}}}} & \ensuremath{\textbf{0.261}_{{\textcolor{gray}{\pm 0.004}}}} & \ensuremath{\underline{0.631}_{{\textcolor{gray}{\pm 0.001}}}} & \ensuremath{0.235_{{\textcolor{gray}{\pm 0.004}}}} & \ensuremath{0.873_{{\textcolor{gray}{\pm 0.002}}}} & \ensuremath{0.365_{{\textcolor{gray}{\pm 0.004}}}} & \ensuremath{\underline{0.822}_{{\textcolor{gray}{\pm 0.005}}}} & \ensuremath{\underline{0.071}_{{\textcolor{gray}{\pm 0.004}}}}  \\
\midrule
\multicolumn{11}{l}{\textbf{\textit{kNNPFN}}} \\
\midrule
\hspace{3mm} Full AWARE & \ensuremath{{\textbf{0.917}}^{**}_{{\textcolor{gray}{\pm 0.000}}}} & \ensuremath{{\textbf{0.538}}^{*}_{{\textcolor{gray}{\pm 0.006}}}} & \ensuremath{{\textbf{0.799}}^{**}_{{\textcolor{gray}{\pm 0.000}}}} & \ensuremath{{\textbf{0.282}}^{**}_{{\textcolor{gray}{\pm 0.000}}}} & \ensuremath{{\textbf{0.673}}^{**}_{{\textcolor{gray}{\pm 0.002}}}} & \ensuremath{{\textbf{0.263}}^{**}_{{\textcolor{gray}{\pm 0.006}}}} & \ensuremath{{\textbf{0.981}}^{**}_{{\textcolor{gray}{\pm 0.000}}}} & \ensuremath{{\textbf{0.724}}^{**}_{{\textcolor{gray}{\pm 0.004}}}} & \ensuremath{{\textbf{0.932}}^{**}_{{\textcolor{gray}{\pm 0.002}}}} & \ensuremath{{\textbf{0.152}}^{**}_{{\textcolor{gray}{\pm 0.005}}}}  \\
\hspace{3mm} Attention Only & \ensuremath{0.885_{{\textcolor{gray}{\pm 0.002}}}} & \ensuremath{0.502_{{\textcolor{gray}{\pm 0.006}}}} & \ensuremath{0.783_{{\textcolor{gray}{\pm 0.001}}}} & \ensuremath{0.274_{{\textcolor{gray}{\pm 0.003}}}} & \ensuremath{0.656_{{\textcolor{gray}{\pm 0.002}}}} & \ensuremath{0.251_{{\textcolor{gray}{\pm 0.001}}}} & \ensuremath{0.954_{{\textcolor{gray}{\pm 0.002}}}} & \ensuremath{0.633_{{\textcolor{gray}{\pm 0.007}}}} & \ensuremath{0.875_{{\textcolor{gray}{\pm 0.003}}}} & \ensuremath{0.091_{{\textcolor{gray}{\pm 0.001}}}}  \\
\hspace{3mm} SNNL Only & \ensuremath{0.898_{{\textcolor{gray}{\pm 0.000}}}} & \ensuremath{0.510_{{\textcolor{gray}{\pm 0.000}}}} & \ensuremath{0.782_{{\textcolor{gray}{\pm 0.000}}}} & \ensuremath{\underline{0.277}_{{\textcolor{gray}{\pm 0.000}}}} & \ensuremath{\underline{0.658}_{{\textcolor{gray}{\pm 0.000}}}} & \ensuremath{\underline{0.254}_{{\textcolor{gray}{\pm 0.000}}}} & \ensuremath{0.957_{{\textcolor{gray}{\pm 0.001}}}} & \ensuremath{0.638_{{\textcolor{gray}{\pm 0.002}}}} & \ensuremath{0.873_{{\textcolor{gray}{\pm 0.004}}}} & \ensuremath{0.093_{{\textcolor{gray}{\pm 0.002}}}}  \\
\hspace{3mm} Balance Only & \ensuremath{0.898_{{\textcolor{gray}{\pm 0.001}}}} & \ensuremath{0.518_{{\textcolor{gray}{\pm 0.002}}}} & \ensuremath{0.783_{{\textcolor{gray}{\pm 0.002}}}} & \ensuremath{0.267_{{\textcolor{gray}{\pm 0.009}}}} & \ensuremath{0.654_{{\textcolor{gray}{\pm 0.002}}}} & \ensuremath{0.251_{{\textcolor{gray}{\pm 0.005}}}} & \ensuremath{0.954_{{\textcolor{gray}{\pm 0.001}}}} & \ensuremath{0.629_{{\textcolor{gray}{\pm 0.001}}}} & \ensuremath{0.873_{{\textcolor{gray}{\pm 0.005}}}} & \ensuremath{0.094_{{\textcolor{gray}{\pm 0.005}}}}  \\
\hspace{3mm} Ensemble Only & \ensuremath{0.900_{{\textcolor{gray}{\pm 0.001}}}} & \ensuremath{0.518_{{\textcolor{gray}{\pm 0.007}}}} & \ensuremath{0.781_{{\textcolor{gray}{\pm 0.001}}}} & \ensuremath{0.266_{{\textcolor{gray}{\pm 0.008}}}} & \ensuremath{0.655_{{\textcolor{gray}{\pm 0.001}}}} & \ensuremath{0.249_{{\textcolor{gray}{\pm 0.002}}}} & \ensuremath{0.959_{{\textcolor{gray}{\pm 0.001}}}} & \ensuremath{0.646_{{\textcolor{gray}{\pm 0.005}}}} & \ensuremath{0.873_{{\textcolor{gray}{\pm 0.001}}}} & \ensuremath{0.092_{{\textcolor{gray}{\pm 0.001}}}}  \\
\hspace{3mm} Adapter Only & \ensuremath{\underline{0.903}_{{\textcolor{gray}{\pm 0.001}}}} & \ensuremath{\underline{0.529}_{{\textcolor{gray}{\pm 0.005}}}} & \ensuremath{\underline{0.790}_{{\textcolor{gray}{\pm 0.002}}}} & \ensuremath{0.274_{{\textcolor{gray}{\pm 0.003}}}} & \ensuremath{0.654_{{\textcolor{gray}{\pm 0.005}}}} & \ensuremath{0.247_{{\textcolor{gray}{\pm 0.006}}}} & \ensuremath{\underline{0.962}_{{\textcolor{gray}{\pm 0.001}}}} & \ensuremath{\underline{0.651}_{{\textcolor{gray}{\pm 0.005}}}} & \ensuremath{\underline{0.894}_{{\textcolor{gray}{\pm 0.016}}}} & \ensuremath{\underline{0.098}_{{\textcolor{gray}{\pm 0.010}}}}  \\
\midrule
\multicolumn{11}{l}{\textbf{\textit{TabDPT}}} \\
\midrule
\hspace{3mm} Full AWARE & \ensuremath{{\textbf{0.925}}^{**}_{{\textcolor{gray}{\pm 0.000}}}} & \ensuremath{{\textbf{0.555}}^{**}_{{\textcolor{gray}{\pm 0.001}}}} & \ensuremath{{\textbf{0.802}}^{**}_{{\textcolor{gray}{\pm 0.001}}}} & \ensuremath{\textbf{0.288}_{{\textcolor{gray}{\pm 0.004}}}} & \ensuremath{{\textbf{0.674}}^{**}_{{\textcolor{gray}{\pm 0.001}}}} & \ensuremath{\textbf{0.266}_{{\textcolor{gray}{\pm 0.003}}}} & \ensuremath{{\textbf{0.990}}^{**}_{{\textcolor{gray}{\pm 0.001}}}} & \ensuremath{{\textbf{0.738}}^{**}_{{\textcolor{gray}{\pm 0.005}}}} & \ensuremath{{\textbf{0.935}}^{**}_{{\textcolor{gray}{\pm 0.002}}}} & \ensuremath{{\textbf{0.160}}^{**}_{{\textcolor{gray}{\pm 0.004}}}}  \\
\hspace{3mm} Attention Only & \ensuremath{0.893_{{\textcolor{gray}{\pm 0.001}}}} & \ensuremath{0.515_{{\textcolor{gray}{\pm 0.005}}}} & \ensuremath{\underline{0.789}_{{\textcolor{gray}{\pm 0.003}}}} & \ensuremath{0.281_{{\textcolor{gray}{\pm 0.006}}}} & \ensuremath{0.659_{{\textcolor{gray}{\pm 0.001}}}} & \ensuremath{0.258_{{\textcolor{gray}{\pm 0.003}}}} & \ensuremath{0.968_{{\textcolor{gray}{\pm 0.001}}}} & \ensuremath{0.688_{{\textcolor{gray}{\pm 0.008}}}} & \ensuremath{0.889_{{\textcolor{gray}{\pm 0.004}}}} & \ensuremath{0.107_{{\textcolor{gray}{\pm 0.003}}}}  \\
\hspace{3mm} SNNL Only & \ensuremath{0.906_{{\textcolor{gray}{\pm 0.000}}}} & \ensuremath{0.529_{{\textcolor{gray}{\pm 0.000}}}} & \ensuremath{0.785_{{\textcolor{gray}{\pm 0.000}}}} & \ensuremath{0.274_{{\textcolor{gray}{\pm 0.000}}}} & \ensuremath{0.659_{{\textcolor{gray}{\pm 0.000}}}} & \ensuremath{0.257_{{\textcolor{gray}{\pm 0.000}}}} & \ensuremath{0.967_{{\textcolor{gray}{\pm 0.001}}}} & \ensuremath{0.689_{{\textcolor{gray}{\pm 0.006}}}} & \ensuremath{0.892_{{\textcolor{gray}{\pm 0.003}}}} & \ensuremath{0.112_{{\textcolor{gray}{\pm 0.005}}}}  \\
\hspace{3mm} Balance Only & \ensuremath{0.907_{{\textcolor{gray}{\pm 0.001}}}} & \ensuremath{\underline{0.543}_{{\textcolor{gray}{\pm 0.001}}}} & \ensuremath{0.787_{{\textcolor{gray}{\pm 0.002}}}} & \ensuremath{0.280_{{\textcolor{gray}{\pm 0.004}}}} & \ensuremath{\underline{0.661}_{{\textcolor{gray}{\pm 0.002}}}} & \ensuremath{0.257_{{\textcolor{gray}{\pm 0.002}}}} & \ensuremath{0.965_{{\textcolor{gray}{\pm 0.001}}}} & \ensuremath{0.682_{{\textcolor{gray}{\pm 0.005}}}} & \ensuremath{0.890_{{\textcolor{gray}{\pm 0.001}}}} & \ensuremath{\underline{0.113}_{{\textcolor{gray}{\pm 0.005}}}}  \\
\hspace{3mm} Ensemble Only & \ensuremath{\underline{0.908}_{{\textcolor{gray}{\pm 0.002}}}} & \ensuremath{0.537_{{\textcolor{gray}{\pm 0.004}}}} & \ensuremath{\underline{0.789}_{{\textcolor{gray}{\pm 0.001}}}} & \ensuremath{\underline{0.284}_{{\textcolor{gray}{\pm 0.005}}}} & \ensuremath{\underline{0.661}_{{\textcolor{gray}{\pm 0.001}}}} & \ensuremath{\underline{0.260}_{{\textcolor{gray}{\pm 0.004}}}} & \ensuremath{\underline{0.969}_{{\textcolor{gray}{\pm 0.001}}}} & \ensuremath{\underline{0.692}_{{\textcolor{gray}{\pm 0.003}}}} & \ensuremath{\underline{0.895}_{{\textcolor{gray}{\pm 0.000}}}} & \ensuremath{0.112_{{\textcolor{gray}{\pm 0.001}}}}  \\
\hspace{3mm} Adapter Only & \ensuremath{0.903_{{\textcolor{gray}{\pm 0.003}}}} & \ensuremath{0.517_{{\textcolor{gray}{\pm 0.001}}}} & \ensuremath{0.773_{{\textcolor{gray}{\pm 0.002}}}} & \ensuremath{0.264_{{\textcolor{gray}{\pm 0.006}}}} & \ensuremath{0.631_{{\textcolor{gray}{\pm 0.004}}}} & \ensuremath{0.237_{{\textcolor{gray}{\pm 0.005}}}} & \ensuremath{0.943_{{\textcolor{gray}{\pm 0.003}}}} & \ensuremath{0.583_{{\textcolor{gray}{\pm 0.007}}}} & \ensuremath{0.848_{{\textcolor{gray}{\pm 0.013}}}} & \ensuremath{0.073_{{\textcolor{gray}{\pm 0.007}}}}  \\

\bottomrule
\end{tabular}}
\end{table*}

\begin{table*}[ht]
\centering
\caption{Ablation study of AWARE on large-scale ICU EHR tasks (MIMIC-IV and eICU). \highlightfour{Results are reported as AUROC and AUPRC (mean ± std over seeds) across clinical prediction tasks. $^\dagger$ Full AWARE version for KNN does not include Adapter Finetuning. Markers ($^{*}$p$<$0.05, $^{**}$p$<$0.01) on Full AWARE indicate it significantly outperforms all ablation variants (paired bootstrap, Holm-Bonferroni).}}
\label{tab:ablation_aware_wo}
\resizebox{\textwidth}{!}{
\renewcommand{\arraystretch}{0.95}
\begin{tabular}{lcccccccccc}
\toprule
\multirow{3}{*}{\textbf{Model}} 
& \multicolumn{6}{c}{\textbf{MIMIC-IV}} 
& \multicolumn{4}{c}{\textbf{eICU}} \\
\cmidrule(lr){2-7} \cmidrule(lr){8-11}
& \multicolumn{2}{c}{\textbf{SEPSIS}} 
& \multicolumn{2}{c}{\textbf{VAP}}
& \multicolumn{2}{c}{\textbf{UTI}} 
& \multicolumn{2}{c}{\textbf{SEPSIS}}
& \multicolumn{2}{c}{\textbf{UTI}} \\
\cmidrule(lr){2-3} \cmidrule(lr){4-5} \cmidrule(lr){6-7}
\cmidrule(lr){8-9} \cmidrule(lr){10-11} 
& AUROC $\uparrow$ & AUPRC $\uparrow$
& AUROC $\uparrow$ & AUPRC $\uparrow$
& AUROC $\uparrow$ & AUPRC $\uparrow$
& AUROC $\uparrow$ & AUPRC $\uparrow$
& AUROC $\uparrow$ & AUPRC $\uparrow$ \\
\midrule
\multicolumn{11}{l}{\textbf{\textit{KNN}}} \\
\midrule
\hspace{3mm} Full AWARE$^\dagger$ & \ensuremath{{\textbf{0.903}}^{**}_{{\textcolor{gray}{\pm 0.001}}}} & \ensuremath{{\textbf{0.507}}^{**}_{{\textcolor{gray}{\pm 0.006}}}} & \ensuremath{{\textbf{0.788}}^{**}_{{\textcolor{gray}{\pm 0.003}}}} & \ensuremath{0.250_{{\textcolor{gray}{\pm 0.005}}}} & \ensuremath{{\textbf{0.671}}^{**}_{{\textcolor{gray}{\pm 0.002}}}} & \ensuremath{{\textbf{0.260}}^{**}_{{\textcolor{gray}{\pm 0.004}}}} & \ensuremath{{\textbf{0.971}}^{**}_{{\textcolor{gray}{\pm 0.001}}}} & \ensuremath{{\textbf{0.711}}^{**}_{{\textcolor{gray}{\pm 0.002}}}} & \ensuremath{{\textbf{0.928}}^{**}_{{\textcolor{gray}{\pm 0.002}}}} & \ensuremath{{\textbf{0.155}}^{**}_{{\textcolor{gray}{\pm 0.003}}}}  \\
\hspace{3mm} w/o Attention & \ensuremath{0.864_{{\textcolor{gray}{\pm 0.002}}}} & \ensuremath{0.440_{{\textcolor{gray}{\pm 0.003}}}} & \ensuremath{0.773_{{\textcolor{gray}{\pm 0.001}}}} & \ensuremath{0.258_{{\textcolor{gray}{\pm 0.004}}}} & \ensuremath{0.632_{{\textcolor{gray}{\pm 0.002}}}} & \ensuremath{0.236_{{\textcolor{gray}{\pm 0.005}}}} & \ensuremath{0.861_{{\textcolor{gray}{\pm 0.004}}}} & \ensuremath{0.348_{{\textcolor{gray}{\pm 0.007}}}} & \ensuremath{0.827_{{\textcolor{gray}{\pm 0.003}}}} & \ensuremath{0.071_{{\textcolor{gray}{\pm 0.004}}}}  \\
\hspace{3mm} w/o SNNL & \ensuremath{\underline{0.887}_{{\textcolor{gray}{\pm 0.001}}}} & \ensuremath{\underline{0.486}_{{\textcolor{gray}{\pm 0.006}}}} & \ensuremath{\underline{0.780}_{{\textcolor{gray}{\pm 0.001}}}} & \ensuremath{\textbf{0.266}_{{\textcolor{gray}{\pm 0.005}}}} & \ensuremath{\underline{0.640}_{{\textcolor{gray}{\pm 0.002}}}} & \ensuremath{\underline{0.241}_{{\textcolor{gray}{\pm 0.004}}}} & \ensuremath{\underline{0.946}_{{\textcolor{gray}{\pm 0.002}}}} & \ensuremath{\underline{0.604}_{{\textcolor{gray}{\pm 0.007}}}} & \ensuremath{\underline{0.869}_{{\textcolor{gray}{\pm 0.001}}}} & \ensuremath{\underline{0.091}_{{\textcolor{gray}{\pm 0.003}}}}  \\
\hspace{3mm} w/o Bl. Sampl. & \ensuremath{0.860_{{\textcolor{gray}{\pm 0.006}}}} & \ensuremath{0.436_{{\textcolor{gray}{\pm 0.008}}}} & \ensuremath{0.771_{{\textcolor{gray}{\pm 0.002}}}} & \ensuremath{\underline{0.259}_{{\textcolor{gray}{\pm 0.006}}}} & \ensuremath{\underline{0.640}_{{\textcolor{gray}{\pm 0.002}}}} & \ensuremath{0.240_{{\textcolor{gray}{\pm 0.002}}}} & \ensuremath{0.865_{{\textcolor{gray}{\pm 0.004}}}} & \ensuremath{0.353_{{\textcolor{gray}{\pm 0.009}}}} & \ensuremath{0.827_{{\textcolor{gray}{\pm 0.004}}}} & \ensuremath{0.073_{{\textcolor{gray}{\pm 0.002}}}}  \\
\hspace{3mm} w/o Ensemble & \ensuremath{0.859_{{\textcolor{gray}{\pm 0.002}}}} & \ensuremath{0.427_{{\textcolor{gray}{\pm 0.005}}}} & \ensuremath{0.766_{{\textcolor{gray}{\pm 0.003}}}} & \ensuremath{0.251_{{\textcolor{gray}{\pm 0.005}}}} & \ensuremath{\underline{0.640}_{{\textcolor{gray}{\pm 0.009}}}} & \ensuremath{0.237_{{\textcolor{gray}{\pm 0.005}}}} & \ensuremath{0.896_{{\textcolor{gray}{\pm 0.002}}}} & \ensuremath{0.430_{{\textcolor{gray}{\pm 0.017}}}} & \ensuremath{0.857_{{\textcolor{gray}{\pm 0.002}}}} & \ensuremath{0.083_{{\textcolor{gray}{\pm 0.002}}}}  \\
\midrule
\multicolumn{11}{l}{\textbf{\textit{kNNPFN}}} \\
\midrule
\hspace{3mm} Full AWARE & \ensuremath{{\textbf{0.917}}^{**}_{{\textcolor{gray}{\pm 0.001}}}} & \ensuremath{{\textbf{0.538}}^{**}_{{\textcolor{gray}{\pm 0.006}}}} & \ensuremath{{\textbf{0.799}}^{**}_{{\textcolor{gray}{\pm 0.001}}}} & \ensuremath{\textbf{0.282}_{{\textcolor{gray}{\pm 0.001}}}} & \ensuremath{{\textbf{0.673}}^{**}_{{\textcolor{gray}{\pm 0.002}}}} & \ensuremath{\textbf{0.263}_{{\textcolor{gray}{\pm 0.006}}}} & \ensuremath{{\textbf{0.981}}^{**}_{{\textcolor{gray}{\pm 0.001}}}} & \ensuremath{{\textbf{0.724}}^{**}_{{\textcolor{gray}{\pm 0.004}}}} & \ensuremath{{\textbf{0.932}}^{**}_{{\textcolor{gray}{\pm 0.002}}}} & \ensuremath{{\textbf{0.152}}^{**}_{{\textcolor{gray}{\pm 0.005}}}}  \\
\hspace{3mm} w/o Attention & \ensuremath{0.900_{{\textcolor{gray}{\pm 0.001}}}} & \ensuremath{0.520_{{\textcolor{gray}{\pm 0.009}}}} & \ensuremath{0.784_{{\textcolor{gray}{\pm 0.003}}}} & \ensuremath{0.275_{{\textcolor{gray}{\pm 0.010}}}} & \ensuremath{0.656_{{\textcolor{gray}{\pm 0.002}}}} & \ensuremath{0.251_{{\textcolor{gray}{\pm 0.003}}}} & \ensuremath{0.954_{{\textcolor{gray}{\pm 0.001}}}} & \ensuremath{0.634_{{\textcolor{gray}{\pm 0.002}}}} & \ensuremath{0.875_{{\textcolor{gray}{\pm 0.006}}}} & \ensuremath{0.095_{{\textcolor{gray}{\pm 0.005}}}}  \\
\hspace{3mm} w/o SNNL & \ensuremath{0.899_{{\textcolor{gray}{\pm 0.001}}}} & \ensuremath{0.514_{{\textcolor{gray}{\pm 0.005}}}} & \ensuremath{0.780_{{\textcolor{gray}{\pm 0.002}}}} & \ensuremath{0.262_{{\textcolor{gray}{\pm 0.006}}}} & \ensuremath{0.656_{{\textcolor{gray}{\pm 0.001}}}} & \ensuremath{0.251_{{\textcolor{gray}{\pm 0.002}}}} & \ensuremath{0.958_{{\textcolor{gray}{\pm 0.001}}}} & \ensuremath{0.650_{{\textcolor{gray}{\pm 0.003}}}} & \ensuremath{0.885_{{\textcolor{gray}{\pm 0.001}}}} & \ensuremath{0.091_{{\textcolor{gray}{\pm 0.004}}}}  \\
\hspace{3mm} w/o Bl. Sampl. & \ensuremath{0.900_{{\textcolor{gray}{\pm 0.001}}}} & \ensuremath{0.520_{{\textcolor{gray}{\pm 0.003}}}} & \ensuremath{0.780_{{\textcolor{gray}{\pm 0.001}}}} & \ensuremath{0.269_{{\textcolor{gray}{\pm 0.001}}}} & \ensuremath{0.657_{{\textcolor{gray}{\pm 0.002}}}} & \ensuremath{0.253_{{\textcolor{gray}{\pm 0.002}}}} & \ensuremath{0.958_{{\textcolor{gray}{\pm 0.001}}}} & \ensuremath{0.641_{{\textcolor{gray}{\pm 0.003}}}} & \ensuremath{0.875_{{\textcolor{gray}{\pm 0.003}}}} & \ensuremath{\underline{0.098}_{{\textcolor{gray}{\pm 0.002}}}}  \\
\hspace{3mm} w/o Ensemble & \ensuremath{0.896_{{\textcolor{gray}{\pm 0.003}}}} & \ensuremath{0.515_{{\textcolor{gray}{\pm 0.009}}}} & \ensuremath{0.783_{{\textcolor{gray}{\pm 0.004}}}} & \ensuremath{0.269_{{\textcolor{gray}{\pm 0.002}}}} & \ensuremath{0.657_{{\textcolor{gray}{\pm 0.002}}}} & \ensuremath{0.254_{{\textcolor{gray}{\pm 0.001}}}} & \ensuremath{0.955_{{\textcolor{gray}{\pm 0.001}}}} & \ensuremath{0.641_{{\textcolor{gray}{\pm 0.003}}}} & \ensuremath{0.880_{{\textcolor{gray}{\pm 0.005}}}} & \ensuremath{0.095_{{\textcolor{gray}{\pm 0.005}}}}  \\
\hspace{3mm} w/o Adapter & \ensuremath{\underline{0.913}_{{\textcolor{gray}{\pm 0.001}}}} & \ensuremath{\underline{0.533}_{{\textcolor{gray}{\pm 0.006}}}} & \ensuremath{\underline{0.795}_{{\textcolor{gray}{\pm 0.001}}}} & \ensuremath{\underline{0.281}_{{\textcolor{gray}{\pm 0.001}}}} & \ensuremath{\underline{0.669}_{{\textcolor{gray}{\pm 0.001}}}} & \ensuremath{\underline{0.259}_{{\textcolor{gray}{\pm 0.003}}}} & \ensuremath{\underline{0.976}_{{\textcolor{gray}{\pm 0.001}}}} & \ensuremath{\underline{0.720}_{{\textcolor{gray}{\pm 0.004}}}} & \ensuremath{\underline{0.927}_{{\textcolor{gray}{\pm 0.002}}}} & \ensuremath{\textbf{0.152}_{{\textcolor{gray}{\pm 0.005}}}}  \\
\midrule
\multicolumn{11}{l}{\textbf{\textit{TabDPT}}} \\
\midrule
\hspace{3mm} Full AWARE & \ensuremath{{\textbf{0.925}}^{**}_{{\textcolor{gray}{\pm 0.001}}}} & \ensuremath{{\textbf{0.555}}^{**}_{{\textcolor{gray}{\pm 0.001}}}} & \ensuremath{{\textbf{0.802}}^{**}_{{\textcolor{gray}{\pm 0.001}}}} & \ensuremath{\textbf{0.288}_{{\textcolor{gray}{\pm 0.004}}}} & \ensuremath{{\textbf{0.674}}^{**}_{{\textcolor{gray}{\pm 0.001}}}} & \ensuremath{\textbf{0.266}_{{\textcolor{gray}{\pm 0.003}}}} & \ensuremath{{\textbf{0.990}}^{**}_{{\textcolor{gray}{\pm 0.001}}}} & \ensuremath{{\textbf{0.738}}^{**}_{{\textcolor{gray}{\pm 0.005}}}} & \ensuremath{{\textbf{0.935}}^{**}_{{\textcolor{gray}{\pm 0.002}}}} & \ensuremath{{\textbf{0.160}}^{**}_{{\textcolor{gray}{\pm 0.004}}}}  \\
\hspace{3mm} w/o Attention & \ensuremath{0.908_{{\textcolor{gray}{\pm 0.001}}}} & \ensuremath{0.539_{{\textcolor{gray}{\pm 0.003}}}} & \ensuremath{0.791_{{\textcolor{gray}{\pm 0.002}}}} & \ensuremath{\underline{0.284}_{{\textcolor{gray}{\pm 0.001}}}} & \ensuremath{0.661_{{\textcolor{gray}{\pm 0.002}}}} & \ensuremath{0.259_{{\textcolor{gray}{\pm 0.001}}}} & \ensuremath{0.965_{{\textcolor{gray}{\pm 0.001}}}} & \ensuremath{0.685_{{\textcolor{gray}{\pm 0.003}}}} & \ensuremath{0.885_{{\textcolor{gray}{\pm 0.003}}}} & \ensuremath{0.109_{{\textcolor{gray}{\pm 0.003}}}}  \\
\hspace{3mm} w/o SNNL & \ensuremath{0.907_{{\textcolor{gray}{\pm 0.002}}}} & \ensuremath{0.541_{{\textcolor{gray}{\pm 0.006}}}} & \ensuremath{0.789_{{\textcolor{gray}{\pm 0.002}}}} & \ensuremath{0.279_{{\textcolor{gray}{\pm 0.003}}}} & \ensuremath{0.661_{{\textcolor{gray}{\pm 0.002}}}} & \ensuremath{0.259_{{\textcolor{gray}{\pm 0.002}}}} & \ensuremath{0.974_{{\textcolor{gray}{\pm 0.001}}}} & \ensuremath{0.723_{{\textcolor{gray}{\pm 0.005}}}} & \ensuremath{0.900_{{\textcolor{gray}{\pm 0.001}}}} & \ensuremath{0.118_{{\textcolor{gray}{\pm 0.003}}}}  \\
\hspace{3mm} w/o Bl. Sampl. & \ensuremath{0.908_{{\textcolor{gray}{\pm 0.001}}}} & \ensuremath{0.542_{{\textcolor{gray}{\pm 0.008}}}} & \ensuremath{0.786_{{\textcolor{gray}{\pm 0.001}}}} & \ensuremath{0.282_{{\textcolor{gray}{\pm 0.004}}}} & \ensuremath{0.661_{{\textcolor{gray}{\pm 0.002}}}} & \ensuremath{0.259_{{\textcolor{gray}{\pm 0.003}}}} & \ensuremath{0.969_{{\textcolor{gray}{\pm 0.001}}}} & \ensuremath{0.688_{{\textcolor{gray}{\pm 0.002}}}} & \ensuremath{0.898_{{\textcolor{gray}{\pm 0.001}}}} & \ensuremath{0.120_{{\textcolor{gray}{\pm 0.006}}}}  \\
\hspace{3mm} w/o Ensemble & \ensuremath{0.905_{{\textcolor{gray}{\pm 0.004}}}} & \ensuremath{0.534_{{\textcolor{gray}{\pm 0.006}}}} & \ensuremath{0.788_{{\textcolor{gray}{\pm 0.002}}}} & \ensuremath{0.282_{{\textcolor{gray}{\pm 0.004}}}} & \ensuremath{0.660_{{\textcolor{gray}{\pm 0.004}}}} & \ensuremath{0.261_{{\textcolor{gray}{\pm 0.003}}}} & \ensuremath{0.966_{{\textcolor{gray}{\pm 0.001}}}} & \ensuremath{0.690_{{\textcolor{gray}{\pm 0.002}}}} & \ensuremath{0.892_{{\textcolor{gray}{\pm 0.006}}}} & \ensuremath{0.114_{{\textcolor{gray}{\pm 0.006}}}}  \\
\hspace{3mm} w/o Adapter & \ensuremath{\underline{0.920}_{{\textcolor{gray}{\pm 0.001}}}} & \ensuremath{\underline{0.550}_{{\textcolor{gray}{\pm 0.004}}}} & \ensuremath{\underline{0.798}_{{\textcolor{gray}{\pm 0.001}}}} & \ensuremath{0.281_{{\textcolor{gray}{\pm 0.003}}}} & \ensuremath{\underline{0.671}_{{\textcolor{gray}{\pm 0.001}}}} & \ensuremath{\underline{0.263}_{{\textcolor{gray}{\pm 0.001}}}} & \ensuremath{\underline{0.985}_{{\textcolor{gray}{\pm 0.001}}}} & \ensuremath{\underline{0.734}_{{\textcolor{gray}{\pm 0.005}}}} & \ensuremath{\underline{0.931}_{{\textcolor{gray}{\pm 0.002}}}} & \ensuremath{\underline{0.159}_{{\textcolor{gray}{\pm 0.004}}}}  \\

\bottomrule
\end{tabular}}
\end{table*}

\begin{table*}[t]
\centering
\caption{Ablation study of AWARE and subsequent enhancements on large-scale ICU EHR tasks (MIMIC-IV and eICU).
Each row progressively adds one component to the backbone. We report AUROC and AUPRC for prediction tasks (mean ± std across 3 seeds). Markers ($^{*}$p$<$0.05, $^{**}$p$<$0.01) on + Adapter (Full AWARE) indicate it significantly outperforms all progressive variants (paired bootstrap, Holm-Bonferroni).}
\label{tab:ablation_aware_progressive}
\resizebox{\textwidth}{!}{
\renewcommand{\arraystretch}{0.95}
\begin{tabular}{lcccccccccc}
\toprule
\multirow{3}{*}{\textbf{Model}} 
& \multicolumn{6}{c}{\textbf{MIMIC-IV}} 
& \multicolumn{4}{c}{\textbf{eICU}} \\
\cmidrule(lr){2-7} \cmidrule(lr){8-11}
& \multicolumn{2}{c}{\textbf{SEPSIS}} 
& \multicolumn{2}{c}{\textbf{VAP}}
& \multicolumn{2}{c}{\textbf{UTI}} 
& \multicolumn{2}{c}{\textbf{SEPSIS}}
& \multicolumn{2}{c}{\textbf{UTI}} \\
\cmidrule(lr){2-3} \cmidrule(lr){4-5} \cmidrule(lr){6-7}
\cmidrule(lr){8-9} \cmidrule(lr){10-11} 
& AUROC $\uparrow$ & AUPRC $\uparrow$
& AUROC $\uparrow$ & AUPRC $\uparrow$
& AUROC $\uparrow$ & AUPRC $\uparrow$
& AUROC $\uparrow$ & AUPRC $\uparrow$
& AUROC $\uparrow$ & AUPRC $\uparrow$ \\
\midrule
\multicolumn{11}{l}{\textbf{\textit{KNN}}} \\
\midrule
\hspace{3mm} Baseline & \ensuremath{0.794_{{\textcolor{gray}{\pm 0.013}}}} & \ensuremath{0.391_{{\textcolor{gray}{\pm 0.026}}}} & \ensuremath{0.709_{{\textcolor{gray}{\pm 0.015}}}} & \ensuremath{0.213_{{\textcolor{gray}{\pm 0.008}}}} & \ensuremath{0.569_{{\textcolor{gray}{\pm 0.025}}}} & \ensuremath{0.209_{{\textcolor{gray}{\pm 0.025}}}} & \ensuremath{0.795_{{\textcolor{gray}{\pm 0.023}}}} & \ensuremath{0.314_{{\textcolor{gray}{\pm 0.027}}}} & \ensuremath{0.734_{{\textcolor{gray}{\pm 0.041}}}} & \ensuremath{0.056_{{\textcolor{gray}{\pm 0.011}}}}  \\
\hspace{3mm} + SNNL & \ensuremath{0.795_{{\textcolor{gray}{\pm 0.002}}}} & \ensuremath{0.394_{{\textcolor{gray}{\pm 0.011}}}} & \ensuremath{0.710_{{\textcolor{gray}{\pm 0.005}}}} & \ensuremath{0.233_{{\textcolor{gray}{\pm 0.011}}}} & \ensuremath{0.586_{{\textcolor{gray}{\pm 0.005}}}} & \ensuremath{0.218_{{\textcolor{gray}{\pm 0.007}}}} & \ensuremath{0.803_{{\textcolor{gray}{\pm 0.006}}}} & \ensuremath{0.325_{{\textcolor{gray}{\pm 0.018}}}} & \ensuremath{0.757_{{\textcolor{gray}{\pm 0.006}}}} & \ensuremath{0.062_{{\textcolor{gray}{\pm 0.001}}}}  \\
\hspace{3mm} + Attention & \ensuremath{0.843_{{\textcolor{gray}{\pm 0.001}}}} & \ensuremath{\underline{0.503}_{{\textcolor{gray}{\pm 0.007}}}} & \ensuremath{0.736_{{\textcolor{gray}{\pm 0.002}}}} & \ensuremath{0.247_{{\textcolor{gray}{\pm 0.007}}}} & \ensuremath{0.622_{{\textcolor{gray}{\pm 0.001}}}} & \ensuremath{0.246_{{\textcolor{gray}{\pm 0.003}}}} & \ensuremath{0.853_{{\textcolor{gray}{\pm 0.005}}}} & \ensuremath{0.467_{{\textcolor{gray}{\pm 0.009}}}} & \ensuremath{0.715_{{\textcolor{gray}{\pm 0.030}}}} & \ensuremath{0.048_{{\textcolor{gray}{\pm 0.014}}}}  \\
\hspace{3mm} + Bl. Sampl. & \ensuremath{\underline{0.860}_{{\textcolor{gray}{\pm 0.001}}}} & \ensuremath{0.502_{{\textcolor{gray}{\pm 0.007}}}} & \ensuremath{\underline{0.754}_{{\textcolor{gray}{\pm 0.004}}}} & \ensuremath{\textbf{0.257}_{{\textcolor{gray}{\pm 0.002}}}} & \ensuremath{\underline{0.635}_{{\textcolor{gray}{\pm 0.003}}}} & \ensuremath{\underline{0.248}_{{\textcolor{gray}{\pm 0.001}}}} & \ensuremath{\underline{0.922}_{{\textcolor{gray}{\pm 0.001}}}} & \ensuremath{\underline{0.681}_{{\textcolor{gray}{\pm 0.002}}}} & \ensuremath{\underline{0.875}_{{\textcolor{gray}{\pm 0.003}}}} & \ensuremath{\underline{0.139}_{{\textcolor{gray}{\pm 0.003}}}}  \\
\hspace{3mm} + Ensemble & \ensuremath{{\textbf{0.903}}^{**}_{{\textcolor{gray}{\pm 0.001}}}} & \ensuremath{\textbf{0.507}_{{\textcolor{gray}{\pm 0.006}}}} & \ensuremath{{\textbf{0.788}}^{**}_{{\textcolor{gray}{\pm 0.003}}}} & \ensuremath{\underline{0.250}_{{\textcolor{gray}{\pm 0.005}}}} & \ensuremath{{\textbf{0.671}}^{**}_{{\textcolor{gray}{\pm 0.002}}}} & \ensuremath{{\textbf{0.260}}^{**}_{{\textcolor{gray}{\pm 0.004}}}} & \ensuremath{{\textbf{0.971}}^{**}_{{\textcolor{gray}{\pm 0.001}}}} & \ensuremath{{\textbf{0.711}}^{**}_{{\textcolor{gray}{\pm 0.002}}}} & \ensuremath{{\textbf{0.928}}^{**}_{{\textcolor{gray}{\pm 0.002}}}} & \ensuremath{{\textbf{0.155}}^{**}_{{\textcolor{gray}{\pm 0.003}}}}  \\
\midrule
\multicolumn{11}{l}{\textbf{\textit{kNNPFN}}} \\
\midrule
\hspace{3mm} Baseline & \ensuremath{0.897_{{\textcolor{gray}{\pm 0.001}}}} & \ensuremath{0.509_{{\textcolor{gray}{\pm 0.001}}}} & \ensuremath{0.782_{{\textcolor{gray}{\pm 0.001}}}} & \ensuremath{0.276_{{\textcolor{gray}{\pm 0.001}}}} & \ensuremath{0.657_{{\textcolor{gray}{\pm 0.001}}}} & \ensuremath{0.253_{{\textcolor{gray}{\pm 0.001}}}} & \ensuremath{0.947_{{\textcolor{gray}{\pm 0.001}}}} & \ensuremath{0.615_{{\textcolor{gray}{\pm 0.001}}}} & \ensuremath{0.867_{{\textcolor{gray}{\pm 0.001}}}} & \ensuremath{0.091_{{\textcolor{gray}{\pm 0.001}}}}  \\
\hspace{3mm} + SNNL & \ensuremath{0.898_{{\textcolor{gray}{\pm 0.001}}}} & \ensuremath{0.510_{{\textcolor{gray}{\pm 0.001}}}} & \ensuremath{0.782_{{\textcolor{gray}{\pm 0.001}}}} & \ensuremath{0.277_{{\textcolor{gray}{\pm 0.001}}}} & \ensuremath{0.658_{{\textcolor{gray}{\pm 0.001}}}} & \ensuremath{0.254_{{\textcolor{gray}{\pm 0.001}}}} & \ensuremath{0.957_{{\textcolor{gray}{\pm 0.001}}}} & \ensuremath{0.638_{{\textcolor{gray}{\pm 0.002}}}} & \ensuremath{0.873_{{\textcolor{gray}{\pm 0.004}}}} & \ensuremath{0.093_{{\textcolor{gray}{\pm 0.002}}}}  \\
\hspace{3mm} + Attention & \ensuremath{0.900_{{\textcolor{gray}{\pm 0.001}}}} & \ensuremath{0.511_{{\textcolor{gray}{\pm 0.001}}}} & \ensuremath{0.784_{{\textcolor{gray}{\pm 0.001}}}} & \ensuremath{0.277_{{\textcolor{gray}{\pm 0.001}}}} & \ensuremath{0.659_{{\textcolor{gray}{\pm 0.001}}}} & \ensuremath{0.254_{{\textcolor{gray}{\pm 0.001}}}} & \ensuremath{0.962_{{\textcolor{gray}{\pm 0.001}}}} & \ensuremath{0.654_{{\textcolor{gray}{\pm 0.007}}}} & \ensuremath{0.879_{{\textcolor{gray}{\pm 0.004}}}} & \ensuremath{0.093_{{\textcolor{gray}{\pm 0.002}}}}  \\
\hspace{3mm} + Bl. Sampl. & \ensuremath{0.906_{{\textcolor{gray}{\pm 0.001}}}} & \ensuremath{0.531_{{\textcolor{gray}{\pm 0.006}}}} & \ensuremath{0.789_{{\textcolor{gray}{\pm 0.001}}}} & \ensuremath{0.279_{{\textcolor{gray}{\pm 0.001}}}} & \ensuremath{0.664_{{\textcolor{gray}{\pm 0.001}}}} & \ensuremath{0.258_{{\textcolor{gray}{\pm 0.004}}}} & \ensuremath{0.971_{{\textcolor{gray}{\pm 0.001}}}} & \ensuremath{0.716_{{\textcolor{gray}{\pm 0.004}}}} & \ensuremath{0.919_{{\textcolor{gray}{\pm 0.002}}}} & \ensuremath{\underline{0.146}_{{\textcolor{gray}{\pm 0.004}}}}  \\
\hspace{3mm} + Ensemble & \ensuremath{\underline{0.913}_{{\textcolor{gray}{\pm 0.001}}}} & \ensuremath{\underline{0.533}_{{\textcolor{gray}{\pm 0.006}}}} & \ensuremath{\underline{0.795}_{{\textcolor{gray}{\pm 0.001}}}} & \ensuremath{\underline{0.281}_{{\textcolor{gray}{\pm 0.001}}}} & \ensuremath{\underline{0.669}_{{\textcolor{gray}{\pm 0.001}}}} & \ensuremath{\underline{0.259}_{{\textcolor{gray}{\pm 0.003}}}} & \ensuremath{\underline{0.976}_{{\textcolor{gray}{\pm 0.001}}}} & \ensuremath{\underline{0.720}_{{\textcolor{gray}{\pm 0.004}}}} & \ensuremath{\underline{0.927}_{{\textcolor{gray}{\pm 0.002}}}} & \ensuremath{\textbf{0.152}_{{\textcolor{gray}{\pm 0.005}}}}  \\
\hspace{3mm} + Adapter & \ensuremath{{\textbf{0.917}}^{**}_{{\textcolor{gray}{\pm 0.001}}}} & \ensuremath{{\textbf{0.538}}^{**}_{{\textcolor{gray}{\pm 0.006}}}} & \ensuremath{{\textbf{0.799}}^{**}_{{\textcolor{gray}{\pm 0.001}}}} & \ensuremath{{\textbf{0.282}}^{**}_{{\textcolor{gray}{\pm 0.001}}}} & \ensuremath{{\textbf{0.673}}^{**}_{{\textcolor{gray}{\pm 0.002}}}} & \ensuremath{\textbf{0.263}_{{\textcolor{gray}{\pm 0.006}}}} & \ensuremath{{\textbf{0.981}}^{**}_{{\textcolor{gray}{\pm 0.001}}}} & \ensuremath{{\textbf{0.724}}^{**}_{{\textcolor{gray}{\pm 0.004}}}} & \ensuremath{{\textbf{0.932}}^{**}_{{\textcolor{gray}{\pm 0.002}}}} & \ensuremath{\textbf{0.152}_{{\textcolor{gray}{\pm 0.005}}}}  \\
\midrule
\multicolumn{11}{l}{\textbf{\textit{TabDPT}}} \\
\midrule
\hspace{3mm} Baseline & \ensuremath{0.905_{{\textcolor{gray}{\pm 0.001}}}} & \ensuremath{0.528_{{\textcolor{gray}{\pm 0.001}}}} & \ensuremath{0.784_{{\textcolor{gray}{\pm 0.001}}}} & \ensuremath{0.274_{{\textcolor{gray}{\pm 0.001}}}} & \ensuremath{0.658_{{\textcolor{gray}{\pm 0.001}}}} & \ensuremath{0.257_{{\textcolor{gray}{\pm 0.001}}}} & \ensuremath{0.961_{{\textcolor{gray}{\pm 0.001}}}} & \ensuremath{0.659_{{\textcolor{gray}{\pm 0.001}}}} & \ensuremath{0.882_{{\textcolor{gray}{\pm 0.001}}}} & \ensuremath{0.103_{{\textcolor{gray}{\pm 0.001}}}}  \\
\hspace{3mm} + SNNL & \ensuremath{0.906_{{\textcolor{gray}{\pm 0.001}}}} & \ensuremath{0.529_{{\textcolor{gray}{\pm 0.001}}}} & \ensuremath{0.785_{{\textcolor{gray}{\pm 0.001}}}} & \ensuremath{0.274_{{\textcolor{gray}{\pm 0.001}}}} & \ensuremath{0.659_{{\textcolor{gray}{\pm 0.001}}}} & \ensuremath{0.257_{{\textcolor{gray}{\pm 0.001}}}} & \ensuremath{0.967_{{\textcolor{gray}{\pm 0.001}}}} & \ensuremath{0.689_{{\textcolor{gray}{\pm 0.006}}}} & \ensuremath{0.892_{{\textcolor{gray}{\pm 0.003}}}} & \ensuremath{0.112_{{\textcolor{gray}{\pm 0.005}}}}  \\
\hspace{3mm} + Attention & \ensuremath{0.908_{{\textcolor{gray}{\pm 0.001}}}} & \ensuremath{0.530_{{\textcolor{gray}{\pm 0.001}}}} & \ensuremath{0.787_{{\textcolor{gray}{\pm 0.001}}}} & \ensuremath{0.275_{{\textcolor{gray}{\pm 0.001}}}} & \ensuremath{0.660_{{\textcolor{gray}{\pm 0.001}}}} & \ensuremath{0.257_{{\textcolor{gray}{\pm 0.001}}}} & \ensuremath{0.973_{{\textcolor{gray}{\pm 0.001}}}} & \ensuremath{0.703_{{\textcolor{gray}{\pm 0.005}}}} & \ensuremath{0.898_{{\textcolor{gray}{\pm 0.003}}}} & \ensuremath{0.113_{{\textcolor{gray}{\pm 0.004}}}}  \\
\hspace{3mm} + Bl. Sampl. & \ensuremath{0.914_{{\textcolor{gray}{\pm 0.001}}}} & \ensuremath{0.546_{{\textcolor{gray}{\pm 0.004}}}} & \ensuremath{0.792_{{\textcolor{gray}{\pm 0.001}}}} & \ensuremath{0.279_{{\textcolor{gray}{\pm 0.003}}}} & \ensuremath{0.666_{{\textcolor{gray}{\pm 0.001}}}} & \ensuremath{0.262_{{\textcolor{gray}{\pm 0.001}}}} & \ensuremath{0.979_{{\textcolor{gray}{\pm 0.001}}}} & \ensuremath{0.730_{{\textcolor{gray}{\pm 0.005}}}} & \ensuremath{0.924_{{\textcolor{gray}{\pm 0.001}}}} & \ensuremath{0.151_{{\textcolor{gray}{\pm 0.006}}}}  \\
\hspace{3mm} + Ensemble & \ensuremath{\underline{0.920}_{{\textcolor{gray}{\pm 0.001}}}} & \ensuremath{\underline{0.550}_{{\textcolor{gray}{\pm 0.004}}}} & \ensuremath{\underline{0.798}_{{\textcolor{gray}{\pm 0.001}}}} & \ensuremath{\underline{0.281}_{{\textcolor{gray}{\pm 0.003}}}} & \ensuremath{\underline{0.671}_{{\textcolor{gray}{\pm 0.001}}}} & \ensuremath{\underline{0.263}_{{\textcolor{gray}{\pm 0.001}}}} & \ensuremath{\underline{0.985}_{{\textcolor{gray}{\pm 0.001}}}} & \ensuremath{\underline{0.734}_{{\textcolor{gray}{\pm 0.005}}}} & \ensuremath{\underline{0.931}_{{\textcolor{gray}{\pm 0.002}}}} & \ensuremath{\underline{0.159}_{{\textcolor{gray}{\pm 0.004}}}}  \\
\hspace{3mm} + Adapter & \ensuremath{{\textbf{0.925}}^{**}_{{\textcolor{gray}{\pm 0.001}}}} & \ensuremath{{\textbf{0.555}}^{**}_{{\textcolor{gray}{\pm 0.001}}}} & \ensuremath{{\textbf{0.802}}^{**}_{{\textcolor{gray}{\pm 0.001}}}} & \ensuremath{{\textbf{0.288}}^{**}_{{\textcolor{gray}{\pm 0.004}}}} & \ensuremath{{\textbf{0.674}}^{**}_{{\textcolor{gray}{\pm 0.001}}}} & \ensuremath{{\textbf{0.266}}^{*}_{{\textcolor{gray}{\pm 0.003}}}} & \ensuremath{{\textbf{0.990}}^{**}_{{\textcolor{gray}{\pm 0.001}}}} & \ensuremath{{\textbf{0.738}}^{**}_{{\textcolor{gray}{\pm 0.005}}}} & \ensuremath{{\textbf{0.935}}^{**}_{{\textcolor{gray}{\pm 0.002}}}} & \ensuremath{{\textbf{0.160}}^{**}_{{\textcolor{gray}{\pm 0.004}}}}  \\

\bottomrule
\end{tabular}}
\end{table*}

\newpage

\begin{table*}[t]
\centering
\caption{Benchmark coverage of models across small-to-medium scale EHR datasets, sourced from OpenML and UCI repositories. RMSE is reported for regression tasks, while AUROC is reported for other classification tasks. Markers ($^{*}$p$<$0.05, $^{**}$p$<$0.01) on AWARE models indicate they significantly outperform all non-AWARE baselines (paired bootstrap, Holm-Bonferroni).}
\label{tab:results}
\label{tab:smallmediumehrs}
\resizebox{\textwidth}{!}{
\renewcommand{\arraystretch}{0.95}
\begin{tabular}{lcccccccccccc}
\toprule
\multirow{2}{*}{\textbf{Model}} & \multicolumn{3}{c}{\textbf{Regression}} & \multicolumn{9}{c}{\textbf{Classification}} \\
\cmidrule(lr){2-4} \cmidrule(lr){5-13}
 & HF $\downarrow$ & OXF-PT $\downarrow$ & TIT $\downarrow$ & SUPPORT2 $\uparrow$ & SSMI $\uparrow$ & DTC $\uparrow$ & GGCM $\uparrow$ & AIDS $\uparrow$ & ILP $\uparrow$ & COVID19 $\uparrow$ & Diabetes $\uparrow$ & Kidney $\uparrow$ \\
\midrule
\multicolumn{13}{c}{\textbf{\textit{Classical and Boosting Models}}} \\
\midrule
LR & \ensuremath{1.360_{\textcolor{gray}{\pm 0.001}}} & \ensuremath{0.413_{\textcolor{gray}{\pm 0.001}}} & \ensuremath{1.070_{\textcolor{gray}{\pm 0.001}}} & \ensuremath{0.836_{\textcolor{gray}{\pm 0.001}}} & \ensuremath{0.692_{\textcolor{gray}{\pm 0.001}}} & \ensuremath{0.984_{\textcolor{gray}{\pm 0.001}}} & \ensuremath{0.883_{\textcolor{gray}{\pm 0.001}}} & \ensuremath{0.710_{\textcolor{gray}{\pm 0.001}}} & \ensuremath{0.757_{\textcolor{gray}{\pm 0.001}}} & \ensuremath{0.727_{\textcolor{gray}{\pm 0.001}}} & \ensuremath{0.630_{\textcolor{gray}{\pm 0.001}}} & \ensuremath{0.688_{\textcolor{gray}{\pm 0.001}}} \\
KNN & \ensuremath{1.394_{\textcolor{gray}{\pm 0.001}}} & \ensuremath{0.551_{\textcolor{gray}{\pm 0.001}}} & \ensuremath{1.253_{\textcolor{gray}{\pm 0.001}}} & \ensuremath{0.782_{\textcolor{gray}{\pm 0.001}}} & \ensuremath{0.645_{\textcolor{gray}{\pm 0.001}}} & \ensuremath{0.500_{\textcolor{gray}{\pm 0.001}}} & \ensuremath{0.500_{\textcolor{gray}{\pm 0.001}}} & \ensuremath{0.702_{\textcolor{gray}{\pm 0.001}}} & \ensuremath{0.500_{\textcolor{gray}{\pm 0.001}}} & \ensuremath{0.692_{\textcolor{gray}{\pm 0.001}}} & \ensuremath{0.479_{\textcolor{gray}{\pm 0.001}}} & \ensuremath{0.500_{\textcolor{gray}{\pm 0.001}}} \\
CatB & \ensuremath{1.366_{\textcolor{gray}{\pm 0.033}}} & \ensuremath{\underline{0.107}_{\textcolor{gray}{\pm 0.013}}} & \ensuremath{0.942_{\textcolor{gray}{\pm 0.096}}} & \ensuremath{0.839_{\textcolor{gray}{\pm 0.006}}} & \ensuremath{0.690_{\textcolor{gray}{\pm 0.002}}} & \ensuremath{0.972_{\textcolor{gray}{\pm 0.007}}} & \ensuremath{0.851_{\textcolor{gray}{\pm 0.013}}} & \ensuremath{0.714_{\textcolor{gray}{\pm 0.004}}} & \ensuremath{0.741_{\textcolor{gray}{\pm 0.006}}} & \ensuremath{0.782_{\textcolor{gray}{\pm 0.005}}} & \ensuremath{\underline{0.676}_{\textcolor{gray}{\pm 0.004}}} & \ensuremath{0.750_{\textcolor{gray}{\pm 0.108}}} \\
LightG & \ensuremath{1.360_{\textcolor{gray}{\pm 0.032}}} & \ensuremath{\textbf{0.092}_{\textcolor{gray}{\pm 0.003}}} & \ensuremath{1.068_{\textcolor{gray}{\pm 0.061}}} & \ensuremath{0.846_{\textcolor{gray}{\pm 0.006}}} & \ensuremath{0.691_{\textcolor{gray}{\pm 0.001}}} & \ensuremath{0.990_{\textcolor{gray}{\pm 0.003}}} & \ensuremath{0.857_{\textcolor{gray}{\pm 0.007}}} & \ensuremath{0.711_{\textcolor{gray}{\pm 0.002}}} & \ensuremath{0.727_{\textcolor{gray}{\pm 0.019}}} & \ensuremath{0.744_{\textcolor{gray}{\pm 0.037}}} & \ensuremath{\textbf{0.677}_{\textcolor{gray}{\pm 0.002}}} & \ensuremath{0.854_{\textcolor{gray}{\pm 0.110}}} \\
XGB & \ensuremath{1.419_{\textcolor{gray}{\pm 0.027}}} & \ensuremath{0.595_{\textcolor{gray}{\pm 0.001}}} & \ensuremath{1.310_{\textcolor{gray}{\pm 0.166}}} & \ensuremath{0.841_{\textcolor{gray}{\pm 0.013}}} & \ensuremath{0.692_{\textcolor{gray}{\pm 0.001}}} & \ensuremath{0.946_{\textcolor{gray}{\pm 0.001}}} & \ensuremath{0.864_{\textcolor{gray}{\pm 0.008}}} & \ensuremath{0.716_{\textcolor{gray}{\pm 0.002}}} & \ensuremath{0.727_{\textcolor{gray}{\pm 0.035}}} & \ensuremath{\underline{0.784}_{\textcolor{gray}{\pm 0.001}}} & \ensuremath{0.671_{\textcolor{gray}{\pm 0.002}}} & \ensuremath{0.865_{\textcolor{gray}{\pm 0.018}}} \\
MLP & \ensuremath{1.358_{\textcolor{gray}{\pm 0.066}}} & \ensuremath{0.236_{\textcolor{gray}{\pm 0.006}}} & \ensuremath{1.147_{\textcolor{gray}{\pm 0.078}}} & \ensuremath{0.837_{\textcolor{gray}{\pm 0.004}}} & \ensuremath{0.676_{\textcolor{gray}{\pm 0.009}}} & \ensuremath{0.933_{\textcolor{gray}{\pm 0.044}}} & \ensuremath{0.870_{\textcolor{gray}{\pm 0.017}}} & \ensuremath{0.705_{\textcolor{gray}{\pm 0.001}}} & \ensuremath{0.643_{\textcolor{gray}{\pm 0.034}}} & \ensuremath{0.774_{\textcolor{gray}{\pm 0.003}}} & \ensuremath{0.633_{\textcolor{gray}{\pm 0.015}}} & \ensuremath{0.604_{\textcolor{gray}{\pm 0.036}}} \\
\midrule
\multicolumn{13}{c}{\textbf{\textit{Deep Tabular Models}}} \\
\midrule
ResNet & \ensuremath{1.486_{\textcolor{gray}{\pm 0.101}}} & \ensuremath{0.368_{\textcolor{gray}{\pm 0.035}}} & \ensuremath{1.045_{\textcolor{gray}{\pm 0.087}}} & \ensuremath{0.839_{\textcolor{gray}{\pm 0.005}}} & \ensuremath{0.691_{\textcolor{gray}{\pm 0.001}}} & \ensuremath{0.933_{\textcolor{gray}{\pm 0.028}}} & \ensuremath{0.878_{\textcolor{gray}{\pm 0.014}}} & \ensuremath{0.707_{\textcolor{gray}{\pm 0.004}}} & \ensuremath{0.723_{\textcolor{gray}{\pm 0.031}}} & \ensuremath{0.776_{\textcolor{gray}{\pm 0.002}}} & \ensuremath{0.647_{\textcolor{gray}{\pm 0.002}}} & \ensuremath{0.750_{\textcolor{gray}{\pm 0.062}}} \\
TabM & \ensuremath{1.409_{\textcolor{gray}{\pm 0.174}}} & \ensuremath{0.423_{\textcolor{gray}{\pm 0.115}}} & \ensuremath{1.054_{\textcolor{gray}{\pm 0.068}}} & \ensuremath{0.804_{\textcolor{gray}{\pm 0.043}}} & \ensuremath{0.690_{\textcolor{gray}{\pm 0.002}}} & \ensuremath{0.976_{\textcolor{gray}{\pm 0.010}}} & \ensuremath{0.868_{\textcolor{gray}{\pm 0.024}}} & \ensuremath{0.702_{\textcolor{gray}{\pm 0.019}}} & \ensuremath{0.739_{\textcolor{gray}{\pm 0.014}}} & \ensuremath{\underline{0.784}_{\textcolor{gray}{\pm 0.008}}} & \ensuremath{0.660_{\textcolor{gray}{\pm 0.005}}} & \ensuremath{0.771_{\textcolor{gray}{\pm 0.219}}} \\
ExcelF & \ensuremath{1.373_{\textcolor{gray}{\pm 0.005}}} & \ensuremath{0.368_{\textcolor{gray}{\pm 0.036}}} & \ensuremath{1.114_{\textcolor{gray}{\pm 0.207}}} & \ensuremath{0.843_{\textcolor{gray}{\pm 0.003}}} & \ensuremath{0.692_{\textcolor{gray}{\pm 0.001}}} & \ensuremath{0.931_{\textcolor{gray}{\pm 0.057}}} & \ensuremath{0.873_{\textcolor{gray}{\pm 0.014}}} & \ensuremath{0.716_{\textcolor{gray}{\pm 0.002}}} & \ensuremath{0.748_{\textcolor{gray}{\pm 0.030}}} & \ensuremath{0.775_{\textcolor{gray}{\pm 0.009}}} & \ensuremath{0.664_{\textcolor{gray}{\pm 0.006}}} & \ensuremath{0.833_{\textcolor{gray}{\pm 0.157}}} \\
TabNet & \ensuremath{1.475_{\textcolor{gray}{\pm 0.164}}} & \ensuremath{0.347_{\textcolor{gray}{\pm 0.016}}} & \ensuremath{1.705_{\textcolor{gray}{\pm 0.365}}} & \ensuremath{0.826_{\textcolor{gray}{\pm 0.005}}} & \ensuremath{0.686_{\textcolor{gray}{\pm 0.005}}} & \ensuremath{0.917_{\textcolor{gray}{\pm 0.016}}} & \ensuremath{0.815_{\textcolor{gray}{\pm 0.031}}} & \ensuremath{0.701_{\textcolor{gray}{\pm 0.016}}} & \ensuremath{0.706_{\textcolor{gray}{\pm 0.075}}} & \ensuremath{0.749_{\textcolor{gray}{\pm 0.012}}} & \ensuremath{0.633_{\textcolor{gray}{\pm 0.004}}} & \ensuremath{0.604_{\textcolor{gray}{\pm 0.157}}} \\
TabT & \ensuremath{1.373_{\textcolor{gray}{\pm 0.016}}} & \ensuremath{0.685_{\textcolor{gray}{\pm 0.003}}} & \ensuremath{1.253_{\textcolor{gray}{\pm 0.009}}} & \ensuremath{0.827_{\textcolor{gray}{\pm 0.021}}} & \ensuremath{0.656_{\textcolor{gray}{\pm 0.005}}} & \ensuremath{0.972_{\textcolor{gray}{\pm 0.013}}} & \ensuremath{0.859_{\textcolor{gray}{\pm 0.032}}} & \ensuremath{0.700_{\textcolor{gray}{\pm 0.016}}} & \ensuremath{0.667_{\textcolor{gray}{\pm 0.016}}} & \ensuremath{0.743_{\textcolor{gray}{\pm 0.012}}} & \ensuremath{0.597_{\textcolor{gray}{\pm 0.005}}} & \ensuremath{0.750_{\textcolor{gray}{\pm 0.062}}} \\
Trompt & \ensuremath{1.353_{\textcolor{gray}{\pm 0.009}}} & \ensuremath{0.694_{\textcolor{gray}{\pm 0.007}}} & \ensuremath{1.256_{\textcolor{gray}{\pm 0.011}}} & \ensuremath{0.853_{\textcolor{gray}{\pm 0.001}}} & \ensuremath{0.692_{\textcolor{gray}{\pm 0.001}}} & \ensuremath{0.987_{\textcolor{gray}{\pm 0.006}}} & \ensuremath{0.880_{\textcolor{gray}{\pm 0.014}}} & \ensuremath{0.718_{\textcolor{gray}{\pm 0.002}}} & \ensuremath{0.742_{\textcolor{gray}{\pm 0.016}}} & \ensuremath{0.781_{\textcolor{gray}{\pm 0.001}}} & \ensuremath{0.638_{\textcolor{gray}{\pm 0.016}}} & \ensuremath{\underline{0.896}_{\textcolor{gray}{\pm 0.095}}} \\
MNCA & \ensuremath{1.387_{\textcolor{gray}{\pm 0.009}}} & \ensuremath{0.314_{\textcolor{gray}{\pm 0.005}}} & \ensuremath{\underline{0.938}_{\textcolor{gray}{\pm 0.179}}} & \ensuremath{0.847_{\textcolor{gray}{\pm 0.007}}} & \ensuremath{0.682_{\textcolor{gray}{\pm 0.002}}} & \ensuremath{0.976_{\textcolor{gray}{\pm 0.002}}} & \ensuremath{0.680_{\textcolor{gray}{\pm 0.173}}} & \ensuremath{0.714_{\textcolor{gray}{\pm 0.001}}} & \ensuremath{0.728_{\textcolor{gray}{\pm 0.018}}} & \ensuremath{0.747_{\textcolor{gray}{\pm 0.002}}} & \ensuremath{0.668_{\textcolor{gray}{\pm 0.007}}} & \ensuremath{\underline{0.896}_{\textcolor{gray}{\pm 0.036}}} \\
\midrule
\multicolumn{13}{c}{\textbf{\textit{Tabular In-Context Learning Models}}} \\
\midrule
TabPFN & \ensuremath{1.345_{\textcolor{gray}{\pm 0.001}}} & \ensuremath{0.570_{\textcolor{gray}{\pm 0.001}}} & \ensuremath{1.256_{\textcolor{gray}{\pm 0.001}}} & \ensuremath{0.848_{\textcolor{gray}{\pm 0.001}}} & \ensuremath{\underline{0.697}_{\textcolor{gray}{\pm 0.001}}} & \ensuremath{0.981_{\textcolor{gray}{\pm 0.001}}} & \ensuremath{0.876_{\textcolor{gray}{\pm 0.001}}} & \ensuremath{0.715_{\textcolor{gray}{\pm 0.001}}} & \ensuremath{\underline{0.758}_{\textcolor{gray}{\pm 0.001}}} & \ensuremath{0.743_{\textcolor{gray}{\pm 0.001}}} & \ensuremath{0.635_{\textcolor{gray}{\pm 0.001}}} & \ensuremath{0.818_{\textcolor{gray}{\pm 0.001}}} \\
kNNPFN & \ensuremath{\underline{1.334}_{\textcolor{gray}{\pm 0.001}}} & \ensuremath{0.518_{\textcolor{gray}{\pm 0.001}}} & \ensuremath{1.246_{\textcolor{gray}{\pm 0.001}}} & \ensuremath{0.850_{\textcolor{gray}{\pm 0.001}}} & \ensuremath{0.653_{\textcolor{gray}{\pm 0.001}}} & \ensuremath{0.989_{\textcolor{gray}{\pm 0.001}}} & \ensuremath{0.884_{\textcolor{gray}{\pm 0.001}}} & \ensuremath{0.714_{\textcolor{gray}{\pm 0.001}}} & \ensuremath{\textbf{0.761}_{\textcolor{gray}{\pm 0.001}}} & \ensuremath{0.714_{\textcolor{gray}{\pm 0.001}}} & \ensuremath{0.621_{\textcolor{gray}{\pm 0.001}}} & \ensuremath{0.793_{\textcolor{gray}{\pm 0.001}}} \\
TabDPT & \ensuremath{1.347_{\textcolor{gray}{\pm 0.001}}} & \ensuremath{0.120_{\textcolor{gray}{\pm 0.001}}} & \ensuremath{1.043_{\textcolor{gray}{\pm 0.001}}} & \ensuremath{0.851_{\textcolor{gray}{\pm 0.001}}} & \ensuremath{0.613_{\textcolor{gray}{\pm 0.001}}} & \ensuremath{0.981_{\textcolor{gray}{\pm 0.001}}} & \ensuremath{0.886_{\textcolor{gray}{\pm 0.001}}} & \ensuremath{0.708_{\textcolor{gray}{\pm 0.001}}} & \ensuremath{0.749_{\textcolor{gray}{\pm 0.001}}} & \ensuremath{0.723_{\textcolor{gray}{\pm 0.001}}} & \ensuremath{0.641_{\textcolor{gray}{\pm 0.001}}} & \ensuremath{\textbf{0.998}_{\textcolor{gray}{\pm 0.001}}} \\
\midrule
\multicolumn{13}{c}{\textbf{\textit{Ours (Tabular In-Context Learning Models + AWARE)}}} \\
\midrule
KNN & \ensuremath{1.481_{\textcolor{gray}{\pm 0.023}}} & \ensuremath{0.595_{\textcolor{gray}{\pm 0.003}}} & \ensuremath{1.070_{\textcolor{gray}{\pm 0.062}}} & \ensuremath{0.834_{\textcolor{gray}{\pm 0.003}}} & \ensuremath{0.691_{\textcolor{gray}{\pm 0.001}}} & \ensuremath{0.964_{\textcolor{gray}{\pm 0.008}}} & \ensuremath{0.848_{\textcolor{gray}{\pm 0.030}}} & \ensuremath{0.711_{\textcolor{gray}{\pm 0.001}}} & \ensuremath{0.712_{\textcolor{gray}{\pm 0.044}}} & \ensuremath{0.781_{\textcolor{gray}{\pm 0.001}}} & \ensuremath{0.649_{\textcolor{gray}{\pm 0.007}}} & \ensuremath{0.510_{\textcolor{gray}{\pm 0.100}}} \\
kNNPFN  & \ensuremath{1.444_{\textcolor{gray}{\pm 0.041}}} & \ensuremath{0.275_{\textcolor{gray}{\pm 0.003}}} & \ensuremath{0.966_{\textcolor{gray}{\pm 0.136}}} & \ensuremath{{\underline{0.874}}^{**}_{\textcolor{gray}{\pm 0.003}}} & \ensuremath{\textbf{0.711}_{\textcolor{gray}{\pm 0.001}}} & \ensuremath{\underline{0.998}_{\textcolor{gray}{\pm 0.002}}} & \ensuremath{\underline{0.887}_{\textcolor{gray}{\pm 0.014}}} & \ensuremath{\underline{0.731}_{\textcolor{gray}{\pm 0.001}}} & \ensuremath{0.727_{\textcolor{gray}{\pm 0.043}}} & \ensuremath{\textbf{0.796}_{\textcolor{gray}{\pm 0.003}}} & \ensuremath{0.653_{\textcolor{gray}{\pm 0.006}}} & \ensuremath{0.813_{\textcolor{gray}{\pm 0.021}}} \\
TabDPT  & \ensuremath{\textbf{1.333}_{\textcolor{gray}{\pm 0.001}}} & \ensuremath{0.119_{\textcolor{gray}{\pm 0.001}}} & \ensuremath{\textbf{0.930}_{\textcolor{gray}{\pm 0.088}}} & \ensuremath{{\textbf{0.875}}^{**}_{\textcolor{gray}{\pm 0.001}}} & \ensuremath{\textbf{0.711}_{\textcolor{gray}{\pm 0.002}}} & \ensuremath{\textbf{0.999}_{\textcolor{gray}{\pm 0.002}}} & \ensuremath{\textbf{0.895}_{\textcolor{gray}{\pm 0.001}}} & \ensuremath{\textbf{0.733}_{\textcolor{gray}{\pm 0.001}}} & \ensuremath{0.739_{\textcolor{gray}{\pm 0.025}}} & \ensuremath{\textbf{0.796}_{\textcolor{gray}{\pm 0.001}}} & \ensuremath{0.671_{\textcolor{gray}{\pm 0.004}}} & \ensuremath{0.848_{\textcolor{gray}{\pm 0.146}}} \\
\midrule
\bottomrule
\end{tabular}
}
\end{table*}
\begin{table*}[t]
\centering
\caption{Performance on hospital-acquired infection prediction tasks on large-scale EHR datasets, including MIMIC-IV and eICU. For each task and each model, we extensively ran hyperparameters tuning for 30 rounds. \highlightfour{Markers ($^{*}$p$<$0.05, $^{**}$p$<$0.01) on AWARE models indicate they significantly outperform all non-AWARE baselines (paired bootstrap, Holm-Bonferroni).}}
\label{tab:hai_results}
\resizebox{\textwidth}{!}{

\renewcommand{\arraystretch}{0.95}
\begin{tabular}{lcccccccccc}
\toprule
\multirow{3}{*}{\textbf{Model}} & \multicolumn{6}{c}{\textbf{MIMIC-IV}} & \multicolumn{4}{c}{\textbf{eICU}} \\
\cmidrule(lr){2-7} \cmidrule(lr){8-11}
 & \multicolumn{2}{c}{SEPSIS} & \multicolumn{2}{c}{VAP} & \multicolumn{2}{c}{UTI} & \multicolumn{2}{c}{SEPSIS} & \multicolumn{2}{c}{UTI} \\
\cmidrule(lr){2-3} \cmidrule(lr){4-5} \cmidrule(lr){6-7} \cmidrule(lr){8-9} \cmidrule(lr){10-11}
 & AUROC $\uparrow$ & AUPRC $\uparrow$ & AUROC $\uparrow$ & AUPRC $\uparrow$ & AUROC $\uparrow$ & AUPRC $\uparrow$ & AUROC $\uparrow$ & AUPRC $\uparrow$ & AUROC $\uparrow$ & AUPRC $\uparrow$ \\
\midrule
\multicolumn{11}{c}{\textbf{\textit{Classical and Boosting Models}}} \\
\midrule
LR & \ensuremath{0.886_{\textcolor{gray}{\pm 0.012}}} & \ensuremath{0.490_{\textcolor{gray}{\pm 0.019}}} & \ensuremath{0.752_{\textcolor{gray}{\pm 0.023}}} & \ensuremath{0.218_{\textcolor{gray}{\pm 0.020}}} & \ensuremath{0.649_{\textcolor{gray}{\pm 0.015}}} & \ensuremath{0.247_{\textcolor{gray}{\pm 0.015}}} & \ensuremath{0.940_{\textcolor{gray}{\pm 0.016}}} & \ensuremath{0.577_{\textcolor{gray}{\pm 0.051}}} & \ensuremath{0.922_{\textcolor{gray}{\pm 0.013}}} & \ensuremath{0.126_{\textcolor{gray}{\pm 0.015}}} \\
KNN & \ensuremath{0.794_{\textcolor{gray}{\pm 0.013}}} & \ensuremath{0.391_{\textcolor{gray}{\pm 0.026}}} & \ensuremath{0.709_{\textcolor{gray}{\pm 0.015}}} & \ensuremath{0.213_{\textcolor{gray}{\pm 0.008}}} & \ensuremath{0.569_{\textcolor{gray}{\pm 0.025}}} & \ensuremath{0.209_{\textcolor{gray}{\pm 0.025}}} & \ensuremath{0.795_{\textcolor{gray}{\pm 0.023}}} & \ensuremath{0.314_{\textcolor{gray}{\pm 0.027}}} & \ensuremath{0.734_{\textcolor{gray}{\pm 0.041}}} & \ensuremath{0.056_{\textcolor{gray}{\pm 0.011}}} \\
CatB & \ensuremath{0.903_{\textcolor{gray}{\pm 0.008}}} & \ensuremath{0.527_{\textcolor{gray}{\pm 0.025}}} & \ensuremath{0.754_{\textcolor{gray}{\pm 0.010}}} & \ensuremath{0.235_{\textcolor{gray}{\pm 0.014}}} & \ensuremath{0.659_{\textcolor{gray}{\pm 0.013}}} & \ensuremath{0.257_{\textcolor{gray}{\pm 0.013}}} & \ensuremath{0.981_{\textcolor{gray}{\pm 0.001}}} & \ensuremath{0.756_{\textcolor{gray}{\pm 0.008}}} & \ensuremath{0.937_{\textcolor{gray}{\pm 0.010}}} & \ensuremath{0.153_{\textcolor{gray}{\pm 0.004}}} \\
LightG & \ensuremath{0.904_{\textcolor{gray}{\pm 0.006}}} & \ensuremath{0.521_{\textcolor{gray}{\pm 0.020}}} & \ensuremath{0.752_{\textcolor{gray}{\pm 0.012}}} & \ensuremath{0.212_{\textcolor{gray}{\pm 0.006}}} & \ensuremath{0.643_{\textcolor{gray}{\pm 0.007}}} & \ensuremath{0.236_{\textcolor{gray}{\pm 0.008}}} & \ensuremath{0.982_{\textcolor{gray}{\pm 0.002}}} & \ensuremath{\underline{0.757}_{\textcolor{gray}{\pm 0.016}}} & \ensuremath{0.920_{\textcolor{gray}{\pm 0.041}}} & \ensuremath{0.128_{\textcolor{gray}{\pm 0.034}}} \\
XGB & \ensuremath{0.914_{\textcolor{gray}{\pm 0.007}}} & \ensuremath{\underline{0.563}_{\textcolor{gray}{\pm 0.032}}} & \ensuremath{0.750_{\textcolor{gray}{\pm 0.018}}} & \ensuremath{0.210_{\textcolor{gray}{\pm 0.018}}} & \ensuremath{0.639_{\textcolor{gray}{\pm 0.022}}} & \ensuremath{0.237_{\textcolor{gray}{\pm 0.018}}} & \ensuremath{\underline{0.983}_{\textcolor{gray}{\pm 0.001}}} & \ensuremath{\textbf{0.764}_{\textcolor{gray}{\pm 0.003}}} & \ensuremath{0.919_{\textcolor{gray}{\pm 0.001}}} & \ensuremath{\underline{0.174}_{\textcolor{gray}{\pm 0.051}}} \\
MLP & \ensuremath{0.899_{\textcolor{gray}{\pm 0.007}}} & \ensuremath{0.516_{\textcolor{gray}{\pm 0.014}}} & \ensuremath{0.766_{\textcolor{gray}{\pm 0.010}}} & \ensuremath{0.233_{\textcolor{gray}{\pm 0.017}}} & \ensuremath{0.629_{\textcolor{gray}{\pm 0.025}}} & \ensuremath{0.241_{\textcolor{gray}{\pm 0.019}}} & \ensuremath{0.967_{\textcolor{gray}{\pm 0.003}}} & \ensuremath{0.670_{\textcolor{gray}{\pm 0.019}}} & \ensuremath{0.905_{\textcolor{gray}{\pm 0.027}}} & \ensuremath{0.114_{\textcolor{gray}{\pm 0.030}}} \\
\midrule
\multicolumn{11}{c}{\textbf{\textit{Deep Tabular Models}}} \\
\midrule
ResNet & \ensuremath{0.882_{\textcolor{gray}{\pm 0.006}}} & \ensuremath{0.488_{\textcolor{gray}{\pm 0.010}}} & \ensuremath{0.733_{\textcolor{gray}{\pm 0.032}}} & \ensuremath{0.205_{\textcolor{gray}{\pm 0.018}}} & \ensuremath{0.600_{\textcolor{gray}{\pm 0.009}}} & \ensuremath{0.219_{\textcolor{gray}{\pm 0.008}}} & \ensuremath{0.972_{\textcolor{gray}{\pm 0.003}}} & \ensuremath{0.707_{\textcolor{gray}{\pm 0.009}}} & \ensuremath{0.896_{\textcolor{gray}{\pm 0.003}}} & \ensuremath{0.094_{\textcolor{gray}{\pm 0.009}}} \\
TabM & \ensuremath{0.870_{\textcolor{gray}{\pm 0.001}}} & \ensuremath{0.463_{\textcolor{gray}{\pm 0.004}}} & \ensuremath{0.767_{\textcolor{gray}{\pm 0.011}}} & \ensuremath{0.258_{\textcolor{gray}{\pm 0.011}}} & \ensuremath{0.645_{\textcolor{gray}{\pm 0.010}}} & \ensuremath{0.251_{\textcolor{gray}{\pm 0.010}}} & \ensuremath{0.981_{\textcolor{gray}{\pm 0.001}}} & \ensuremath{0.753_{\textcolor{gray}{\pm 0.006}}} & \ensuremath{\underline{0.941}_{\textcolor{gray}{\pm 0.006}}} & \ensuremath{0.130_{\textcolor{gray}{\pm 0.006}}} \\
ExcelF & \ensuremath{\underline{0.917}_{\textcolor{gray}{\pm 0.001}}} & \ensuremath{\textbf{0.570}_{\textcolor{gray}{\pm 0.007}}} & \ensuremath{0.796_{\textcolor{gray}{\pm 0.003}}} & \ensuremath{0.258_{\textcolor{gray}{\pm 0.005}}} & \ensuremath{0.657_{\textcolor{gray}{\pm 0.007}}} & \ensuremath{0.247_{\textcolor{gray}{\pm 0.006}}} & \ensuremath{0.976_{\textcolor{gray}{\pm 0.001}}} & \ensuremath{0.673_{\textcolor{gray}{\pm 0.029}}} & \ensuremath{\textbf{0.945}_{\textcolor{gray}{\pm 0.007}}} & \ensuremath{\textbf{0.177}_{\textcolor{gray}{\pm 0.004}}} \\
TabNet & \ensuremath{0.893_{\textcolor{gray}{\pm 0.009}}} & \ensuremath{0.506_{\textcolor{gray}{\pm 0.019}}} & \ensuremath{0.612_{\textcolor{gray}{\pm 0.102}}} & \ensuremath{0.133_{\textcolor{gray}{\pm 0.040}}} & \ensuremath{0.543_{\textcolor{gray}{\pm 0.028}}} & \ensuremath{0.180_{\textcolor{gray}{\pm 0.014}}} & \ensuremath{0.963_{\textcolor{gray}{\pm 0.008}}} & \ensuremath{0.649_{\textcolor{gray}{\pm 0.041}}} & \ensuremath{0.589_{\textcolor{gray}{\pm 0.080}}} & \ensuremath{0.018_{\textcolor{gray}{\pm 0.004}}} \\
TabT & \ensuremath{0.886_{\textcolor{gray}{\pm 0.004}}} & \ensuremath{0.487_{\textcolor{gray}{\pm 0.002}}} & \ensuremath{0.727_{\textcolor{gray}{\pm 0.012}}} & \ensuremath{0.198_{\textcolor{gray}{\pm 0.009}}} & \ensuremath{0.610_{\textcolor{gray}{\pm 0.016}}} & \ensuremath{0.221_{\textcolor{gray}{\pm 0.014}}} & \ensuremath{0.955_{\textcolor{gray}{\pm 0.004}}} & \ensuremath{0.609_{\textcolor{gray}{\pm 0.011}}} & \ensuremath{0.892_{\textcolor{gray}{\pm 0.002}}} & \ensuremath{0.080_{\textcolor{gray}{\pm 0.005}}} \\
Trompt & \ensuremath{0.903_{\textcolor{gray}{\pm 0.005}}} & \ensuremath{0.510_{\textcolor{gray}{\pm 0.013}}} & \ensuremath{0.784_{\textcolor{gray}{\pm 0.012}}} & \ensuremath{0.261_{\textcolor{gray}{\pm 0.013}}} & \ensuremath{0.636_{\textcolor{gray}{\pm 0.012}}} & \ensuremath{0.239_{\textcolor{gray}{\pm 0.008}}} & \ensuremath{0.981_{\textcolor{gray}{\pm 0.002}}} & \ensuremath{0.747_{\textcolor{gray}{\pm 0.014}}} & \ensuremath{0.940_{\textcolor{gray}{\pm 0.010}}} & \ensuremath{0.139_{\textcolor{gray}{\pm 0.045}}} \\
MNCA & \ensuremath{0.888_{\textcolor{gray}{\pm 0.008}}} & \ensuremath{0.505_{\textcolor{gray}{\pm 0.019}}} & \ensuremath{0.716_{\textcolor{gray}{\pm 0.018}}} & \ensuremath{0.202_{\textcolor{gray}{\pm 0.017}}} & \ensuremath{0.622_{\textcolor{gray}{\pm 0.012}}} & \ensuremath{0.227_{\textcolor{gray}{\pm 0.007}}} & \ensuremath{0.975_{\textcolor{gray}{\pm 0.002}}} & \ensuremath{0.727_{\textcolor{gray}{\pm 0.011}}} & \ensuremath{0.931_{\textcolor{gray}{\pm 0.002}}} & \ensuremath{0.129_{\textcolor{gray}{\pm 0.008}}} \\
\midrule
\multicolumn{11}{c}{\textbf{\textit{Tabular In-Context Learning Models}}} \\
\midrule
TabPFN & \ensuremath{0.905_{\textcolor{gray}{\pm 0.002}}} & \ensuremath{0.445_{\textcolor{gray}{\pm 0.009}}} & \ensuremath{0.783_{\textcolor{gray}{\pm 0.006}}} & \ensuremath{0.212_{\textcolor{gray}{\pm 0.006}}} & \ensuremath{0.639_{\textcolor{gray}{\pm 0.006}}} & \ensuremath{0.226_{\textcolor{gray}{\pm 0.003}}} & \ensuremath{0.821_{\textcolor{gray}{\pm 0.133}}} & \ensuremath{0.291_{\textcolor{gray}{\pm 0.166}}} & \ensuremath{0.641_{\textcolor{gray}{\pm 0.068}}} & \ensuremath{0.025_{\textcolor{gray}{\pm 0.007}}} \\
kNNPFN & \ensuremath{0.906_{\textcolor{gray}{\pm 0.001}}} & \ensuremath{0.514_{\textcolor{gray}{\pm 0.001}}} & \ensuremath{0.789_{\textcolor{gray}{\pm 0.001}}} & \ensuremath{0.279_{\textcolor{gray}{\pm 0.001}}} & \ensuremath{{0.664}_{\textcolor{gray}{\pm 0.001}}} & \ensuremath{0.256_{\textcolor{gray}{\pm 0.001}}} & \ensuremath{0.956_{\textcolor{gray}{\pm 0.001}}} & \ensuremath{0.621_{\textcolor{gray}{\pm 0.001}}} & \ensuremath{0.876_{\textcolor{gray}{\pm 0.001}}} & \ensuremath{0.092_{\textcolor{gray}{\pm 0.001}}} \\
TabDPT & \ensuremath{0.905_{\textcolor{gray}{\pm 0.001}}} & \ensuremath{0.528_{\textcolor{gray}{\pm 0.001}}} & \ensuremath{0.784_{\textcolor{gray}{\pm 0.001}}} & \ensuremath{0.274_{\textcolor{gray}{\pm 0.001}}} & \ensuremath{0.658_{\textcolor{gray}{\pm 0.001}}} & \ensuremath{0.257_{\textcolor{gray}{\pm 0.001}}} & \ensuremath{0.961_{\textcolor{gray}{\pm 0.001}}} & \ensuremath{0.659_{\textcolor{gray}{\pm 0.001}}} & \ensuremath{0.882_{\textcolor{gray}{\pm 0.001}}} & \ensuremath{0.103_{\textcolor{gray}{\pm 0.001}}} \\
\midrule
\multicolumn{11}{c}{\textbf{\textit{Ours (Tabular In-Context Learning Models + AWARE)}}} \\
\midrule
KNN  & \ensuremath{0.903_{\textcolor{gray}{\pm 0.001}}} & \ensuremath{0.507_{\textcolor{gray}{\pm 0.006}}} & \ensuremath{0.788_{\textcolor{gray}{\pm 0.003}}} & \ensuremath{0.250_{\textcolor{gray}{\pm 0.005}}} & \ensuremath{{0.671}_{\textcolor{gray}{\pm 0.002}}} & \ensuremath{0.260_{\textcolor{gray}{\pm 0.004}}} & \ensuremath{0.971_{\textcolor{gray}{\pm 0.001}}} & \ensuremath{0.711_{\textcolor{gray}{\pm 0.002}}} & \ensuremath{0.928_{\textcolor{gray}{\pm 0.002}}} & \ensuremath{0.155_{\textcolor{gray}{\pm 0.003}}} \\
kNNPFN  & \ensuremath{\underline{0.917}_{\textcolor{gray}{\pm 0.001}}} & \ensuremath{0.538_{\textcolor{gray}{\pm 0.006}}} & \ensuremath{\underline{0.799}_{\textcolor{gray}{\pm 0.001}}} & \ensuremath{{\underline{0.282}}^{**}_{\textcolor{gray}{\pm 0.001}}} & \ensuremath{\underline{0.673}^{**}_{\textcolor{gray}{\pm 0.002}}} & \ensuremath{\underline{0.263}^{**}_{\textcolor{gray}{\pm 0.006}}} & \ensuremath{0.981_{\textcolor{gray}{\pm 0.001}}} & \ensuremath{0.724_{\textcolor{gray}{\pm 0.004}}} & \ensuremath{0.932_{\textcolor{gray}{\pm 0.002}}} & \ensuremath{0.152_{\textcolor{gray}{\pm 0.005}}} \\
TabDPT  & \ensuremath{{\textbf{0.925}}^{**}_{\textcolor{gray}{\pm 0.001}}} & \ensuremath{0.555_{\textcolor{gray}{\pm 0.001}}} & \ensuremath{{\textbf{0.802}}^{**}_{\textcolor{gray}{\pm 0.001}}} & \ensuremath{{\textbf{0.288}}^{**}_{\textcolor{gray}{\pm 0.004}}} & \ensuremath{\textbf{0.674}^{**}_{\textcolor{gray}{\pm 0.001}}} & \ensuremath{\textbf{0.266}^{**}_{\textcolor{gray}{\pm 0.003}}} & \ensuremath{\textbf{0.990}^{**}_{\textcolor{gray}{\pm 0.001}}} & \ensuremath{0.738_{\textcolor{gray}{\pm 0.005}}} & \ensuremath{0.935_{\textcolor{gray}{\pm 0.002}}} & \ensuremath{0.160_{\textcolor{gray}{\pm 0.004}}} \\
\midrule
\bottomrule
\end{tabular}
}
\end{table*}
\clearpage

\subsection{Supplementary Note 2 | Efficiency Trade-offs}
\label{sec:apdx:efficiency}

\begin{figure*}[ht]
    \setlength{\lineskip}{0pt}
    \centering
    \begin{tikzpicture}
        \node[anchor=north west,inner sep=0pt] at (0,0){\includegraphics[width=0.48\linewidth]{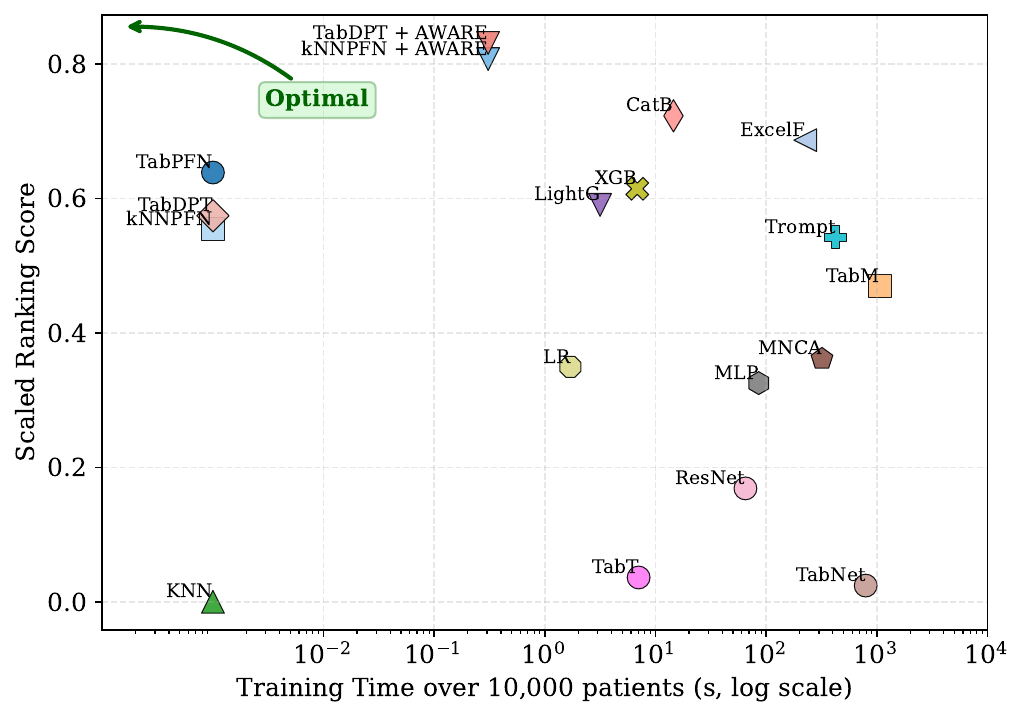}};
        \node[font=\sffamily\bfseries\large] at (0ex, -0.8ex) {a};
    \end{tikzpicture}
    \begin{tikzpicture}
        \node[anchor=north west,inner sep=0pt] at (0,0){\includegraphics[width=0.48\linewidth]{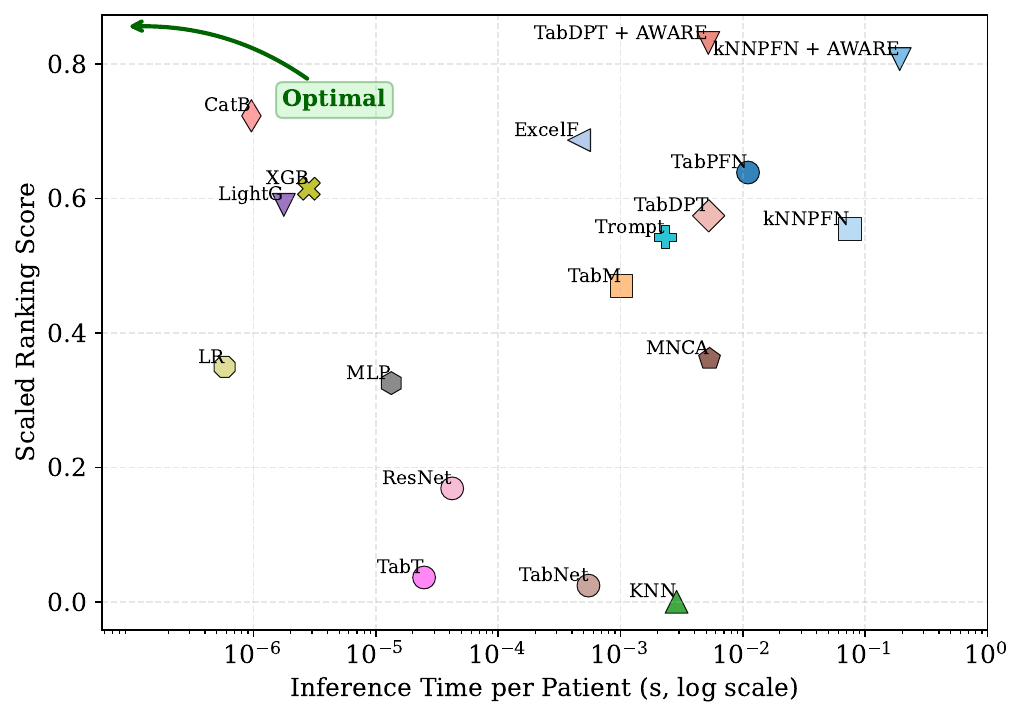}};
        \node[font=\sffamily\bfseries\large] at (0ex, 0ex) {b};
    \end{tikzpicture}
    \caption{Training and inference efficiency trade-offs across model classes on large-scale EHR datasets. The left panel reports relative predictive ranking versus training time (measured until convergence with early stopping), while the right panel reports ranking versus per-patient inference time. Hyperparameter tuning and pretraining costs are excluded. \highlightrevdel{Models based on in-context learning incur no task-specific training cost.} \highlightrev{Vanilla (non-AWARE) in-context learning models incur no task-specific training cost; AWARE-enhanced variants additionally train a lightweight retrieval encoder (and, where applicable, an adapter) per dataset/task, at substantially lower cost than end-to-end training.}}
    \label{fig:apdx:efficiency}
\end{figure*}

Supplementary Figure~\ref{fig:apdx:efficiency} summarizes the trade-offs between computational efficiency and predictive performance across model classes on large-scale EHR datasets. We report relative ranking scores (higher is better) against measured training and inference time. Hyperparameter tuning time is excluded, and for models requiring supervised training, training time reflects convergence with early stopping. \highlightrevdel{For PFN-based and other ICL models, which do not require task-specific training, training time is effectively negligible.} \highlightrev{For vanilla (non-AWARE) PFN-based and other ICL models, which do not require task-specific training, training time is effectively negligible; AWARE-enhanced variants incur additional but comparatively small training cost, quantified below.}

\paragraph{Training Efficiency.}
In terms of training cost, ICL-based models exhibit a distinct advantage. TabPFN and TabDPT incur no dataset-specific training time, yet achieve mid-to-high performance rankings, placing them favorably among all evaluated models. \highlightrevdel{This contrasts with Classical k-nearest neighbors, which is also training-free but exhibits substantially lower predictive rankings.} Gradient-boosted models (e.g., XGBoost, LightGBM) occupy an intermediate regime, requiring modest training time while achieving competitive performance. In contrast, deep tabular and retrieval-based models such as ModernNCA and TabR incur significantly higher training costs while achieving lower or comparable rankings. The highest-ranked deep tabular models (e.g., ExcelFormer, Trompt) require the great amount of training time and computational resources, reflecting the cost of end-to-end representation learning on large EHR cohorts. Our AWARE adapters are the most notable where they achieve both highest rankings while being training time-efficient.

\paragraph{Inference Efficiency.}
Inference-time costs reveal a different pattern. Classical linear and tree-based models remain the fastest at inference, with per-patient prediction times substantially lower than all other model classes. Among neural models, shallow architectures such as MLPs and ResNet-style tabular models achieve moderate inference efficiency. PFN-based ICL models, including TabPFN and TabDPT, exhibit inference times comparable to other transformer-based tabular models and retrieval-augmented approaches. Retrieval-based variants, particularly kNNPFN, incur the highest inference cost due to explicit neighbor search and context construction, especially with AWARE incorporated. Importantly, despite slower inference relative to classical models, PFN-based approaches remain competitive with deep tabular models that require both training and inference overhead.

\paragraph{Implications for Clinical Use.}
Taken together, these results highlight a key efficiency trade-off for in-context learning in EHR prediction. PFN-based models offer a favorable balance between predictive performance and training efficiency by eliminating the need for task-specific retraining, which can be advantageous in settings with limited computational resources or rapidly changing clinical tasks. However, this benefit is partially offset by increased inference-time cost, particularly for retrieval-augmented variants. These findings suggest that ICL-based approaches are best suited for scenarios where retraining is infeasible or costly, while inference latency remains acceptable, rather than as drop-in replacements for lightweight classical models in latency-critical clinical workflows.

\clearpage

\subsection{Supplementary Note 3 | Performance on Synthetic EHR Data}

We evaluated representative models on fully synthetic EHR datasets generated using Synthea \cite{walonoski_synthea_2018}, an open-source patient population simulator that produces realistic yet fully synthetic longitudinal health records based on rule-driven disease modules and publicly available epidemiological statistics. The evaluated tasks include stroke and lung cancer prediction \cite{chen_simulation_2022}. Across all evaluated methods—logistic regression, XGBoost, TabPFN and TabDPT—predictive performance approached ceiling levels, as seen in Supplementary Table \ref{tab:results:synthea}. This behavior is expected given the rule-based generative process underlying Synthea, in which disease progression and outcomes are governed by explicit state-transition logic and probabilistic modules. As a result, target labels are often directly inferable from a small subset of observable features, limiting the need for complex representation learning or retrieval-based inference.

\begin{table}[hbt]
\centering
\caption{AUROC performance on fully synthetic EHR data generated using Synthea for two binary prediction tasks: lung cancer and stroke. Performance across model classes approaches ceiling levels due to the rule-based generative process underlying Synthea. Results are intended as an upper-bound sanity check rather than evidence of real-world generalization.}
\label{tab:results:synthea}
\renewcommand{\arraystretch}{0.95}
\begin{tabular}{lcc}
\toprule
\textbf{Model} & \textbf{Lung Cancer} $\uparrow$ & \textbf{Stroke} $\uparrow$ \\
\midrule
\multicolumn{3}{c}{\textbf{\textit{Classical and Ensemble Models}}} \\
\midrule
Logistic Regression & \ensuremath{0.993_{\textcolor{gray}{\pm 0.00}}} & \ensuremath{0.992_{\textcolor{gray}{\pm 0.00}}} \\
XGBoost            & \ensuremath{0.999_{\textcolor{gray}{\pm 0.00}}} & \ensuremath{0.999_{\textcolor{gray}{\pm 0.00}}} \\
\midrule
\multicolumn{3}{c}{\textbf{\textit{Tabular In-Context Learning Models}}} \\
\midrule
TabPFN  & \ensuremath{1.0_{\textcolor{gray}{\pm 0.00}}} & \ensuremath{0.999_{\textcolor{gray}{\pm 0.00}}} \\
TabDPT  & \ensuremath{0.992_{\textcolor{gray}{\pm 0.00}}} & \ensuremath{0.978_{\textcolor{gray}{\pm 0.00}}} \\
\midrule
\bottomrule
\end{tabular}
\end{table}

Within this saturated regime, PFN-based models achieve performance comparable to strong classical baselines without task-specific retraining or feature engineering. While this does not indicate superiority over established methods, it demonstrates that PFNs are compatible with simulator-driven clinical data and do not exhibit inductive bias mismatches that hinder learning of disease logic. Importantly, this compatibility suggests that the limitations of PFN-based and RA-TICL methods observed on real-world EHR datasets are unlikely to stem from representational capacity alone, but instead arise from challenges inherent to clinical data as described in the main text.

\subsection{Supplementary Note 4 | Dataset Preprocessing and Feature Extraction}
\label{sec:apdx:datasets}

We conduct experiments using two publicly available ICU EHR datasets: MIMIC-IV (v3.1) and eICU-CRD (v2.0). For both datasets, analyses are restricted to adult patients and constructed at the ICU-stay level.

In MIMIC-IV, patients aged $\geq 18$ years are included if ICU stays satisfy duration requirements for hospital-acquired infection prediction. To ensure infections are acquired during hospitalization rather than present at admission, only ICU stays of at least 48 hours are considered. A 24-hour observation window is used to extract patient features, followed by a 24-hour prediction gap, after which the model predicts the onset of hospital-acquired infections. Outcomes include three clinically relevant infections commonly monitored in critical care settings: sepsis, urinary tract infection (UTI), and ventilator-associated pneumonia (VAP). These outcomes are identified using diagnosis codes and clinical event records mapped to standardized infection definitions where available.

In eICU-CRD, adult patients aged 18–89 years with ICU stays between 12 hours and 10 days are included. Clinical events recorded during the first 12 hours of ICU admission are used as the observation window. Prediction targets similarly focus on hospital-acquired infections, including sepsis and urinary tract infection, defined using diagnosis codes, treatment indicators, and infection-related clinical variables when available in the dataset.

Structured features are extracted from multiple EHR sources, including diagnoses, treatments or procedures, medications, laboratory measurements, vital signs, and physiological variables. In MIMIC-IV, diagnosis and procedure codes are encoded using frequency counts, medications are aggregated by mean administration rate and dosage, and charted measurements are treated as continuous variables summarized within fixed temporal buckets (e.g., 2-hour intervals). In eICU-CRD, diagnosis and treatment information is parsed from hierarchical strings and standardized across hospitals. Laboratory measurements and vital signs are normalized and summarized using minimum, maximum, mean, and standard deviation over the observation window, with additional physiological variables derived from APACHE-related features.

Variable-length time series are converted into fixed-length representations through temporal aggregation. Missing continuous measurements in MIMIC-IV are handled via forward- and backward-filling followed by median imputation. In eICU-CRD, missing categorical variables are imputed using the mode, while continuous variables are normalized and missing values are set to zero after normalization. The resulting datasets consist of fixed-length feature matrices with one row per ICU stay, enabling consistent input representations across all evaluated models.

\begin{table*}[ht]
\centering
\caption{
Summary of EHR datasets used in this study, spanning diverse clinical cohorts, task types, and feature modalities.
Feature type abbreviations:
DEM = demographics;
VIT = vital signs;
LAB = laboratory results;
CODE = medical codes / diagnoses;
GEN = genetic / molecular features;
PHYS = physiological signals;
ADMIN = administrative variables;
TMP = temporal features.
}
\label{tab:dataset_stats}
\resizebox{\textwidth}{!}{
\renewcommand{\arraystretch}{0.95}
\begin{tabular}{lccccccc}
\toprule
\textbf{Dataset Name} & \textbf{\#Rows} & \textbf{\#Features} & \textbf{Task} & \textbf{Imbalance Ratio} & \textbf{Cohort} & \textbf{Feature Types} \\
\midrule
\rowcolor{lightgray}
\multicolumn{7}{c}{\textbf{Synthetic EHR datasets}} \\
Synthea-Stroke
\cite{chen_simulation_2022} &
16{,}055 & 793 & Binary & -- &
Multi-domain &
DEM, VIT, LAB, CODE \\
Synthea-LungCancer
\cite{chen_simulation_2022} &
4{,}384 & 773 & Binary & -- &
Oncology &
DEM, VIT, LAB, CODE \\

\rowcolor{lightgray}
\multicolumn{7}{c}{\textbf{Small-to-medium real-world EHR datasets}} \\
Heart Failure (Platelets Counts)
\cite{chicco2020machine} &
299 & 12 & Regression & -- &
Cardiovascular &
DEM, LAB \\
OXF – Parkinsons Telemonitoring
\cite{tsanas2009accurate} &
5{,}875 & 22 & Regression & -- &
Neurological &
DEM, PHYS, TMP \\
TunedIT – ICU
 &
200 & 19 & Binary & -- &
ICU &
DEM, VIT, LAB \\
SUPPORT2
\cite{knaus2020support} &
9{,}105 & 46 & Binary & 2.13 &
Multi-domain &
DEM, VIT, LAB, ADMIN \\
Sepsis Survival Minimal
\cite{sepsis_survival_minimal_clinical_records_827} &
110{,}341 & 4 & Binary & 12.57 &
Infectious &
DEM \\
Differentiated Thyroid Cancer
\cite{differentiated_thyroid_cancer_recurrence_915} &
383 & 16 & Binary & 2.55 &
Oncology &
DEM, LAB \\
Glioma Grading (Clinical + Mutation)
\cite{glioma_grading_clinical_and_mutation_features_759} &
839 & 23 & Binary & 1.38 &
Oncology &
DEM, GEN \\
AIDS Classification 50{,}000 &
50{,}000 & 23 & Binary & 2.22 &
Infectious &
DEM, LAB \\
Indian Liver Patient
\cite{ilpd_(indian_liver_patient_dataset)_225} &
583 & 11 & Binary & 2.54 &
Metabolic &
DEM, LAB \\
COVID-19 Hospital Treatment Plan
\cite{wu2020understanding} &
318{,}438 & 18 & Multiclass & 31.69 &
Infectious &
DEM, ADMIN \\
Diabetes130US
\cite{diabetes_130-us_hospitals_for_years_1999-2008_296} &
101{,}766 & 49 & Binary & 4.83 &
Metabolic &
DEM, LAB \\
Kidney
\cite{mcgilchrist1991regression} &
76 & 6 & Binary & 1.12 &
Metabolic &
DEM, LAB \\

\rowcolor{lightgray}
\multicolumn{7}{c}{\textbf{Large-scale longitudinal real-world EHR datasets}} \\

MIMIC-IV (SEPSIS)
\cite{johnson_mimic-iv_2023} &
35{,}885 & 3{,}400 & Binary & 8.408 &
Infectious &
DEM, VIT, LAB, TMP \\

MIMIC-IV (UTI)
\cite{johnson_mimic-iv_2023} &
24{,}145 & 3{,}440 & Binary & 5.226 &
Infectious &
DEM, VIT, LAB, TMP \\

MIMIC-IV (VAP)
\cite{johnson_mimic-iv_2023} &
24{,}495 & 3{,}440 & Binary & 10.38 &
Infectious &
DEM, VIT, LAB, TMP \\


eICU (SEPSIS)
\cite{pollard2018eicu} &
120{,}442 & 655 & Binary & 14.91 &
Infectious &
DEM, VIT, LAB, TMP \\

eICU (UTI)
\cite{pollard2018eicu} &
124{,}828 & 655 & Binary & 70.65 &
Infectious &
DEM, VIT, LAB, TMP \\

HIPE (CPE)
\cite{pham2025explainable} &
45{,}333 & 28{,}566 & Binary & 434.89 &
Infectious &
DEM, CODE, ADMIN, TMP \\
HIPE (RDMT)
\cite{pham_forecasting_2024} &
78{,}142 & 28{,}779 & Binary & 7.48 &
ICU &
DEM, CODE, ADMIN, TMP \\
\bottomrule
\end{tabular}
}
\end{table*}

\subsection{Supplementary Note 5 | Model Descriptions}
\label{sec:apdx:algorithms}

This section provides brief descriptions of the models included in our benchmark, focusing on their architectural characteristics and relevance to EHR prediction tasks.

\paragraph{TabTransformer} \cite{huang_tabtransformer_2020} models categorical features using contextual embeddings derived from Transformer layers, enabling richer interactions between categorical variables. It has been shown to outperform standard multilayer perceptrons (MLPs) on tabular data and to exhibit robustness to noisy or missing categorical inputs.

\paragraph{TabNet} \cite{arik2021tabnet} is a deep learning architecture for tabular data that performs instance-wise feature selection through a sequential attention mechanism. By selecting a sparse subset of features at each decision step, TabNet provides a degree of interpretability while maintaining competitive predictive performance.

\paragraph{ExcelFormer} \cite{chen_excelformer_2024}is a Transformer-based architecture designed for tabular data, incorporating a semi-permeable attention mechanism and gated linear units. It also introduces interpolation-based data augmentation strategies (Hid-Mix and Feat-Mix) to improve robustness under irregular target functions.

\paragraph{Trompt} \cite{chen2023trompt} is a prompt-inspired neural architecture for tabular learning that derives sample-specific feature importance through learned prompts. By dynamically adjusting feature relevance at inference time, Trompt aims to bridge the performance gap between deep tabular models and tree-based methods.

\paragraph{TabM} \cite{gorishniy2024tabm} is a parameter-efficient deep tabular model that ensembles multiple MLP submodels with shared weights. This design improves training efficiency and generalization while mitigating issues such as dead neurons commonly observed in deep networks.

\paragraph{ModernNCA} \cite{ye2024revisiting} revisits classical Neighborhood Components Analysis (NCA) and integrates modern optimization and representation learning techniques to support differentiable nearest-neighbor retrieval for tabular prediction tasks.

\paragraph{TabPFN} \cite{hollmann_tabpfn_2023} is a prior-fitted network pretrained on a large corpus of synthetically generated tabular datasets derived from causal models. It performs classification via in-context learning, enabling inference-time adaptation without task-specific fine-tuning.

\paragraph{kNNPFN} \cite{thomas_retrieval_nodate} (also referred to as TabPFN-kNN) augments the TabPFN prior-fitted architecture with an explicit $k$-nearest-neighbor retrieval step at inference time. For each query instance, a subset of similar training examples is selected using a distance metric in feature space, and only this retrieved context is provided to the PFN model. This strategy reduces the computational burden associated with conditioning on large training sets while introducing a similarity-based inductive bias. TabPFN-kNN thus represents a simple yet effective retrieval-augmented extension of PFN-based in-context learning, enabling scalable inference-time adaptation on larger tabular datasets.

\paragraph{TabDPT} \cite{ma_tabdpt_2025} is a retrieval-augmented tabular in-context learning model that extends the TabPFN paradigm to larger and more heterogeneous datasets. Instead of conditioning on the full training set, TabDPT constructs a query-specific context via retrieval and processes the retrieved examples jointly with the query through a Transformer encoder. By combining pretrained PFN initialization with explicit context construction, TabDPT enables scalable inference-time adaptation while maintaining the feature-order invariance and amortized Bayesian inference properties of PFN-based models.

\highlightrev{
\subsection{Supplementary Note 6 | Hyperparameter Search Spaces}
\label{sec:apdx:hyperparams}

Supplementary Table~\ref{tab:hyperparameter_ranges} reports the full hyperparameter search space used for each classical machine learning and deep tabular baseline. All models were tuned independently for each dataset using 30 validation trials with a matched tuning budget, as described in the Section \ref{sec:experiments}. Log-uniform ranges are sampled on a logarithmic scale, uniform ranges are sampled linearly, and categorical hyperparameters are sampled uniformly at random from the listed set. Parameters marked ``(optional)'' were included as part of the search space but could be disabled by the tuning procedure (e.g., regularization terms set to zero or omitted).

\begin{longtable}{ll}
\caption{Hyperparameter search spaces used for tuning classical machine learning and deep tabular baselines. }
\label{tab:hyperparameter_ranges} \\
\toprule
\textbf{Hyperparameter} & \textbf{Range / Type} \\
\midrule
\endfirsthead
\toprule
\textbf{Hyperparameter} & \textbf{Range / Type} \\
\midrule
\endhead
\bottomrule
\endfoot

\multicolumn{2}{l}{\textbf{\textit{Classical Machine Learning Methods}}} \\
\midrule
\multicolumn{2}{l}{\textbf{Logistic Regression}} \\
\quad C & Log-uniform $[10^{-5}, 5.0]$ \\
\quad penalty & Categorical $\{\ell_2, \text{none}\}$ \\
\quad max\_iter & Integer $[50, 500]$ \\
\midrule
\multicolumn{2}{l}{\textbf{XGBoost}} \\
\quad learning\_rate & Log-uniform $[10^{-5}, 1.0]$ \\
\quad max\_depth & Integer $[3, 10]$ \\
\quad min\_child\_weight & Log-uniform $[10^{-8}, 10^{5}]$ \\
\quad subsample & Uniform $[0.5, 1.0]$ \\
\quad colsample\_bytree & Uniform $[0.5, 1.0]$ \\
\quad colsample\_bylevel & Uniform $[0.5, 1.0]$ \\
\quad alpha (L1 reg.) & Log-uniform $[10^{-8}, 100.0]$ (optional) \\
\quad lambda (L2 reg.) & Log-uniform $[10^{-8}, 100.0]$ (optional) \\
\quad gamma & Log-uniform $[10^{-8}, 100.0]$ (optional) \\
\midrule
\multicolumn{2}{l}{\textbf{LightGBM}} \\
\quad learning\_rate & Log-uniform $[0.001, 1.0]$ \\
\quad num\_leaves & Integer $[10, 100]$ \\
\quad max\_depth & Integer $[3, 10]$ \\
\quad min\_child\_weight & Log-uniform $[10^{-5}, 0.1]$ \\
\quad min\_child\_samples & Integer $[2, 100]$ \\
\quad subsample & Uniform $[0.5, 1.0]$ \\
\quad colsample\_bytree & Uniform $[0.5, 1.0]$ \\
\quad reg\_lambda & Log-uniform $[10^{-5}, 1.0]$ (optional) \\
\midrule
\multicolumn{2}{l}{\textbf{CatBoost}} \\
\quad learning\_rate & Log-uniform $[10^{-5}, 1.0]$ \\
\quad depth & Integer $[3, 10]$ \\
\quad l2\_leaf\_reg & Log-uniform $[1.0, 10.0]$ \\
\quad bagging\_temperature & Uniform $[0.0, 1.0]$ \\
\quad leaf\_estimation\_iterations & Integer $[1, 10]$ \\
\midrule
\multicolumn{2}{l}{\textbf{MLP}} \\
\quad d\_layers & 1--8 hidden layers, width $[64, 512]$ \\
\quad dropout & Uniform $[0.0, 0.5]$ (optional) \\
\quad lr & Log-uniform $[10^{-5}, 0.01]$ \\
\quad weight\_decay & Log-uniform $[10^{-6}, 0.001]$ (optional) \\
\midrule
\multicolumn{2}{l}{\textbf{\textit{Deep Tabular Methods}}} \\
\midrule
\multicolumn{2}{l}{\textbf{TabNet}} \\
\quad lr & Uniform $[0.001, 0.01]$ \\
\quad gamma & Uniform $[1.0, 2.0]$ \\
\quad n\_steps & Integer $[3, 10]$ \\
\quad n\_independent & Integer $[1, 5]$ \\
\quad n\_shared & Integer $[1, 5]$ \\
\quad momentum & Uniform $[0.01, 0.4]$ \\
\midrule
\multicolumn{2}{l}{\textbf{TabTransformer}} \\
\quad dim & Categorical $\{32, 64, 128, 256\}$ \\
\quad depth & Categorical $\{1, 2, 3, 6, 12\}$ \\
\quad heads & Categorical $\{2, 4, 8\}$ \\
\quad attn\_dropout & Uniform $[0.0, 0.5]$ \\
\quad ff\_dropout & Uniform $[0.0, 0.5]$ \\
\quad lr & Log-uniform $[10^{-5}, 0.1]$ \\
\quad weight\_decay & Log-uniform $[10^{-6}, 0.001]$ (optional) \\
\midrule
\multicolumn{2}{l}{\textbf{ResNet}} \\
\quad n\_layers & Integer $[1, 8]$ \\
\quad $d$ (embedding size) & Integer $[64, 512]$ \\
\quad d\_hidden\_factor & Uniform $[1.0, 4.0]$ \\
\quad hidden\_dropout & Uniform $[0.0, 0.5]$ \\
\quad residual\_dropout & Uniform $[0.0, 0.5]$ (optional) \\
\quad lr & Log-uniform $[10^{-5}, 0.01]$ \\
\quad weight\_decay & Log-uniform $[10^{-6}, 0.001]$ (optional) \\
\quad activation / normalization & Fixed at ReLU / BatchNorm \\
\midrule
\multicolumn{2}{l}{\textbf{TabM}} \\
\quad backbone.n\_blocks & Integer $[1, 5]$ \\
\quad backbone.d\_block & Integer $[64, 1024]$ \\
\quad backbone.dropout & Uniform $[0.0, 0.5]$ \\
\quad num\_embeddings.n\_frequencies & Integer $[16, 96]$ \\
\quad num\_embeddings.frequency\_scale & Log-uniform $[0.005, 10.0]$ \\
\quad num\_embeddings.d\_embedding & Integer $[16, 64]$ \\
\quad lr & Log-uniform $[0.0001, 0.005]$ \\
\quad weight\_decay & Log-uniform $[0.0001, 0.1]$ (optional) \\
\quad architecture & PLREmbeddings (lite), MLP backbone, tabm, $k=32$ \\
\midrule
\multicolumn{2}{l}{\textbf{Trompt}} \\
\quad n\_cycles & Integer $[3, 8]$ \\
\quad $d$ & Categorical $\{32, 64, 128, 256\}$ \\
\quad $P$ & Categorical $\{32, 64, 128, 256\}$ \\
\quad lr & Log-uniform $[10^{-5}, 0.001]$ \\
\quad weight\_decay & Log-uniform $[10^{-6}, 0.001]$ \\
\midrule
\multicolumn{2}{l}{\textbf{ExcelFormer}} \\
\quad n\_layers & Integer $[2, 5]$ \\
\quad d\_token & Categorical $\{8, 16, 32, 64, 128\}$ \\
\quad attention\_dropout & Uniform $[0.0, 0.5]$ \\
\quad ffn\_dropout & Uniform $[0.0, 0.5]$ \\
\quad residual\_dropout & Uniform $[0.0, 0.5]$ (optional) \\
\quad mix\_type & Categorical $\{$none, feat\_mix, hidden\_mix$\}$ \\
\quad lr & Log-uniform $[10^{-5}, 0.001]$ \\
\quad weight\_decay & Log-uniform $[10^{-6}, 0.001]$ \\
\midrule
\multicolumn{2}{l}{\textbf{ModernNCA}} \\
\quad dim & Integer $[64, 1024]$ \\
\quad d\_block & Integer $[64, 1024]$ \\
\quad n\_blocks & Integer $[0, 2]$ (optional) \\
\quad dropout & Uniform $[0.0, 0.5]$ \\
\quad num\_embeddings.n\_frequencies & Integer $[16, 96]$ \\
\quad num\_embeddings.frequency\_scale & Log-uniform $[0.005, 10.0]$ \\
\quad num\_embeddings.d\_embedding & Integer $[16, 64]$ \\
\quad lr & Log-uniform $[10^{-5}, 0.1]$ \\
\quad weight\_decay & Log-uniform $[10^{-6}, 0.001]$ (optional) \\
\end{longtable}

}

\end{document}